\documentclass[preprint,12pt]{elsarticle}
\usepackage[utf8]{inputenc}
\usepackage{times}
\usepackage{amsthm,amsmath,amsfonts,latexsym,amssymb}
\usepackage{graphics,fancyhdr,graphicx}
\usepackage[percent]{overpic}
\usepackage{multirow}
\usepackage{float}
\usepackage{fleqn}
\usepackage{subcaption}
\usepackage{mathtools}
\usepackage{algorithm}
\usepackage{mathrsfs}
\usepackage{makecell}
\usepackage[noend]{algpseudocode}
\usepackage{verbatim}
\usepackage{rotating}
\usepackage{bm}
\usepackage{color}

\usepackage{colortbl}
\usepackage[table]{xcolor}

\usepackage{geometry}

\usepackage[onehalfspacing]{setspace}

\newif\ifrevision
\revisionfalse

\definecolor{darkteal}{RGB}{0,120,120}
\definecolor{listinggray}{gray}{0.9}
\definecolor{lbcolor}{rgb}{0.9,0.9,0.9}
\definecolor{Darkgreen}{RGB}{0,100,0}

\ifrevision
    \newcommand{\tcb}[1]{\textcolor{blue}{#1}}  
    \newcommand{\tcr}[1]{\textcolor{red}{#1}}   
    \newcommand{\tcd}[1]{\textcolor{darkteal}{#1}} 
\else
    \newcommand{\tcb}[1]{#1}
    \newcommand{\tcr}[1]{#1}
    \newcommand{\tcd}[1]{#1}
\fi

\biboptions{numbers,sort&compress}

\begin{document}

\begin{frontmatter}

\title{Physics-Informed Laplace Neural Operator for Solving Partial Differential Equations}

\author[aff1]{Heechang Kim}
\ead{heechangkim@postech.ac.kr}
\author[aff2,aff4]{Qianying Cao}
\ead{qianying_cao@brown.edu}
\author[aff1]{Hyomin Shin}
\ead{zhainl@postech.ac.kr}
\author[aff3]{Seungchul Lee}
\ead{seunglee@kaist.ac.kr}
\author[aff2]{George Em Karniadakis}
\ead{george_karniadakis@brown.edu}
\author[aff1]{Minseok Choi\corref{cor1}}
\cortext[cor1]{Corresponding author}
\ead{mchoi@postech.ac.kr}

\affiliation[aff1]{organization={Department of Mathematics, Pohang University of Science and Technology (POSTECH)},
            city={Pohang},
            postcode={37673},
            country={Republic of Korea}}

\affiliation[aff2]{organization={Division of Applied Mathematics, Brown University},
            city={Providence},
            postcode={02906},
            state={Rhode Island},
            country={United States}}

\affiliation[aff3]{organization={Department of Mechanical Engineering, Korea Advanced Institute of Science and Technology (KAIST)},
            city={Daejeon},
            postcode={34141},
            country={Republic of Korea}}

\affiliation[aff4]{organization={Department of Mechanical Engineering, Rochester Institute of Technology},
            city={Rochester},
            postcode={14623},
            state={New York},
            country={United States}}

\begin{abstract}
    Neural operators have emerged as fast surrogate solvers for parametric partial differential equations (PDEs). However, purely data-driven models often require extensive training data and can generalize poorly, especially in small-data regimes and under unseen (out-of-distribution) input functions that are not represented in the training data. To address these limitations, we propose the \emph{Physics-Informed Laplace Neural Operator} (PILNO), which enhances the Laplace Neural Operator (LNO) by embedding governing physics into training through PDE, boundary condition, and initial condition residuals.
    To improve expressivity, we first introduce an \emph{Advanced LNO} (ALNO) backbone that retains a pole--residue transient representation while replacing the steady-state branch with an FNO-style Fourier multiplier. To make physics-informed training both data-efficient and robust, PILNO further leverages (i) \emph{virtual inputs}: an unlabeled ensemble of input functions spanning a broad spectral range that provides abundant physics-only supervision and explicitly targets out-of-distribution (OOD) regimes; and (ii) \emph{temporal-causality weighting}: a time-decaying reweighting of the physics residual that prioritizes early-time dynamics and stabilizes optimization for time-dependent PDEs.
    Across four representative benchmarks---Burgers' equation, Darcy flow, a reaction--diffusion system, and a forced KdV equation---PILNO consistently improves accuracy in small-data settings (e.g., $N_{\text{train}}\le 27$), reduces run-to-run variability across random seeds, and achieves stronger OOD generalization \tcb{with respect to input function statistics} than purely data-driven baselines.
\end{abstract}


\end{frontmatter}

%
%
\section{Introduction}\label{sec:intro}
    Recent advances in machine learning, supported by modern GPUs and software frameworks~\cite{abadi2016tensorflow, bradbury2018jax, paszke2019pytorch}, have accelerated progress in scientific computing for partial differential equations (PDEs)~\cite{raissi2019physics, li2020fourier, lu2021learning, jung2024ceens, faroughi2024physics, propp2025transfer}. In many applications, learned surrogate models complement classical numerical solvers~\cite{hughes2012finite} by enabling fast evaluation across varying geometries, parameters, and boundary/initial conditions~\cite{faroughi2024physics, cardoso2025exactly, de2022generic}. Within this landscape, operator learning has emerged as a promising paradigm: it aims to learn mappings between infinite-dimensional function spaces, providing fast surrogates for parametric PDEs in many scientific and engineering applications~\cite{kovachki2023neural, lu2022comprehensive}.

    Operator learning methods can be broadly grouped into two main directions. The first is exemplified by DeepONet~\cite{lu2021learning}, which leverages the universal approximation theorem for nonlinear operators~\cite{chen1995universal, back2002universal} and uses a branch--trunk architecture to encode the input functions and the coordinates of evaluation points. DeepONet’s branch--trunk decomposition has since motivated a broad class of extensions and variants~\cite{lu2022comprehensive, oommen2022learning, cao2024deep, kontolati2023influence, lee2024training, de2023bi}. In parallel, the second direction comprises integral--kernel neural operators~\cite{li2020multipole, li2020neural, li2020fourier, kovachki2023neural, tripura2022wavelet, cao2023lno, bonev2023spherical}, which generalize deep networks to the operator setting by composing learnable integral operators with pointwise nonlinearities. Representative examples include the Graph Neural Operator (GNO)~\cite{li2020neural}, the Fourier Neural Operator (FNO)~\cite{li2020fourier}, and the Laplace Neural Operator (LNO)~\cite{cao2023lno}, where kernel integration is parameterized via a pole--residue representation in the Laplace domain~\cite{lin1976probabilistic, hu2013signal, cao2023laplace}. These approaches have demonstrated strong empirical performance in a wide range of applications~\cite{pathak2022fourcastnet, ranade2022thermal, jin2025characterization, ahmadi2025physics}.

    Despite these successes, existing neural operators face two persistent challenges. \emph{(i) Data efficiency:} high accuracy typically requires many paired input--output samples, which may be expensive to obtain from high-fidelity simulations or experiments~\cite{hersbach2020era5, li2024physics,de2023bi}. In small-data regimes, limited supervision often fails to constrain fine-scale dynamics, and performance can degrade substantially~\cite{lee2024training, kontolati2023influence}. \emph{(ii) Robust generalization:} 
    \tcb{test error can increase sharply when test inputs differ from the training distribution, a setting commonly referred to as out-of-distribution (OOD) generalization~\cite{zhu2023reliable}.} These limitations motivate operator-learning approaches that remain reliable both in small-data regimes and under OOD inputs.

    Physics-informed operator learning addresses these issues by augmenting the training objective with PDE constraints~\cite{li2024physics, wang2021learning, goswami2023physics, navaneeth2024physics, faroughi2024physics}. Inspired by physics-informed neural networks (PINNs)~\cite{raissi2019physics}, such methods penalize PDE, boundary-condition, and initial-condition residuals evaluated on model predictions, guiding the learned operator toward physically consistent solutions even when observations are scarce or partially missing~\cite{rosofsky2023applications, kaewnuratchadasorn2024physics, jiao2024solving}. For example, PINO~\cite{li2024physics} uses multi-resolution residual evaluation to improve accuracy, and PI-DeepONet~\cite{wang2021learning} improves performance when training and testing conditions differ moderately. Zhu et al.~\cite{zhu2023reliable} study extrapolation systematically and propose physics- and data-informed adaptation strategies to correct pre-trained operators under distribution shift.
    
    However, existing physics-informed operator frameworks remain limited in how they address small-data training and better generalization to OOD inputs. First, many approaches incorporate physics primarily on the given training distribution and then rely on fine-tuning to handle extrapolation~\cite{zhu2023reliable}, rather than training the operator from the outset to be robust across a deliberately broadened family of inputs (e.g., sweeping over correlation length-scales or spectral content). Second, many physics-informed operator methods still require intermediate-to-large training data sets~\cite{wang2021learning,li2024physics} and rarely quantify how test performance scales as the number of training data decreases. In particular, there is little quantitative analysis of the extent to which physics-informed operator architectures can temper the growth of test error as training data become extremely scarce. These gaps open a central question: to what extent can physics-informed operator learning genuinely improve data efficiency and robustness on test inputs that differ from the training distribution, compared with purely data-driven neural operators?

    In this work, we propose the \emph{Physics-Informed Laplace Neural Operator} (PILNO), which builds on an Advanced Laplace Neural Operator (ALNO) backbone and incorporates physics-informed training with two mechanisms designed for data efficiency and better generalization. ALNO retains the pole--residue transient component of LNO~\cite{lin1976probabilistic, kreyszig2008advanced, hu2013signal, hu2016pole} while replacing the steady-state response with an FNO-style Fourier multiplier~\cite{li2020fourier}, improving expressivity without sacrificing interpretability. Note that ALNO was previously outlined in the supplementary material of our prior work~\cite{cao2023lno} and is fully developed in this paper. On top of ALNO, PILNO incorporates: (i) \emph{virtual inputs}, an unlabeled ensemble of admissible inputs spanning a broad range of input spectra, which provides abundant physics-only supervision and explicitly targets OOD regimes; and (ii) \emph{temporal-causality weighting}~\cite{wang2022respecting, jung2024ceens}, a time-decaying reweighting of the physics residual that emphasizes early-time dynamics at the beginning of the training, moves forward in time, and stabilizes optimization for time-dependent PDEs. By combining PDE/BC/IC residuals with supervised data when available, PILNO reduces dependence of labeled training data and improves physical fidelity relative to purely data-driven LNO, particularly in small-data and OOD settings. While physics-informed losses, virtual inputs, and causal training strategies have been explored in the PINNs, to the best of our knowledge, this work is the first to systematically integrate these mechanisms into a Laplace neural operator with explicit transient-steady separation, and quantify their impact on operator generalization, data efficiency, and OOD robustness. Our main contributions are:

    \begin{itemize}
        \item \textbf{\tcb{Advanced Laplace Neural Operator}.} We provide a complete formulation of ALNO with an explicit transient--steady decomposition: a pole--residue transient representation and a Fourier-multiplier steady-state representation, yielding a more expressive and efficient backbone.
        \item \textbf{Virtual inputs.} We introduce a label-free input ensemble that broadens the training spectrum, improves data efficiency and OOD generalization compared to data-driven LNO.
        \item \textbf{Temporal-causality weighting. (semigroup-aware residual weighting)} We propose a causal, time-decaying weighting of the physics residual that emphasizes early-time dynamics; from an operator viewpoint, this stabilizes the learned evolution operators and mitigates error accumulation in long-horizon predictions, and stabilizes optimization.
        \item \textbf{Numerical experiments.} We systematically evaluate and ablate these components at the operator level on four representative PDE benchmarks, quantifying gains in accuracy in small-data regimes, robustness across random seeds, and OOD generalization.
    \end{itemize}

    The rest of the paper is organized as follows. Section~\ref{sec:prelim} reviews LNO and introduces ALNO. Section~\ref{sec:pilno} presents PILNO, including physics residuals, virtual inputs, and temporal-causality weighting. Section~\ref{sec:numerical_results} reports numerical results on four representative examples: Burgers' equation (initial condition $\to$ solution mapping), Darcy flow (coefficient $\to$ solution mapping), reaction–diffusion system (forcing $\to$ solution mapping), and a forced Korteweg-de Vries (KdV) equation (forcing, boundary, initial condition combined $\to$ solution mapping) together with ablation studies on data efficiency and OOD generalization under a common evaluation protocol. Section~\ref{sec:conclusion} concludes with a summary, limitations, and directions for future work.

%
%
\section{Problem setting and Laplace Neural Operator Backbone}\label{sec:prelim}
    This section first formalizes the PDE problem setting and notation, \tcb{reviews the LNO formulation, and then introduces the ALNO backbone used in PILNO.}

\subsection{Problem setting and operator learning formulation}\label{subsec:problem_setting}
    Let $\mathcal{D}\subset \mathbb{R}^d$ be a spatial domain and $t \in [0,T]$ a time horizon. We consider (possibly parameterized) time-dependent PDEs of the form
    \begin{equation}\label{eq:pde_tuple}
        \begin{aligned}
        \frac{\partial u}{\partial t} + \mathcal{N}[u;\,c] &= f \quad &&\text{in }\mathcal{D}\times(0,T],\\
        \mathcal{B}(u) &= 0 \quad &&\text{on }\partial\mathcal{D}\times(0,T],\\
        u(\cdot,0) &= u_0, &&
        \end{aligned}
    \end{equation}
    where $u: \mathcal{D}\times[0,T] \to \mathbb{R}^m$ is the $m$-dimensional solution field, $\mathcal{N}$ is a (possibly nonlinear) differential operator, $f$ is a forcing term, and $\mathcal{B}$ is a boundary operator (periodic or Dirichlet/Neumann, depending on the problem). We denote the initial condition by  $u_0$ and collect the problem data in $a = (c, u_0, f) \in \mathcal{A}$, where $c$ represents PDE coefficients and $\mathcal{A}$ is the admissible input space. 

    The PDE~\eqref{eq:pde_tuple} induces a solution operator
    \begin{equation}
        \mathcal{G} : \mathcal{A} \to \mathcal{U}, \qquad a = (c,u_0,f) \mapsto u = \mathcal{G}(a),
    \end{equation}
    where $\mathcal{U}$ denotes a suitable space of spatio-temporal fields on $\mathcal{D}\times[0,T]$ (e.g. $\mathcal{U} = C([0,T];L^2(\mathcal{D};\mathbb{R}^m))$). We approximate $\mathcal{G}$ by a parametric neural operator $\mathcal{G}_\theta:\mathcal{A} \rightarrow \mathcal{U}$ with learnable parameters $\theta \in \Theta$:
    \begin{equation}
        \hat u = \mathcal{G}_\theta(a), \qquad \theta \in \Theta,
    \end{equation}
    and, for brevity, we henceforth omit the explicit dependence of $c, u_0, f$ on $a$.
    
    Let $S_{\text{data}}=\{(a^i,u^i)\}_{i=1}^{N_{\text{data}}}$ be a paired dataset, where each $a^i\in \mathcal{A}$ collects the coefficients, initial conditions, and forcing for a given PDE instance, and $u^i\in \mathcal{U}$ is the corresponding reference solution. In the purely data-driven setting, $\theta$ is learned by minimizing the empirical loss
    \begin{equation}
        \label{eqn:data_loss}
        \mathcal{L}_{\text{data}}(\theta) = \frac{1}{N_{\text{data}}}\sum_{i=1}^{N_{\text{data}}}
        \big\|\mathcal{G}_\theta(a^i)-u^i\big\|^2
    \end{equation}
    where $\| \cdot \|$ denotes the discrete $L_2$ norm over the space-time grid. We denote by $\theta^*$ a minimizer of $\mathcal{L}_\text{data}$ over $\Theta$. In Section~\ref{sec:pilno}, we extend this data-driven operator learning formulation to a physics-informed setting.
    
\subsection{Laplace Neural Operator}\label{subsec:lno}
    \tcb{We briefly recall the original LNO formulation proposed by Cao et al.~\cite{cao2023lno}.} For notational simplicity, in this subsection we focus on the special case where the only varying input is the forcing term $f$, and the PDE coefficients $c$ and the initial condition $u_0$ are fixed. First, the input $f\in\mathbb{R}^{d_x}$ is lifted to a higher-dimensional representation $v=\mathcal{P}(f)\in\mathbb{R}^{d_z}$ by a lift operator $\mathcal{P}$. Then, a single LNO layer computes
    \[
        u_1(t) = \sigma\big( (\mathcal{K}v)(t) + \mathcal{W}v(t) \big), \quad n = 1,2,\dots,d
    \]
    where \(\sigma\) is a nonlinear activation, \(\mathcal{W}\) is a linear transformation, and \(\mathcal{K}\) is a kernel integral operator
    \begin{equation}
        (\mathcal{K}v)(t) = \int_0^t \kappa(t-\tau)\,v(\tau)\,\mathrm{d}\tau 
        = (\kappa * v)(t),
        \label{eq:kernel_integral}
    \end{equation}
    under the simplifying assumption that the kernel is translation invariant, $\kappa(t,\tau) = \kappa(t-\tau)$. Applying the Laplace transform to the kernel integral of \eqref{eq:kernel_integral} yields
    \begin{equation}
        U_1(s) = \mathscr{L}\{(\kappa *v)(t)\}(s)
        = K_\phi(s)\,V(s),
        \label{eq:after_LT}
    \end{equation}
    where $V(s) = \mathscr{L}\{v\}(s)$. Here, $K_\phi(s)$ is directly parameterized by 
    \begin{equation}
        K_\phi(s) = \sum_{n=1}^N \frac{\beta_n}{s-\mu_n},
        \label{eq:K_phi}
    \end{equation}
    with learnable poles $\{\mu_n\}_{n=1}^N$ and residues $\{\beta_n\}_{n=1}^N$ (collectively $\phi = (\mu_1,\dots,\mu_N,\beta_1,\dots,\beta_N)$), where $N$ denotes the number of system modes. If we expand the input as a sum of exponentials
    \begin{equation}
        v(t) = \sum_{\ell} \alpha_\ell e^{i\omega_\ell t},
        \label{eq:expand_of_v}
    \end{equation}
    then applying the inverse Laplace transform to \eqref{eq:after_LT} leads to an explicit time-domain decomposition
    \begin{align}
        (\mathcal{K}^{\text{lno}}v)(t) &= \mathscr{L}^{-1} \left\{U_1(s) \right\}(t) \notag \\
        &= \underbrace{\sum_{n=1}^N 
            \Big( \sum_{\ell=-\infty}^\infty \frac{\beta_n \alpha_\ell}{\mu_n - i\omega_\ell} \Big)
            e^{\mu_n t}}_{\text{transient}}
        \;+\;
        \underbrace{\sum_{\ell=-\infty}^\infty 
            \Big( \sum_{n=1}^N \frac{\alpha_\ell \beta_n}{i\omega_\ell - \mu_n} \Big)
            e^{i\omega_\ell t}}_{\text{steady-state}}.
        \label{eq:lno_transient_steady}
    \end{align}
    The first term corresponds to a transient response governed by the system poles \(\{\mu_n\}\), while the second term captures the steady-state response operated in the frequency domain. After applying the LNO layer iteratively, we use a projection operator $\mathcal{Q}:\mathbb{R}^{d_z}\to\mathcal{U}$ to map the final latent representation back to the solution space. The same construction extends to more general inputs $a = (c,u_0,f)$ by lifting all components into the latent representation.

\subsection{\tcb{Advanced Laplace Neural Operator: decoupled transient-steady branches}}\label{subsec:alno}
    The Laplace layer output in Eq.~\eqref{eq:lno_transient_steady} decomposes the response into a transient and a steady-state term. This pole-residue representation provides a physically interpretable parameterization and enables LNO to capture non-periodic signals and rich transient dynamics more effectively than purely Fourier-based operators. However, because both components are tied to the same pole--residue parameterization, the original LNO can be overly restrictive in deeper architectures and limit the representation of complex dynamics, as we also observe empirically in Appendix~A. To increase flexibility while preserving interpretability, we retain the transient component in pole--residue form, where the poles encode characteristic time scales, and replace the steady component with a learnable Fourier multiplier $H(\omega)$ in the spirit of FNO~\cite{li2020fourier}.

    We next present a complete mathematical formulation of the idea---previously outlined in the supplementary material of our work~\cite{cao2023lno}---that the steady-state term of the LNO can be expressed as a Fourier multiplier. Applying the Fourier transform to the kernel integral in Eq.~\eqref{eq:kernel_integral} gives
    \begin{equation}
        \tilde{U}_1(\omega)
        = \mathcal{F}\{(k*v)(t)\}(\omega)
        = \mathcal{F}\{\kappa\}(\omega)\,\tilde{V}(\omega)
        = \tilde{K}_\phi(\omega)\,\tilde{V}(\omega),
    \end{equation}
    where $\tilde V(\omega) = \mathcal{F}\{v\}(\omega)$ and $\tilde K_\phi(\omega) = \mathcal{F}\{\kappa\}(\omega)$ denotes the Fourier transform of the kernel. Formally, by evaluating the Laplace transfer function on the imaginary axis, we recover this Fourier multiplier via
    \begin{equation}
        \tilde{K}_\phi(\omega) = K_\phi(i\omega)
    \end{equation}
    Using the expansion $v(t) = \sum_{\ell} \alpha_\ell e^{i\omega_\ell t}$, we have $\tilde V(\omega_\ell) = \alpha_\ell$, and on a discrete frequency grid $\{\omega_\ell\}$ we obtain
    \begin{equation}
        \tilde U_1(\omega_\ell) = K_\phi(i\omega_\ell)\,\tilde V(\omega_\ell) = K_\phi(i\omega_\ell)\alpha_\ell
    \end{equation}
    Finally, the inverse discrete Fourier transform yields
    \begin{equation}
        \mathcal{F}^{-1}\{\tilde{U}_1(\omega)\}(t)  = \sum_{\ell=-\infty}^{\infty} \tilde{U}_1(\omega_\ell)\,e^{i\omega_\ell t}
        = \sum_{\ell=-\infty}^{\infty} \alpha_\ell K_\phi(i\omega_\ell)\,e^{i\omega_\ell t}.
    \end{equation}

    rewriting $K_\phi(i\omega_\ell)$ by Eq.~\eqref{eq:K_phi} yields
    \begin{equation}
        \mathcal{F}^{-1}\{\tilde{U}_1(\omega)\}(t)  =\sum_{\ell=-\infty}^{\infty} \alpha_\ell K_\phi(i\omega_\ell)\,e^{i\omega_\ell t}
        = \sum_{\ell=-\infty}^\infty 
            \Big( \sum_{n=1}^N \frac{\alpha_\ell \beta_n}{i\omega_\ell - \mu_n} \Big)
            e^{i\omega_\ell t}
        \label{eq:FNO_formula}
    \end{equation}
    The representation in Eq.~\eqref{eq:FNO_formula} coincides with the steady-state term in Eq.~\eqref{eq:lno_transient_steady}. Consequently, an FNO layer with Fourier multiplier $H(\omega)$ acting as
    \[
        (\mathcal{K}^{\text{fno}}v)(t) = \mathcal{F}^{-1}\!\big[\,H(\omega)\,\mathcal{F}\{v\}(\omega)\,\big](t)
    \]
    recovers the steady–state branch of LNO when $H(\omega)$ is chosen to approximate $K_\phi(i\omega)$. In this sense, the LNO steady–state response can be viewed as a particular Fourier neural operator, while LNO augments it with an additional transient branch governed by the learned poles $\{\mu_n\}$.
    
    These observations motivate the \tcb{ALNO}, in which we explicitly decouple the two branches:
    
    \begin{equation}
        (\mathcal{K}^{\text{alno}}v)(t)
        = \sum_{n=1}^N \Big(\sum_{\ell=-\infty}^\infty \frac{\beta_n \alpha_\ell}{\mu_n - i\omega_\ell}\Big)\,e^{\mu_n t}
        \;+\;
        \mathcal{F}^{-1}\!\Big[\,H(\omega)\,\mathcal{F}\{v\}(\omega)\,\Big].
        \label{eq:alno_split}
    \end{equation}
    In ALNO, the transient branch remains a pole--residue system with learnable poles $\{\mu_n\}$ and residues $\{\beta_n\}$, while the steady branch is implemented as an FNO-style spectral convolution via a flexible multiplier $H(\omega)$. This decoupled design preserves the interpretability of LNO through explicit transient poles, while the more expressive spectral steady branch supports deeper and more accurate operator models.

%
%
\section{\tcb{Physics-Informed Laplace Neural Operator}}\label{sec:pilno}
    Building on the ALNO backbone introduced in Section~\ref{sec:prelim}, \tcb{we define PILNO as a physics-informed extension of ALNO. Specifically, PILNO augments the data-driven ALNO training objective with PDE, boundary-condition (BC), and initial-condition (IC) residuals. It further leverages label-free “virtual inputs” and a temporal-causality weighting scheme to improve data efficiency and robustness.} In the following, we detail the physics-informed objective, the construction of the virtual-input ensemble, and the temporal-causality weighting scheme.

    In addition, PILNO combines supervised data (if available) with physics/BC/IC residuals:
    \[
    \mathcal{L} = 
    \lambda_{\text{pde}}\mathcal{L}_{\text{pde}}
    +\lambda_{\text{bc}}\mathcal{L}_{\text{bc}}
    +\lambda_{\text{ic}}\mathcal{L}_{\text{ic}}
    +\underbrace{\lambda_\text{data}\mathcal{L}_{\text{data}}}_{\text{if available}}
    \]
    with definitions deferred to Sec.~\ref{subsec:pilno_residual}.
    
\subsection{Physics-informed residuals}\label{subsec:pilno_residual}
    We first define the physics-informed training objective. Following prior physics-informed operator learning methods~\cite{raissi2019physics,li2024physics,wang2021learning}, we evaluate PDE and BC/IC residuals directly on model predictions---via automatic differentiation or consistent finite differences---thereby turning the governing physics into additional supervision signals.
     
     A physics-only regime---training solely on PDE/BC/IC residuals without paired input-output data---is possible; however, when paired datasets are available, the data term typically supplies stronger, lower-variance gradients and a smoother optimization landscape. Given a paired dataset $S_{\text{data}} = \{(a^i,u^i)\}_{i=1}^{N_{\text{data}}}$, the data, PDE, boundary, and initial condition losses are defined as follows:
    \begin{align}
        \mathcal{L}_{\text{data}}(\theta)
          &= \frac{1}{N_{\text{data}}}\sum_{i=1}^{N_{\text{data}}}
              \big\| u_\theta^i - u^i \big\|_{2}^2, \\
        \mathcal{L}_{\text{pde}}(\theta)
          &= \frac{1}{N_{\text{data}}}\sum_{i=1}^{N_\text{data}}\left\| \frac{\partial u_\theta^i}{\partial t} + \mathcal{N}[u_\theta^i;c^i] - f^i \right\|_2^2, \\
        \mathcal{L}_{\text{ic}}(\theta)
          &= \frac{1}{N_{\text{data}}}\sum_{i=1}^{N_\text{data}}\left\| u_\theta^i(\cdot,0) - u^i(\cdot,0) \right\|_2^2, \\
        \mathcal{L}_{\text{bc}}(\theta)
          &= \frac{1}{N_{\text{data}}}\sum_{i=1}^{N_{\text{data}}}
              \big\| \mathcal{B}\big(u_\theta^i\big) \big\|_{2}^2,
    \end{align}
    where $u_\theta^i = \mathcal{G}_\theta(a^i)$. The base physics-informed objective is then
    \begin{equation}
        \mathcal{L}(\theta)
        = \lambda_{\text{data}}\mathcal{L}_{\text{data}}
        + \lambda_{\text{pde}}\mathcal{L}_{\text{pde}}
        + \lambda_{\text{ic}}\mathcal{L}_{\text{ic}}
        + \lambda_{\text{bc}}\mathcal{L}_{\text{bc}}.
    \end{equation}
    A purely physics-driven regime is recovered by setting $\lambda_{\text{data}}=0$, in which case no solution labels $u^i$ are available. By employing the above training loss, we can directly embed the governing physics into the training objective, thereby improving the model's learning performance. 
    
    \begin{figure}[ht]
        \centering
        \includegraphics[width=\linewidth]{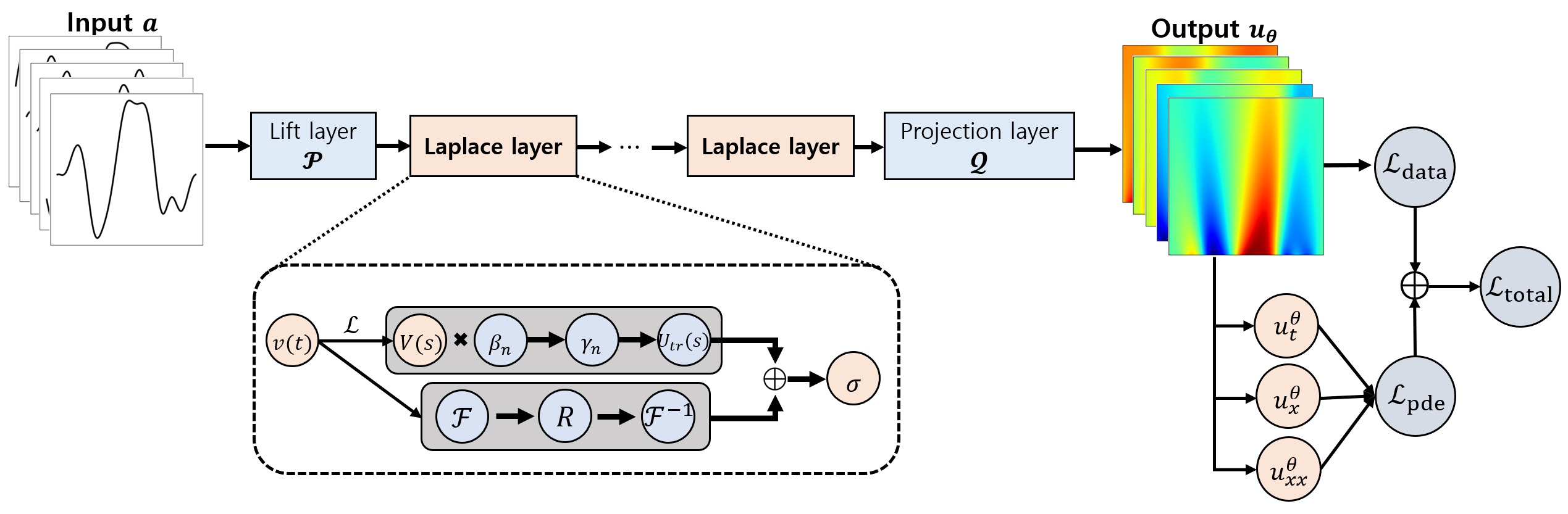}
        \caption{\textbf{Schematic of PILNO architecture.} Given input data $a$, a lifting operator $\mathcal{P}$ embeds the input into a higher-dimensional latent space; the temporal module applies an advanced laplace layer, and a projection map $\mathcal{Q}$ returns the output field. Training uses supervised data (when available) together with physics-based losses: PDE residuals (via derivatives such as \(u_t, u_x, u_{xx}\)) and boundary/initial-condition penalties. The total objective combines data loss with PDE residuals to enforce the governing equations during training.}
        \label{fig:pilno_structure}
    \end{figure}
    
\subsection{Virtual inputs}\label{subsec:virtual_dataset}
    Generating paired data (inputs with ground-truth solutions) is often expensive: each sample may require costly experiments or high-fidelity time-dependent simulations. In practice, this typically places neural operators in a small-data regime, where only a limited set of initial/forcing conditions and coefficients are observed. In such settings, we would like physics-informed training to compensate for the lack of labels and stabilize learning.

    However, when physics residuals are applied only to this small paired dataset, the supervision is effectively restricted to the narrow spectrum of inputs present in those samples (e.g., a limited range of correlation length-scales). We observe that this can bias the operator toward those few configurations, leading to overfitting and degraded generalization---both on \tcb{OOD} inputs and even on trained distribution test cases. In other words, naive physics-informed training does not by itself resolve the small-data problem and may even reinforce the spectral bias of the training distribution.

    \tcr{To decouple physics supervision from label availability, we make virtual inputs a central component of PILNO. We construct an unlabeled input ensemble—"virtual inputs"—by sampling admissible problem data \(a\) (typically the initial condition, and when relevant the forcing and PDE coefficients) without computing the corresponding reference solutions. These virtual inputs contribute only through the PDE, IC, and BC residuals, thereby providing label-free physics supervision.}

\begin{figure}[!htbp]
        \centering
        \begin{overpic}[width=0.9\linewidth]{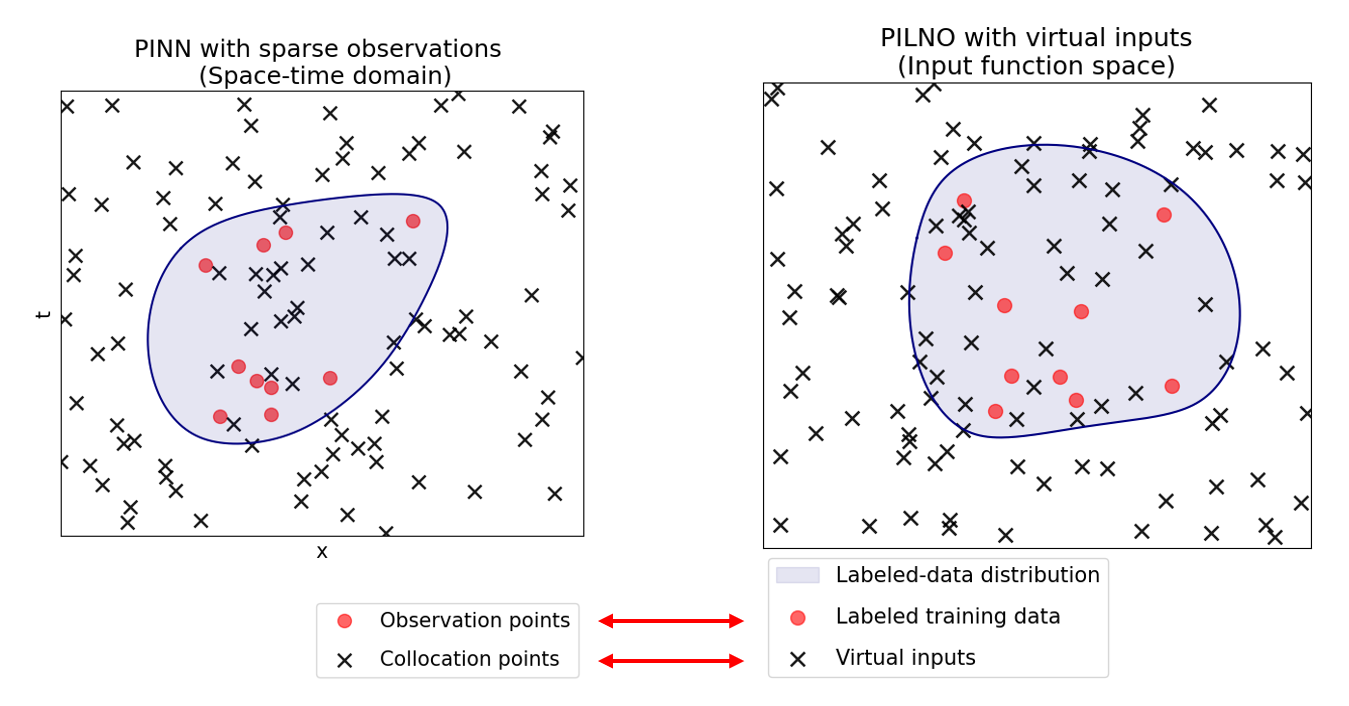}
            \put(23, -0.7){\scriptsize (a)}
            \put(78, -0.7){\scriptsize (b)}
        \end{overpic}

        \caption{\textbf{Schematic illustration of virtual inputs in PILNO.}
(a) In sparse-observation PINNs, collocation points supplement limited solution observations by enforcing physics residuals over the space--time domain.
(b) In PILNO, virtual inputs supplement limited labeled training data by enforcing physics residuals over the admissible input-function space.
}
        \label{fig:virtual_inputs}
    \end{figure}

    \tcr{The role of virtual inputs in PILNO is analogous to that of collocation points in sparse-observation PINNs, but lifted from the space-time domain to the input-function space. In PINNs, sparse observation points may cover only a limited portion of the space-time domain. Therefore, additional collocation points, where no solution observations are available, are introduced to provide label-free physics supervision and enforce the governing equations more broadly over the space-time domain. Similarly, in PILNO, a small paired dataset may cover only a limited portion of the admissible input-function space. We therefore introduce virtual inputs, for which no reference solutions are computed, and use them only through PDE, IC, and BC residuals. This extends label-free physics supervision beyond the labeled input distribution and helps constrain the learned operator over a broader family of admissible PDE inputs. Figure~\ref{fig:virtual_inputs} illustrates this analogy schematically; the input-function space in the PILNO panel should be interpreted as a low-dimensional schematic projection of the generally high- or infinite-dimensional admissible input space.}

    \tcr{Formally, we generate virtual inputs \(a_{\mathrm{virt}}\) from a prescribed virtual-input distribution or sampling rule, denoted by \(\mathbb{P}_{\mathrm{virt}}\), over the admissible input space \(\mathcal{A}\). The choice of \(\mathbb{P}_{\mathrm{virt}}\) can be adapted to the structure of the target problem and the intended deployment regime. For example, virtual inputs may be drawn from Gaussian random fields with prescribed spectrum/energy, generated using Latin hypercube sampling over finite-dimensional PDE parameters, or designed using problem-specific prior knowledge. The key requirement is that the generated virtual inputs remain admissible problem data, so that the PDE, IC, and BC residuals can be evaluated without requiring reference solution labels.}

    \tcr{In this work, we instantiate \(\mathbb{P}_{\mathrm{virt}}\) using Gaussian random fields. Whereas many physics-informed operator-learning approaches apply residual constraints primarily on the available training distribution~\cite{li2024physics, wang2021learning}, we choose \(\mathbb{P}_{\mathrm{virt}}\) so that the virtual ensemble spans broad spectral ranges, for example by varying correlation length-scales. This design provides label-free physics supervision over a deliberately broadened input range rather than only over the narrow paired training distribution and, as shown in Section~\ref{sec:numerical_results}, improves robustness to OOD input regimes.}

    Let \(\mathcal{L}_\text{data}, \mathcal{L}_\text{pde}, \mathcal{L}_\text{ic}, \mathcal{L}_\text{bc}\) be defined as in Section~\ref{subsec:pilno_residual}. During training, each mini-batch mixes paired and virtual samples. Paired samples \((a^i,u^i)\) contribute both data and physics losses. Virtual samples $a_{\text{virt}}^j$ contribute only physics terms,
    \begin{align}
      \mathcal{L}_{\text{pde}}^{\text{virt}}(\theta)
        &= \frac{1}{N_{\text{virt}}}\sum_{j=1}^{N_{\text{virt}}}
           \Big\| \frac{\partial u_\theta^j}{\partial t} + \mathcal{N}[u_\theta^j;c^j] - f^j \Big\|_2^2, \\
      \mathcal{L}_{\text{ic}}^{\text{virt}}(\theta)
        &= \frac{1}{N_{\text{virt}}}\sum_{j=1}^{N_{\text{virt}}}
           \big\| u_\theta^j(\cdot,0) - u_0^j \big\|_2^2, \\
      \mathcal{L}_{\text{bc}}^{\text{virt}}(\theta)
        &= \frac{1}{N_{\text{virt}}}\sum_{j=1}^{N_{\text{virt}}}
           \big\| \mathcal{B}(u_\theta^j) \big\|_2^2,
    \end{align}
    and the total objective is
    \begin{equation}
      \mathcal{L}(\theta)
      = \lambda_{\text{data}}\mathcal{L}_{\text{data}}
      + \lambda_{\text{pde}}\big(\mathcal{L}_{\text{pde}} + \mathcal{L}_{\text{pde}}^{\text{virt}}\big)
      + \lambda_{\text{ic}}\big(\mathcal{L}_{\text{ic}} + \mathcal{L}_{\text{ic}}^{\text{virt}}\big)
      + \lambda_{\text{bc}}\big(\mathcal{L}_{\text{bc}} + \mathcal{L}_{\text{bc}}^{\text{virt}}\big).
    \end{equation}

    Because virtual inputs are label-free, we can generate them abundantly and diversify their spectra, which (i) \tcb{supplies rich label-free supervision without requiring additional reference solutions,} (ii) improves data efficiency in the small-data regime, and (iii) \tcb{improves generalization beyond the paired data distribution by providing label-free physics supervision over a broader input range.} Conversely, when virtual inputs are absent and physics residuals are applied only to a small set of paired instances, the supervision concentrates on a narrow input spectrum; empirically, we observe that this exacerbates overfitting and degrades generalization in the small-data regime (see Section~\ref{subsec:darcy}).
        
\subsection{Temporal causality weighting}\label{subsec:time_causal}
    When computing physics losses for time-dependent PDEs, it is important to respect temporal causality: minimizing the residual at late times can be misleading if early-time predictions remain inaccurate, because early discrepancies can dominate downstream dynamics (“late-time masking”). Following causal training strategies~\cite{wang2022respecting,jung2024ceens}, we introduce temporal-causality weighting (TCW), which biases the physics residual toward early times so that the model first fits the short-time evolution before learning later states. This choice can also be interpreted from an operator viewpoint through error propagation under repeated composition of rollout-based operators.

    \begin{equation}\label{eq:rewrite_pde_residual}
        \begin{aligned}
        \mathcal{L}_\text{pde}(\theta;a^i) &= \left\| \frac{\partial u_\theta^i}{\partial t} + \mathcal{N}[u_\theta^i] \right\|^2 \\
        &= \sum_{k=0}^K \left\| \frac{\partial u_\theta^i}{\partial t}(\cdot,t_k) + \mathcal{N}[u_\theta^i](\cdot,t_k)  \right\|^2 \\
        &= \sum_{k=0}^K \mathcal{L}_r(t_k, \theta;a^i)
        \end{aligned}
    \end{equation}
    We denote by \(\mathcal{L}_r(t_k,\theta;a^i)\) the per-time-step residual at time \(t_k\), so that \(\mathcal{L}_\text{pde}\) is the sum of these per-time-step terms over the temporal grid. To enforce temporal causality, we reweight these contributions by a positive, monotonically decreasing schedule $w_k \coloneqq w(t_k)$. We then normalize the weights so that their average is one, which keeps the overall loss scale stable:
    \[
        \tilde w_k \;=\; \frac{w_k}{\frac{1}{K+1}\sum_{j=0}^K w_j}, \qquad k=0,\dots,K.
    \]
    Finally the causal PDE loss is defined as
    \begin{equation}\label{eq:temporal_causality_weighting}
        \mathcal{L}^{\text{causal}}_{\text{pde}}(\theta;a^i)
        = \sum_{k=0}^K \tilde w_k\,\mathcal{L}_r(t_k,\theta;a^i).
    \end{equation}

    Typical choices for $w(t)$ include
    \begin{equation}
        \label{eq:time_weight_forms}
        \text{(Exp)}\quad w(t)=e^{-\gamma t}, 
        \quad
        \text{(Inv)}\quad w(t)=\frac{1}{1+\alpha t},
        \quad
        \text{(Piecewise)}\quad 
        w(t)=
        \begin{cases}
        w_{\max}, & t\le t_0,\\
        w_{\min}, & t> t_0,
        \end{cases}
    \end{equation}

    with hyperparameters $\gamma,\alpha>0$, $w_{\max}>w_{\min}>0$, and a cutoff $t_0$. Adaptive weighting based on gradients or scale of each per-time-step residual is also possible. 

    \tcb{TCW} stabilizes optimization for time-dependent PDEs by emphasizing early-time residuals, preventing late-time minimization from masking early errors. \tcb{This idea is inspired by causal training strategies for PINNs~\cite{wang2022respecting}, but the present use differs in its level of application. In causal training for PINNs, the objective is to learn the solution of a single PDE instance with a coordinate-based network ($u_\theta(x,t)$), and temporal weighting is used to enforce a causal training order along that single trajectory. In contrast, PILNO learns a solution operator ($\mathcal{G}_\theta : a \mapsto u$) over a family of input functions; hence TCW reweights time-sliced residuals evaluated over labeled and virtual input functions. Thus, in PILNO, TCW acts as an operator-level temporal curriculum rather than a hard time-marching constraint. In our implementation, this curriculum is realized by a normalized monotone time-decaying schedule, which is simple and stable in mini-batch operator learning. In practice, combining TCW with the physics residual and virtual inputs yields faster and more stable training, and improves long-horizon accuracy in the benchmarks of Section~\ref{sec:numerical_results}.} 
    
    To further motivate TCW from an operator perspective, consider a discrete-time flow map $u(\cdot,t_{k+1}) = \mathcal{S}_{\Delta t}(u(\cdot, t_k))$ (a semigroup in the autonomous case), and a learned one-step operator $\mathcal{R}_\theta$ used in rollout form $u_\theta(\cdot,t_{k+1}) = \mathcal{R}_\theta(u_\theta(\cdot,t_k))$. Writing the rollout error $e_k \coloneqq u_\theta(\cdot, t_{k}) - u(\cdot,t_k)$ and the one-step defect $\delta_k \coloneqq \mathcal{R}_\theta (u(\cdot,t_k)) - \mathcal{S}_{\Delta t}(u(\cdot,t_k))$, a first-order decomposition yields $e_{k+1} = \mathcal{R}_\theta (u_\theta(\cdot,t_k)) - \mathcal{R}_\theta(u(\cdot,t_k)) + \delta_k \approx J_ke_k + \delta_k$, where $J_k = \partial \mathcal{R}_\theta / \partial u$ is the Jacobian evaluated along the trajectory. Unrolling this recursion exposes the multiplicative (product) structure induced by composition:
    
    \begin{equation}
        e_K \approx \left( \prod_{j=0}^{K-1} J_j \right)e_0 + \sum_{k=0}^{K-1}\left( \prod_{j=k+1}^{K-1}J_j \right)\delta_k,
    \end{equation}
    
    so defects incurred at early times are propagated through longer Jacobian products and can dominate long-horizon error. In particular, if $\mathcal{R}_\theta$ is (locally) $L$-Lipschitz, we can induce
    \begin{equation}
        \| e_K \| \leq L^K\|e_0\| + \sum_{k=0}^{K-1}L^{K-1-k}\| \delta_k \|,
    \end{equation}
    showing that early-time defects can have a larger downstream influence due to repeated composition. Although PILNO is not an explicit autoregressive rollout model (it predicts the full space-time field in one forward pass), our physics loss still decomposes across time slices (Eq.~\eqref{eq:rewrite_pde_residual}); without temporal causality weighting, optimization can reduce late-time residuals while leaving early-time inconsistencies that would dominate under causal evolution. TCW serves as a tractable surrogate for these downstream sensitivity factors by prioritizing early-time residuals, thereby promoting causally coherent long-horizon predictions.

    Figure~\ref{fig:pilno_structure} presents a schematic of PILNO's core architecture. In addition, Algorithm~\ref{alg:pilno} details the training workflow, including how virtual inputs are integrated and how temporal-causality weighting is applied within the physics-informed loss. 

    To mitigate overfitting and enhance generalization performance, we construct a physics dataset by augmenting the labeled training inputs $a$ with virtual inputs $a_{\text{virt}}$. Formally,
    \[
        \{a_{\text{phy}}^k\}_{k=1}^{N_{\text{train}}+N_{\text{virt}}}
        := \{a^i\}_{i=1}^{N_{\text{train}}} \cup \{a_{\text{virt}}^j\}_{j=1}^{N_{\text{virt}}}.
    \]
    We then evaluate the physics-based losses $\mathcal{L}_{\text{pde}}, \mathcal{L}_{\text{bc}}, \mathcal{L}_{\text{ic}}$ on $\{a_{\text{phy}}^k\}$, while the data loss $\mathcal{L}_{\text{data}}$ is computed only on the labeled subset $\{(a^i,u^i)\}_{i=1}^{N_{\text{train}}}$.

\begin{algorithm}[htbp]
    
    \caption{Training procedure for PILNO}
    \textbf{Input:} training set $\{(a^i,u^i)\}_{i=1}^{N_{\text{train}}}$,
    virtual inputs $\{a_{\text{virt}}^j\}_{j=1}^{N_{\text{virt}}}$ (optional),
    weights $\lambda_{\text{data}},\lambda_{\text{pde}},\lambda_{\text{bc}},\lambda_{\text{ic}}$,
    temporal weights $w_k = w(t_k)$ (optional)
    
    \begin{algorithmic}[1]
    \State Construct physics set $\{a_{\text{phy}}^k\} = \{a^i\} \cup \{a_{\text{virt}}^j\}$
    \State Initialize parameters $\theta$
    \For{$\text{epoch} = 1,\dots,E$}
      \For{each batch $a_{\text{phy}}$ from physics set}
        \State Sample (optionally) paired batch $(a,u)$ from training set
        \State Compute prediction $u_\theta = \mathcal{G}_\theta(a_{\text{phy}})$
        \If{temporal causality weighting is used}
          \State $\ell_{\text{pde}} \gets \mathcal{L}_{\text{pde}}^{\text{causal}}(\theta; a_{\text{phy}})$
        \Else
          \State $\ell_{\text{pde}} \gets \mathcal{L}_{\text{pde}}(\theta; a_{\text{phy}})$
        \EndIf
        \State Compute $\ell_{\text{bc}}, \ell_{\text{ic}}$ on $a_{\text{phy}}$
        \If{paired labels are available in batch}
          \State $\ell_{\text{data}} \gets \mathcal{L}_{\text{data}}(\theta; a,u)$
        \Else
          \State $\ell_{\text{data}} \gets 0$
        \EndIf
        \State $\ell_{\text{total}} \gets \lambda_{\text{data}} \ell_{\text{data}}
              + \lambda_{\text{pde}} \ell_{\text{pde}}
              + \lambda_{\text{bc}} \ell_{\text{bc}}
              + \lambda_{\text{ic}} \ell_{\text{ic}}$
        \State Update $\theta$ using $\nabla_\theta \ell_{\text{total}}$
      \EndFor
    \EndFor
    \end{algorithmic}
    \label{alg:pilno}
\end{algorithm}

%
%
\section{Numerical Results}\label{sec:numerical_results}

\paragraph{Experimental setup and baselines}
    We evaluate PILNO in the small-data regime and under \tcb{OOD} tests on four canonical PDE benchmarks. Throughout all experiments, we instantiate \tcb{ALNO} from Section~\ref{sec:prelim} as the purely data-driven backbone, and build PILNO on top of the same architecture by adding physics-informed training, virtual inputs, and temporal-causality weighting. For notational convenience, we refer to this \tcb{data-driven} ALNO baseline simply as "LNO" unless otherwise specified. As summarized in Table~\ref{tab:results_table}, PILNO achieves substantially lower relative $L_2$ error than the LNO backbone in small-data settings ($N_{\text{train}}=25$ for Burgers and reaction–diffusion, $N_{\text{train}}=10$ for Darcy flow, and $N_\text{train}=27$ for forced KdV). Across all benchmarks, we enforce PDE/BC/IC residuals for PILNO and, when enabled, virtual inputs and temporal-causality weighting to further enhance stability and robustness for time-dependent problems: virtual inputs and temporal-causality for Burgers' equation, reaction-diffusion equation, and forced KdV equation, virtual inputs for Darcy flow.

\paragraph{Tasks and operator mappings}
    We consider four representative operator-learning tasks that span both time-dependent and steady-state problems and probe different types of input operators. 
    \begin{itemize}
        \item Burgers’ equation: learn the mapping from the initial condition $u_0(x)$ to the spatio-temporal solution $u(x,t)$, i.e., $u_0 \mapsto u(\cdot,\cdot)$.
        \item Darcy flow: learn the mapping from the coefficient/permeability field $a(x)$ to the steady pressure solution $u(x)$, i.e., $a \mapsto u(\cdot)$.
        \item Reaction–diffusion system: mapping either the initial condition $u_0(x)$ or the forcing profile $f(x)$ to the spatio-temporal solution $u(x,t)$.
        \item Forced \tcb{KdV} equation: learn the mapping from forcing term $f_x(x,t)$, initial condition $u_0(x)$, boundary condition $u(-L,t), u(L,t), u_x(L,t)$, and PDE coefficients $\alpha, \beta$ to the spatio-temporal solution $u(x,t)$. In short, $[f_x(x,t),u(L,t),u(-L,t),u_x(L,t),u(x,0),\alpha,\beta] \mapsto u(x,t)$.
    \end{itemize}
    This suite of benchmarks therefore covers the mappings of the IC$\to$solution, coefficient$\to$solution, forcing$\to$solution, as well as more complex mixed regimes where initial condition, boundary conditions, coefficients and forcing are provided jointly as inputs to the solution,  providing a diverse set of operator regimes to assess data efficiency and \tcb{OOD} generalization. Table~\ref{tab:tasks} summarizes the operator-learning task, data-efficiency setting, and generalization test setting for each benchmark.

    \begin{table}[!htbp]
        \centering
        \small
        \setlength{\tabcolsep}{4pt}
        \renewcommand{\arraystretch}{1.2}
        \begin{tabular}{l|c|c|c}
        \hline
        Benchmark
        & \makecell[c]{Operator map}
        & Data-efficiency setting
        & Generalization / test setting \\
        \hline
        Burgers' equation
        & \makecell[c]{Initial condition, \\ \(u_0(x) \mapsto u(x,t)\)}
        & \makecell[c]{Fixed length-scale \(\ell=0.5\); \\ varying \(N_{\mathrm{train}}\)}
        & \makecell[c]{OOD over initial-condition statistics: \\
        \(\ell_{\mathrm{train}},\ell_{\mathrm{test}} \in \{0.5,1.0,\dots,5.0\}\)} \\
        \hline
        Darcy flow
        & \makecell[c]{Coefficient field, \\ \(a(x) \mapsto u(x)\)}
        & \makecell[c]{Fixed training distribution; \\ varying \(N_{\mathrm{train}}\)}
        & \makecell[c]{In-distribution test over unseen \\ permeability realizations; \\no OOD test} \\
        \hline
        \makecell[l]{Reaction-diffusion\\Task A}
        & \makecell[c]{Initial condition, \\ \(u_0(x) \mapsto u(x,t)\)}
        & \makecell[c]{Not varied; \\ fixed \(N_{\mathrm{train}}=25\)}
        & \makecell[c]{OOD over initial-condition statistics: \\
        \(\ell_{\mathrm{train}},\ell_{\mathrm{test}} \in \{0.5,1.0,\dots,5.0\}\)} \\
        \hline
        \makecell[l]{Reaction-diffusion\\Task B}
        & \makecell[c]{Forcing term, \\ \(f(x) \mapsto u(x,t)\)}
        & \makecell[c]{Fixed length-scale \(\ell=0.5\); \\ varying \(N_{\mathrm{train}}\)}
        & \makecell[c]{OOD over forcing-function statistics: \\
        \(\ell_{\mathrm{train}},\ell_{\mathrm{test}} \in \{0.5,1.0,\dots,5.0\}\)} \\
        \hline
        Forced KdV
        & \makecell[c]{Mixed inputs, \\
        \([f_x,u_0,\mathrm{BC},\alpha,\beta]\)\\
        \(\mapsto u(x,t)\)}
        & \makecell[c]{Fixed analytic families; \\ varying \(N_{\mathrm{train}}\)}
        & \makecell[c]{In-distribution test over the same \\ analytic families; no OOD test} \\
        \hline
        \end{tabular}
        \caption{Summary of benchmark problems, operator mappings, data-efficiency settings, and generalization test settings.}
        \label{tab:tasks}
    \end{table}

\paragraph{Architectures and training protocol}
    All models are initialized with the Glorot–normal scheme and trained for up to $E=1000$ epochs. We use the Adam optimizer with $(\beta_1=0.9,\ \beta_2=0.999)$, weight decay $10^{-4}$, and a step learning-rate scheduler (step size 100, decay factor $\gamma=0.5$). \tcb{We use Adam rather than full-batch L-BFGS because PILNO optimizes a mini-batch operator-level objective over many input functions including virtual inputs, making full-batch quasi-Newton updates memory-intensive in our setting.} Unless otherwise noted, PILNO and LNO share this optimizer and schedule, and are matched in parameter count to ensure a fair capacity comparison. Problem-specific architectural details (e.g., network width, number of modes in the spectral layers, and temporal-causality weighting schedules) are reported in the corresponding subsections for each PDE.

    \begin{table}[ht]
        \centering
        \begin{tabular}{l|c|c}
            \hline
            \multicolumn{1}{c|}{\multirow{2}{*}{Example}} &
              \multicolumn{2}{c}{Relative $L_2$ error} \\
              \cline{2-3}
            \multicolumn{1}{c|}{} &
              \multicolumn{1}{c|}{PILNO} &
              \multicolumn{1}{c}{Data-driven baseline} \\
              \hline
            Burgers & 1.420e-2 & 9.386e-2 \\
            Darcy & 1.235e-2 & 1.467e-1  \\
            Reaction-Diffusion & 2.563e-2 & 1.229e-1 \\
            Forced KdV & 1.372e-1 & 4.713e-1 \\
            \hline
        \end{tabular}
        \caption{Relative $L_2$ errors of PILNO and data-driven baselines over 5 experiments on small training data for four PDE problems ($N_{\text{train}}=25$ for Burgers, Reaction-Diffusion, $N_{\text{train}}=10$ for Darcy flow, and $N_\text{train}=27$ for forced KdV). The data-driven baseline is LNO for Burgers, Darcy flow, and reaction--diffusion, and DeepOMamba for forced KdV.}
        \label{tab:results_table}
    \end{table}
    
\paragraph{Data generation and evaluation protocols}
    For Burgers' equation, Darcy flow, and the reaction-diffusion examples, inputs are sampled from Gaussian random fields with an exponentiated squared-sine kernel parametrized by a correlation length-scale \(\ell\). The forced KdV benchmark follows a separate analytic-family data-generation protocol described in Section~4.4. 

    Depending on the benchmark, we use the following evaluation protocols:
    \begin{itemize}
        \item Data-efficiency experiments: To isolate the effect on the number of paired samples, we fix a single length–scale (typically a relatively oscillatory field, small $\ell$) and vary only the training-set size $N_{\text{train}}$. These experiments quantify how rapidly the test error grows as paired data become scarce.
        \item OOD generalization experiments: To probe robustness across input statistics, we generate ten datasets with $\ell \in \{0.5,1.0,\dots,5.0\}$. For each $\ell_{\text{train}}$, a model is trained on the corresponding dataset and evaluated on the test sets of all $\ell_{\text{test}}$, yielding a $10\times 10$ grid of train–test combinations. The resulting error heatmaps summarize in-distribution (diagonal) and out-of-distribution (off-diagonal) performance.
    \end{itemize}
    The specific choice of $\ell$ used in each data-efficiency plot, and the precise role of $\ell_{\text{train}}$ and $\ell_{\text{test}}$ in the OOD heatmaps, are detailed in the subsections for each PDE.

\subsection{Burgers' equation}\label{subsec:burgers}
    The first example uses Burgers' equation to assess PILNO in terms of data efficiency and OOD generalization, compared to the purely data-driven LNO. Burgers’ equation is a prototypical nonlinear PDE arising in fluid mechanics and wave propagation:
    
    \begin{equation}
        \begin{aligned}
            &\frac{\partial u}{\partial t} + u\frac{\partial u}{\partial x} - \nu \frac{\partial ^2 u}{\partial x^2} = 0, \quad x\in[0,1], \quad t\in[0,1],\\
            &u(t,0) = u(t,1)
        \end{aligned}
        \label{eqn:burgers}
    \end{equation}
    where $u(t,x)$ denotes the solution field (e.g., velocity) and $\nu$ is the viscosity coefficient. We set $\nu = 0.01$ for this example. 

    Reference solutions are generated by a high-fidelity solver: spatial derivatives are approximated using a fourth-order central finite-difference scheme on a uniform grid with 128 points, and the resulting semi-discrete system is advanced in time using the classical explicit fourth-order Runge–Kutta method (RK4) with time step $\Delta t = 10^{-4}$. Initial conditions $u_0$ are sampled from a zero-mean Gaussian random field $\mu = \mathcal{GP}(0,k_{\text{exp-sin}})$, with an exponentiated sine-squared covariance kernel
    \[
        k_{\text{exp-sin}}(x,x') = \sigma^2
        \exp\!\left(
            -\frac{2\sin^2\big(\pi(x-x')/p\big)}{\ell^2}
        \right),
    \]
    with $p=1$ and $\sigma = 0.2$, and a length-scale parameter $\ell$ that controls the spatial smoothness. The operator-learning models are trained and evaluated on a downsampled space–time grid: the interval $[0,1]$ is discretized with step size $h = 1/25$, yielding $26$ snapshots $\{t_k\}_{k=0}^{25}$, and we use $N_x = 32$ equidistant spatial nodes in $[0,1]$. Each trajectory is therefore represented on a $26\times32$ grid. In all Burgers’ experiments, the input to LNO/PILNO is the initial condition $u_0(x)$ and the target is the corresponding spatio-temporal solution $u(x,t)$.

\paragraph{Data efficiency of PILNO}
    To assess performance in the small-data regime, we fix the length-scale at $\ell = 0.5$ and vary only the number of paired training samples $N_{\text{train}} \in \{25, 50, 100, 150, 200\}$. For each $N_{\text{train}}$, LNO and PILNO are trained on identical paired datasets; in addition, PILNO uses $N_{\text{virt}} = 1000$ virtual inputs drawn from the same Gaussian random field prior. All reported results are averaged over five random seeds.

\begin{figure}[h]
        \centering
        \includegraphics[height=8cm]{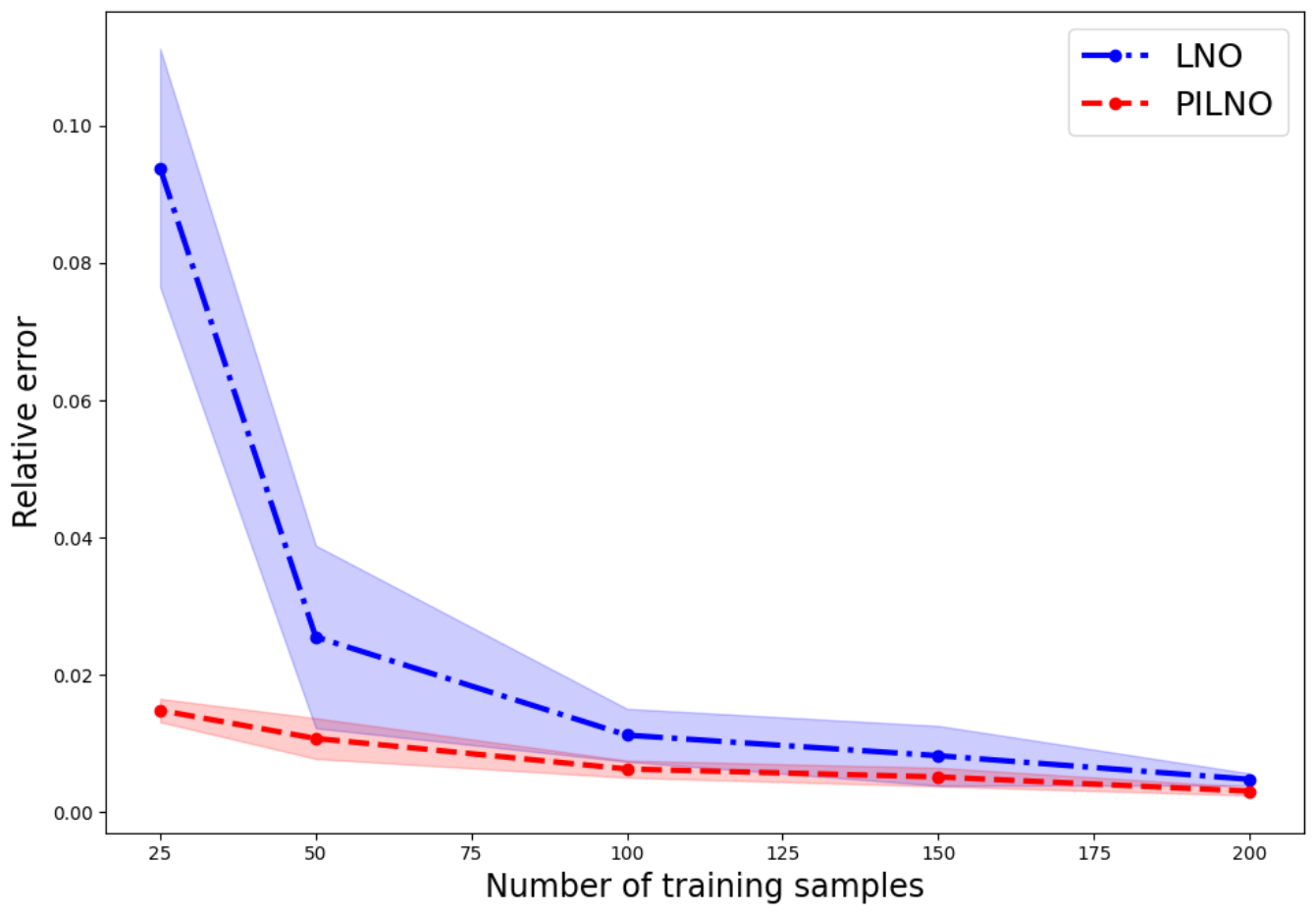}
        \caption{Relative $L_2$ errors over the number of training samples for Burgers' equation. Points denote means and shaded bands indicate $\pm1$ standard deviation over five random seeds. As $N_\text{train}$ decreases, the error of LNO increases sharply ($9.38\%$ at $N_\text{train}=25$), whereas PILNO maintains lower error level ($1.42\%$ at $N_\text{train}=25$).}
        \label{fig:burgers_pilno_ntrain}
    \end{figure}

    Figure~\ref{fig:burgers_pilno_ntrain} reports the relative $L_2$ test error versus $N_{\text{train}}$. As $N_{\text{train}}$ decreases, LNO’s error increases sharply (reaching approximately $9.35\%$ at $N_{\text{train}}=25$), while PILNO remains much more stable (about $1.42\%$ at $N_{\text{train}}=25$). In particular, PILNO with $25$ training samples attains a test error comparable to LNO trained with $100$ samples (roughly $1.1\%$), indicating a substantial gain in data efficiency. This improvement is attributable to the physics-informed structure and the additional supervision from virtual inputs, which together regularize the operator and curb overfitting in the small-data regime.
    
\begin{figure}[H]
        \centering
            \centering \includegraphics[height=7.5cm]{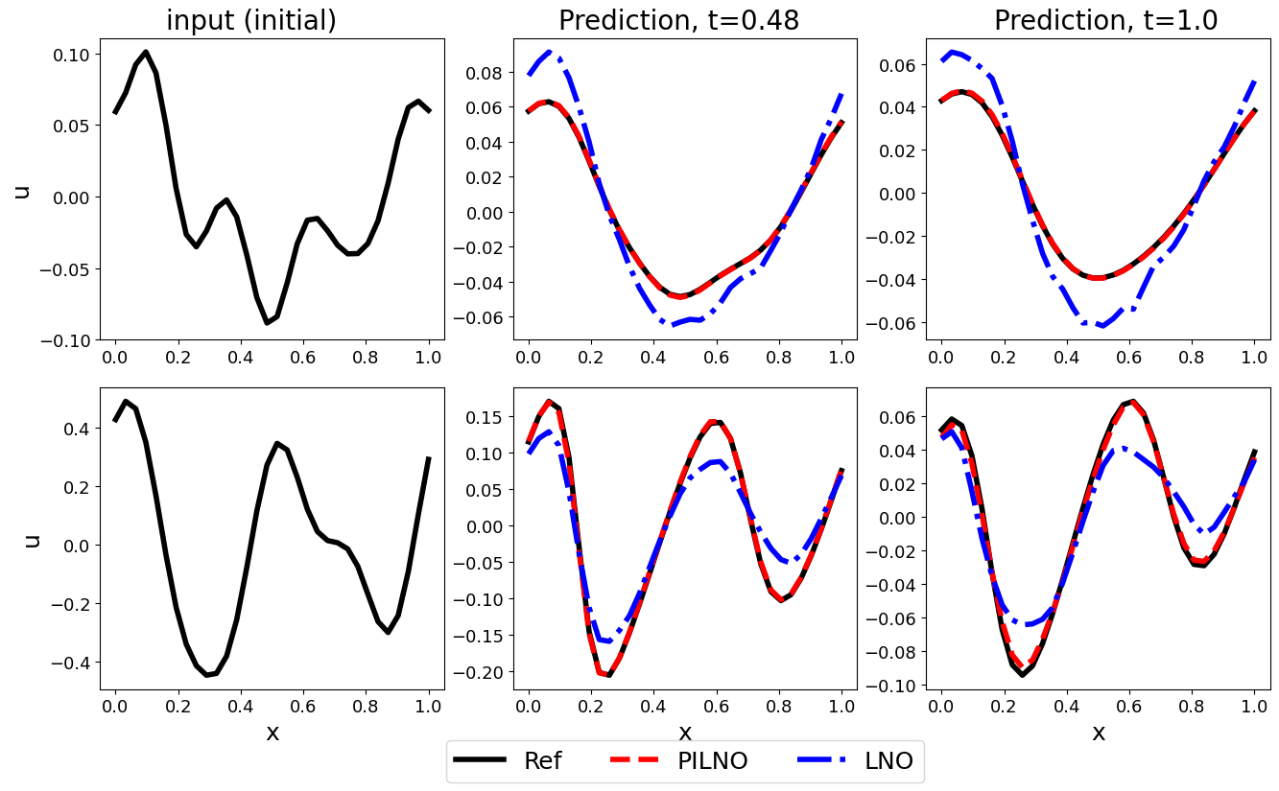}
        \caption{Worst–case time–slicing comparisons on the test set for the two LNO cases with the largest relative $L_2$ error. PILNO approximates the reference solution more accurately than LNO.} 
        \label{fig:burgers_comparison_pilno_alno}
    \end{figure}

    To further illustrate the qualitative behavior, Figure~\ref{fig:burgers_comparison_pilno_alno} shows time-slice comparisons for the two test cases on which LNO incurs its largest relative $L_2$ errors. For these same initial conditions, PILNO produces markedly more accurate predictions, tracking the reference solution more faithfully. This suggests that the physics-driven training objective in PILNO effectively constrains the governing physics and helps preserve key spatio-temporal features even under limited supervision.
    
\begin{figure}[H]
        \centering
        \begin{subfigure}[t]{0.49\linewidth}
            \centering \includegraphics[width=\linewidth]{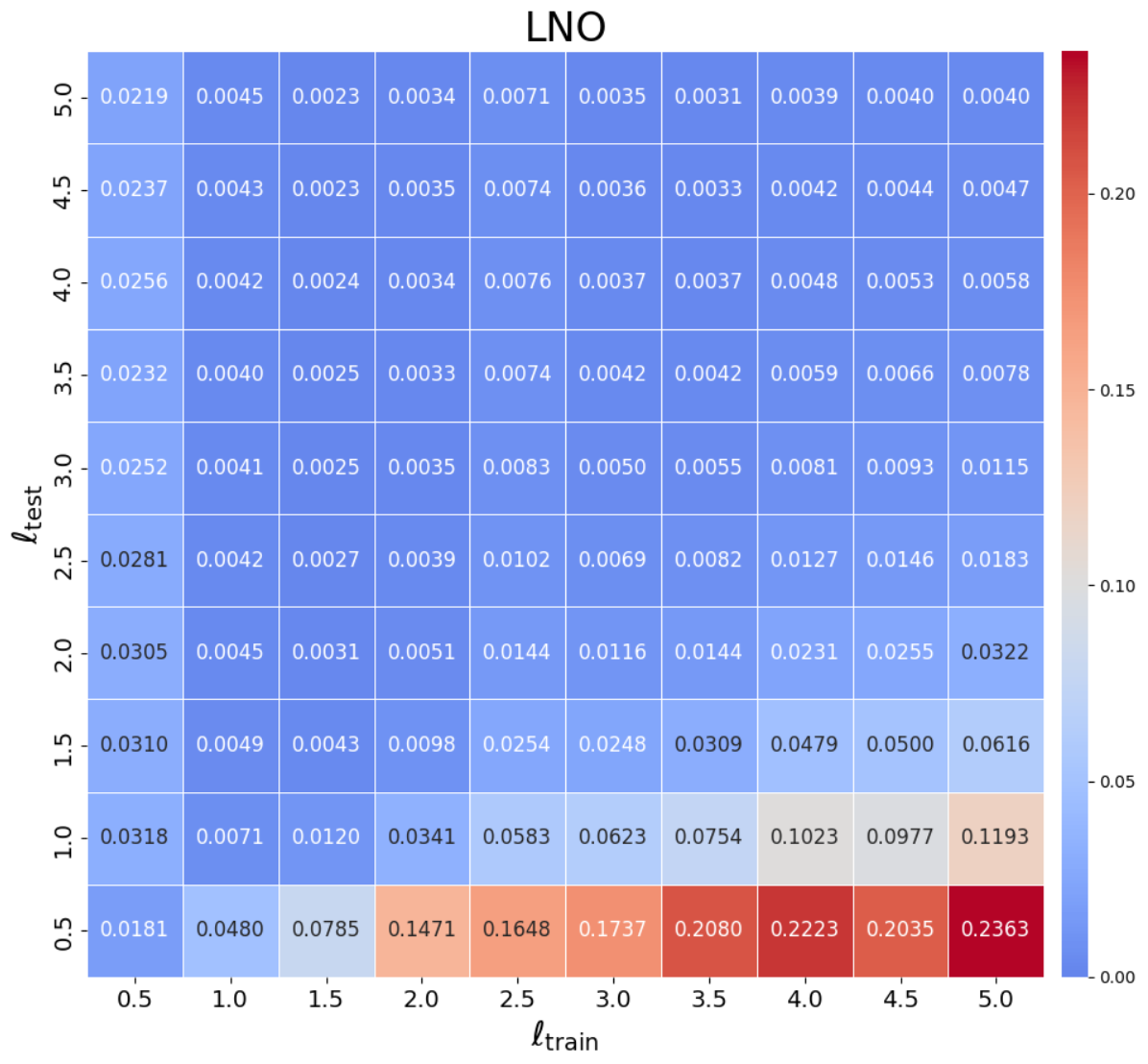}
            \caption{}
        \end{subfigure}
        \begin{subfigure}[t]{0.49\linewidth}
            \centering \includegraphics[width=\linewidth]{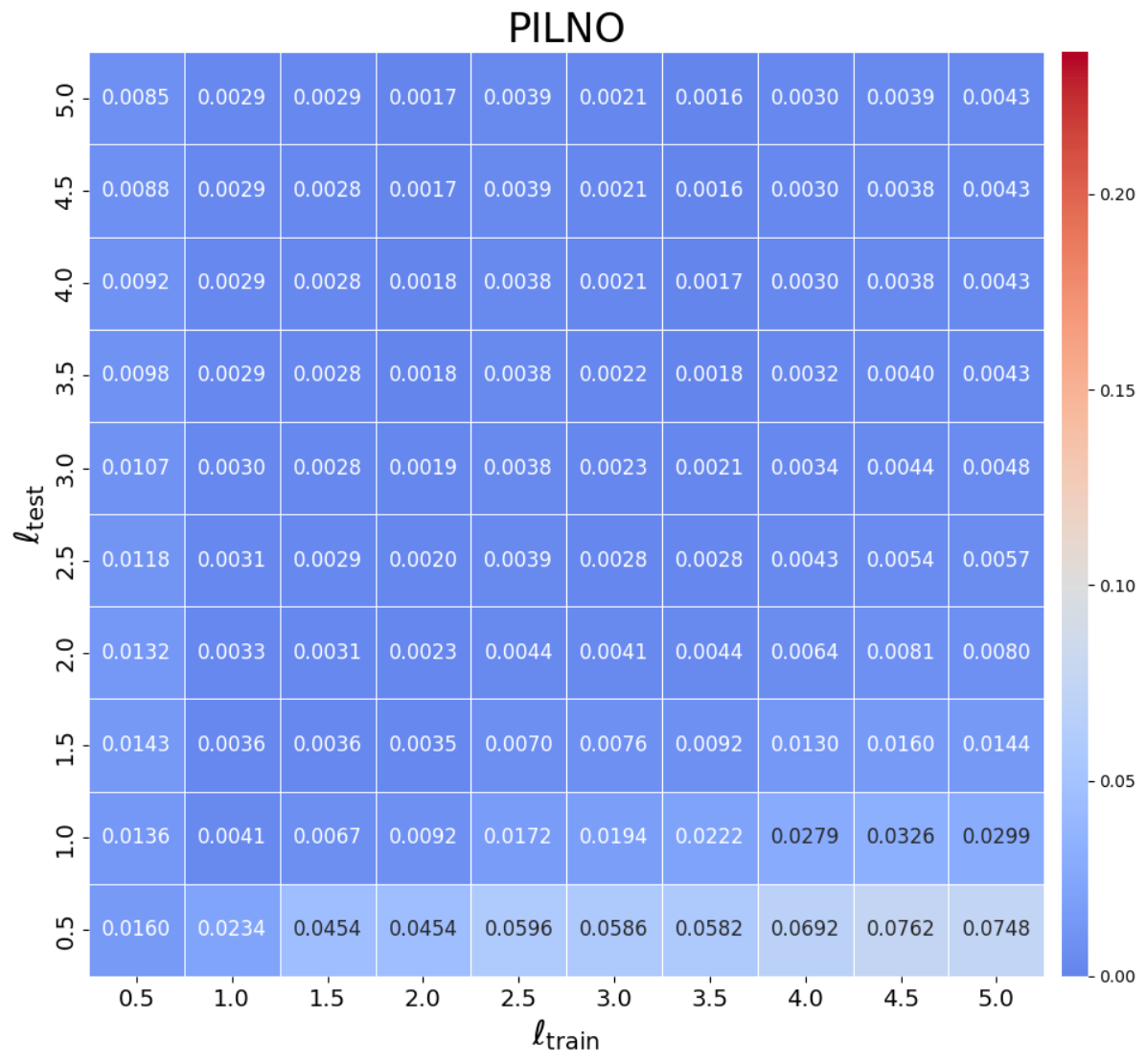}
            \caption{}
        \end{subfigure}
        \caption{Generalization across initial–condition length–scales. Heatmaps show relative $L^2$ error when training at length–scale $\ell_{\text{train}}$ (horizontal axis) and testing at $\ell_{\text{test}}$ (vertical axis), for (a) LNO and (b) PILNO. PILNO maintains low errors even when extrapolating from smooth ($\ell=5.0$) to oscillatory ($\ell=0.5$) regimes. PILNO consistently demonstrates better generalization, showing robust predictive performance even on data distributions not included during training.}
        \label{fig:burgers_generalization}
    \end{figure}

\paragraph{OOD generalization performance}
    We next evaluate OOD generalization with respect to the correlation length-scale of the initial condition. We construct ten independent datasets by sampling $u_0(x)$ from the same Gaussian random field at length-scales $\ell \in \{0.5, 1.0, \dots, 5.0\}$. For each $\ell_{\text{train}}$, we train LNO and PILNO using $N_{\text{train}} = 200$ paired samples (and $N_{\text{virt}}=1000$ virtual inputs for PILNO), then evaluate on test sets drawn from all ten length-scales $\ell_{\text{test}}$. Each $(\ell_{\text{train}},\ell_{\text{test}})$ pair defines one cell in a $10\times10$ error heatmap, shown separately for LNO and PILNO in Figure~\ref{fig:burgers_generalization}.

    When trained on smooth initial conditions ($\ell_{\text{train}} = 5.0$) and tested on highly oscillatory fields ($\ell_{\text{test}}=0.5$), LNO suffers from substantial degradation, with relative $L_2$ error around $23.6\%$. In contrast, PILNO, despite being trained only on smooth data, maintains significantly lower error in this challenging extrapolation regime (about $7\%$), and exhibits uniformly improved performance off the diagonal. These results indicate that embedding the governing PDE into the operator-learning objective, together with virtual inputs that broaden the input spectrum, leads to robust OOD generalization across a wide range of initial-condition regularities.

    Overall, the Burgers’ equation experiments show that PILNO achieves a more favorable trade-off between accuracy and data efficiency than the purely data-driven LNO. Physics residuals, combined with virtual inputs, yield more stable training and lower error both in the extreme small-data regime and under pronounced mismatches between training and test initial-condition statistics.

\subsection{Darcy flow}\label{subsec:darcy}
    We next consider the two-dimensional steady-state Darcy flow on the unit square $\Omega=(0,1)^2$, a second-order linear elliptic PDE with Dirichlet boundary conditions. This example evaluates PILNO in the small-data regime and highlights the role of virtual inputs for improving performance when only a limited number of labeled samples is available. The governing equation is
    
    \begin{equation}\label{eqn:darcy}
        -\nabla\!\cdot\!\big(a(x)\,\nabla u(x)\big) \;=\; f(x)\quad \text{in }\Omega=(0,1)^2, \qquad
        u(x)\;=\;g(x)\quad \text{on }\partial\Omega,
    \end{equation}
    where $u:\Omega\to\mathbb{R}$ denotes the pressure field, $a\in L^{\infty}(\Omega)$ is the piecewise–constant permeability coefficient, and $f\equiv 1$ is a fixed forcing (following~\cite{li2020fourier,li2024physics}). Unless otherwise stated, we take homogeneous Dirichlet boundary condition $g\equiv 0$. The learning task is to approximate the solution operator from the permeability field to the corresponding pressure solution,
    \begin{equation}\label{eqn:darcy_operator}
    \mathcal{G}_{\Theta}:\ a(x)\ \longmapsto\ u(x),
    \end{equation}
    using the PILNO training strategy from Section~\ref{sec:pilno} (data loss when available, plus PDE/BC residuals).

\paragraph{Random permeability field and data generation}
    We follow the data-generation procedure of PINO~\cite{li2020fourier, li2024physics}. First, we sample a mean-zero Gaussian random field $\phi:D\to\mathbb{R}$ on the periodic domain $D=[0,1]\times[0,1]$ with covariance operator
    \[
    \phi(\cdot) \sim \mathcal{N}\bigl(0,(-\Delta+9I)^{-2}\bigr).
    \]
    The permeability field $a:D\to\{3,12\}$ is obtained by a sign-thresholding map
    \[
        a(x) = \psi(\phi(x)), \qquad \psi(z) =  
        \begin{cases}
        12, & z \ge 0,\\
        3,  & z < 0.
        \end{cases}
    \]
    so that each realization consists of high- and low-permeability regions.

    For each sampled permeability $a$, we solve~\eqref{eqn:darcy} on a fine $241\times 241$ grid to obtain an accurate reference solution $u$. These fine-grid solutions are used as ground truth for evaluation and to construct coarse training labels. To mimic a realistic low-resolution data setting, we construct the supervised training pairs $(a,u)$ on a coarse $11\times 11$ grid by downsampling the reference solutions. To inject higher-resolution physics information without requiring fine-grid labels, we additionally evaluate the PDE residual $\mathcal{L}_{\text{pde}}$ on an intermediate $61\times 61$ collocation grid, while the supervised data loss remains defined only on the $11\times 11$ grid. Thus, training combines coarse-grid data supervision with a finer-grid physics loss. 

    Zero boundary conditions are enforced for all methods by multiplying the mollifier
    \[
        m(x,y) = \sin(\pi x)\,\sin(\pi y),
    \]
    so that the network outputs satisfy $u=0$ on $\partial\Omega$ by construction. Unless otherwise noted, we report the mean and standard deviation of the relative $L_2$ error over $100$ test instances, evaluated at both the coarse ($11\times 11$) and fine ($61\times 61$) resolutions. The primary metric is the relative $L_2$ error of the pressure field.    

    \begin{table}[ht]
        \centering
        \renewcommand{\arraystretch}{1.4}
        \begin{tabular}{c c c c c}
            \hline
            & FNO & LNO & PIFNO & PILNO \\
            \hline
            \makecell[l]{Relative \(L_2\) Error at\\ low resolution ($11\times11$)}
            & 5.41e-2 & 1.29e-2 & 5.23e-2 & 1.19e-2 \\
            \hline
            \makecell[l]{Relative \(L_2\) Error at\\ high resolution ($61\times61$)}
            & 9.01e-2 & 3.73e-2 & 1.56e-2 & 1.21e-2 \\
            \hline
        \end{tabular}
        \caption{Relative $L_2$ errors on Darcy flow. All models are trained on $N_\text{train}=800$ coarse $11\times 11$ samples. PIFNO and PILNO additionally enforce the PDE residual on a finer $61\times 61$ collocation grid, which substantially improves test accuracy compared with data-driven baselines trained only on coarse labels.}
        \label{tab:darcy_error}
    \end{table}
    Table~\ref{tab:darcy_error} summarizes the performance of FNO, LNO, PIFNO, and PILNO when trained with $N_{\text{train}}=800$ paired samples. PIFNO and PILNO incorporate the same coarse labels as FNO and LNO, but also exploit a higher-resolution PDE residual evaluated on the $61\times 61$ collocation grid. As a result, their errors remain low even when evaluated on the finer grid (1.56\% and 1.21\%, respectively), whereas the purely data-driven models suffer noticeable degradation when moving from $11\times 11$ to $61\times 61$.

\begin{figure}[ht]
        \centering
        \includegraphics[height=9cm]{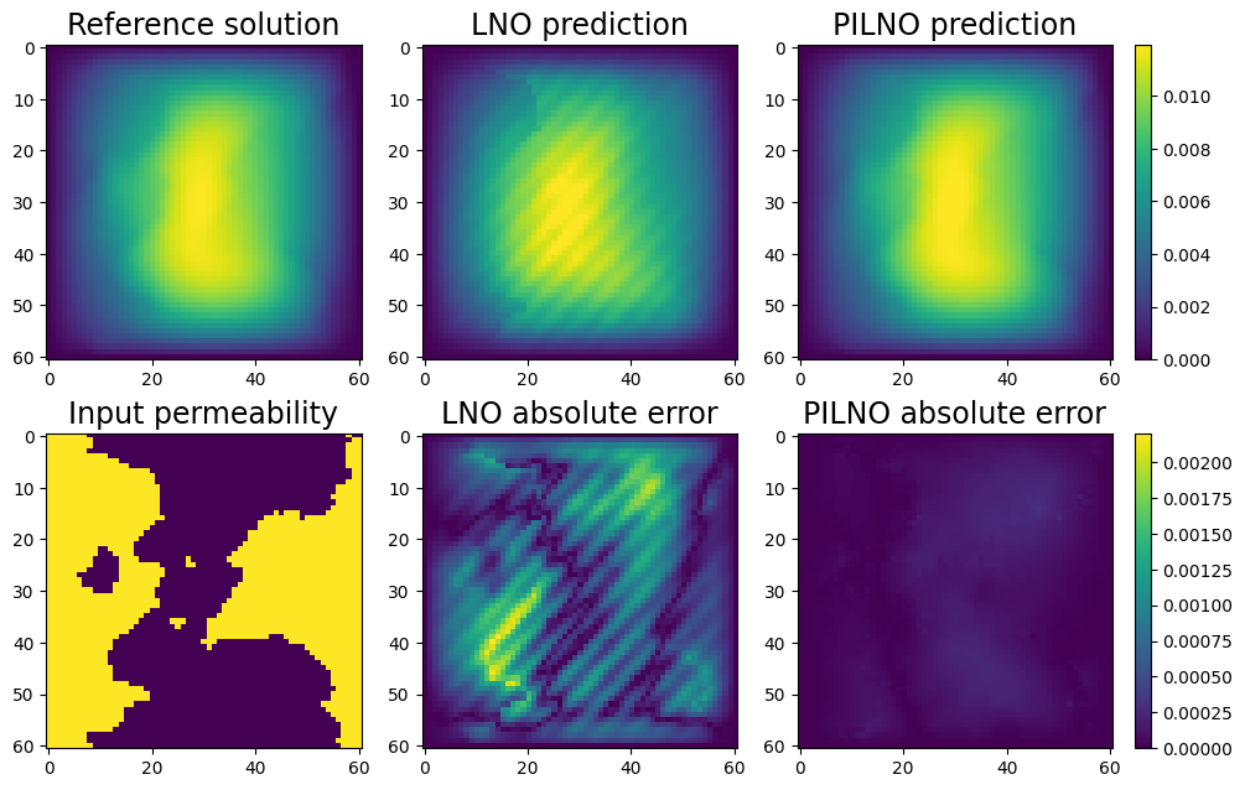}
        \caption{Comparison at $N_\text{train}=10$ (Darcy flow) on $61\times61$ resolution. LNO attains a relative $L_2$ error of 14.67\%, whereas PILNO achieves 1.23\%, indicating substantially better performance by PILNO under extremely limited training data.}
        \label{fig:darcy_plot}
    \end{figure}

\begin{figure}[ht]
        \centering
        \begin{subfigure}{0.49\linewidth}
            \centering \includegraphics[height=5.7cm]{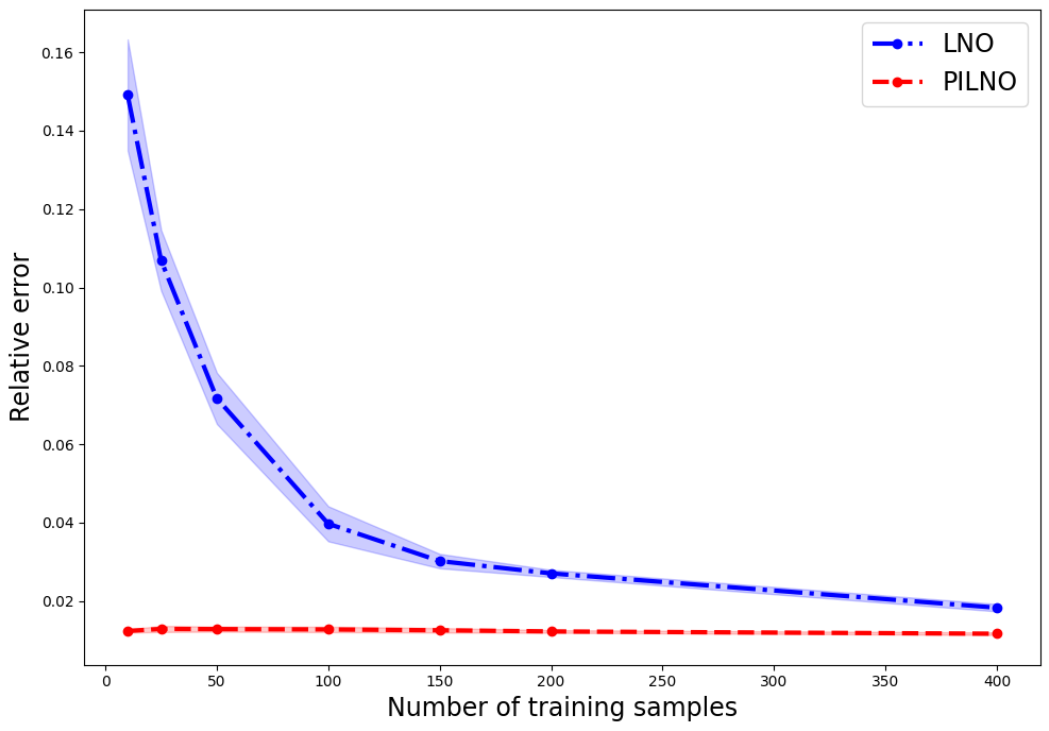}
            \caption{}
        \end{subfigure}
        \centering
        \begin{subfigure}{0.49\linewidth}
            \centering \includegraphics[height=5.7cm]{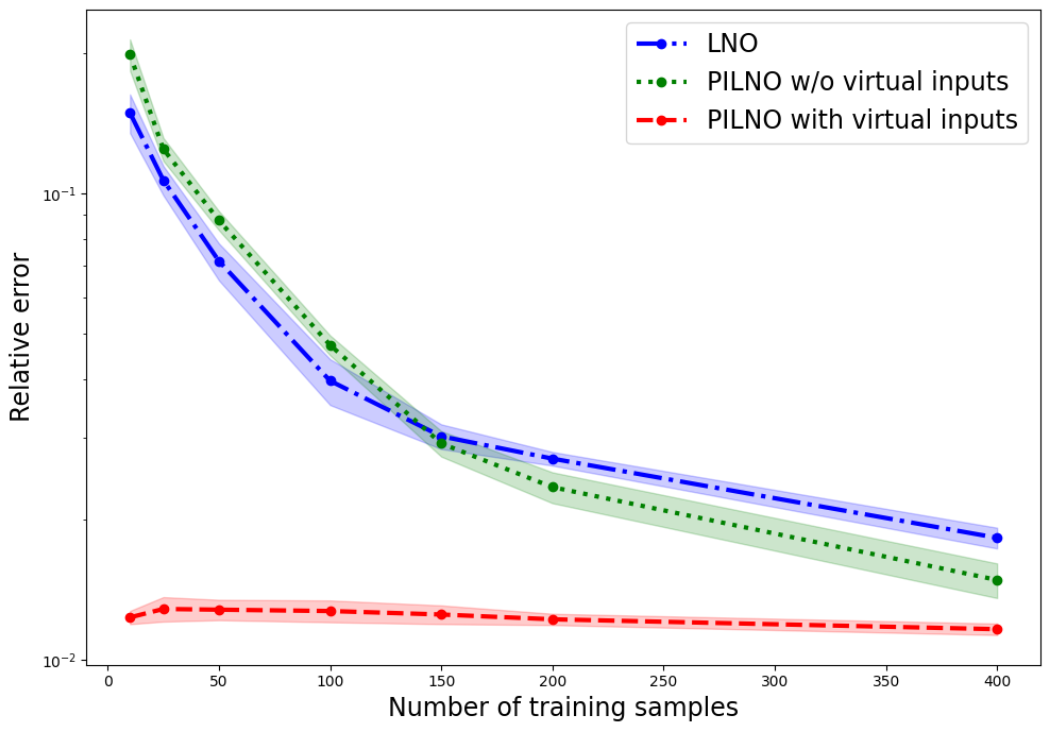}
            \caption{}
        \end{subfigure}
        \caption{Error plots on number of training samples and the role of virtual inputs for Darcy flow. (a) Relative $L_2$ test error versus the number of training samples for LNO and PILNO. Points denote means and shaded bands indicate $\pm 1$ standard deviation over five random seeds. As $N_\text{train}$ decreases, the error of LNO rises sharply (reaching $\approx 14.9\%$ at $N_\text{train}=10$), whereas PILNO maintains low error levels (around $1.2\%$ across all $N_\text{train}$). (b) Effect of removing virtual inputs. When physics-informed models are trained without virtual inputs (i.e., physics residuals applied only to paired data), PILNO outperforms purely data-driven baselines only for larger training sets ($N_\text{train}\gtrsim 200$). For smaller training sets ($N_\text{train}\lesssim 200$), this advantage diminishes or even disappears, yielding errors comparable to data-driven methods and underscoring the importance of virtual inputs in the small-data regime.}
        \label{fig:darcy_ntrain_pilno}
    \end{figure}

    Figure~\ref{fig:darcy_ntrain_pilno}-(a) shows the relative $L_2$ test error as a function of the number of training samples. Each model is trained five times with different random seeds; points denote means and shaded bands indicate $\pm1$ standard deviation. As the number of paired samples decreases, the error of LNO increases sharply (reaching $\approx 14.9\%$ at $N_\text{train}=10$). In contrast, PILNO maintains low error across all $N_\text{train}$, remaining close to its $N_\text{train}=400$ performance (about $1.2\%$). Figure~\ref{fig:darcy_plot} further compares LNO and PILNO for the extreme small-data case $N_\text{train}=10$. Because zero boundary conditions are enforced as a hard constraint via the mollifier, both methods respect the boundary shape, but LNO exhibits large interior discrepancies (relative $L_2 \approx 14.9\%$), whereas PILNO closely matches the reference solution with a relative error of $\approx 1.2\%$.

\paragraph{Importance of virtual inputs}
    Figure~\ref{fig:darcy_ntrain_pilno}-(b) examines the effect of virtual inputs. As discussed in Section~\ref{subsec:virtual_dataset}, enforcing PDE residuals can improve generalization by injecting the governing physics into the model. However, when the physics residuals are applied only to the limited set of paired samples, supervision is confined to a narrow input dataset; in the small-data regime, this can exacerbate overfitting and degrade test accuracy. Consistent with this, physics-informed models without virtual inputs outperform purely data-driven baselines when $N_\text{train}\gtrsim 200$, but their advantage diminishes and can even vanish as $N_\text{train}$ becomes very small. These results highlight that virtual inputs are crucial for reliable physics-informed operator learning under limited data.

\subsection{Reaction-Diffusion equation}\label{subsec:reaction_diffusion}
    We consider a one-dimensional reaction-diffusion equation with a quadratic reaction term
    \begin{equation}\label{eqn:reaction_diffusion}
        D\frac{\partial^2 u}{\partial x^2} + k\,u^2 - \frac{\partial u}{\partial t} \;=\; f(x),
        \qquad (x,t)\in[0,1]\times[0,1],
    \end{equation}
    where $u(t,x)$ is a scalar field, $D>0$ is the diffusivity, $k\ge 0$ the reaction strength, and $f(x)$ a time-independent forcing. Following prior operator-learning studies (e.g., LNO~\cite{cao2023lno} and PI-DeepONet~\cite{wang2021learning}), we set $D=0.01$ and $k=0.01$. Depending on the task, we either prescribe a nontrivial initial condition $u_0(x)$ with zero forcing, or a nontrivial forcing $f(x)$ with zero initial and boundary conditions.

    We study two operator mappings under a common experimental procedure:
    \begin{itemize}
        \item \textbf{Task A (IC $\rightarrow$ solution, zero forcing):} learns the mapping $u_0(x) \mapsto u(x,t)$ with $f\equiv 0$, used to evaluate out-of-distribution (OOD) generalization.
        \item \textbf{Task B (forcing $\rightarrow$ solution, zero IC/BC):} learns the mapping $f(x) \mapsto u(x,t)$ with zero initial and boundary conditions. This setting is more challenging and is used to probe both small-data performance and the impact of temporal-causality weighting.
    \end{itemize}

    \paragraph{Common discretization and data generation}
    Reference solutions are generated using the same high-fidelity solver as in the Burgers' experiments: spatial derivatives are approximated by a fourth-order central finite difference scheme on a fine uniform grid with $N_x=151$ points, and the semi-discrete system is advanced in time using the classical explicit fourth-order Runge-Kutta method with time step $\Delta t=$ 1e-4. The resulting solutions are downsampled to a uniform grid with $N_x = 51$ spatial points and $N_t = 51$ time snapshots on $[0,1]\times[0,1]$, so that each solution is represented on a $51\times 51$ space-time grid. In both tasks, the input to LNO/PILNO is a one-dimensional function ($u_0$ for Task A, $f$ for Task B), and the target is the corresponding solution $u(x,t)$ on this grid.

    For both tasks, input functions are sampled from the same Gaussian process family used in the Burgers' example, namely a zero-mean Gaussian random field with exponentiated squared sine kernel. For each correlation length-scale $\ell>0$,
    \[
        u_0(\cdot) \sim \mathcal{GP}\bigl(0,k^{(\ell)}_{\text{exp-sin}}\bigr),
        \qquad
        f(\cdot) \sim \mathcal{GP}\bigl(0,k^{(\ell)}_{\text{exp-sin}}\bigr),
    \]
    with covariance
    \[
        k^{(\ell)}_{\text{exp-sin}}(x,x')
        = \sigma^2 \exp\!\left(
            -\frac{2\sin^2\big(\pi(x-x')/p\big)}{\ell^2}
          \right),
    \]
    where we fix $p=1$ and $\sigma=0.2$. For Task A, this kernel is used to sample initial conditions $u_0$ with $f\equiv 0$; for Task B, it is used to sample forcings $f$ with $u_0\equiv 0$ and homogeneous Dirichlet boundary conditions enforced via a mollifier.

    Similar to the Burgers' experiments, we consider two complementary evaluation regimes. For data-efficiency studies, we fix a representative length-scale (here $\ell=0.5$) and vary only the number of paired training samples $N_{\text{train}}$, while keeping the test distribution and test set size fixed. For OOD generalization, we construct ten datasets, one for each length-scale $\ell\in\{0.5,1.0,\dots,5.0\}$, and for a given $\ell_{\text{train}}$ we train a model on $N_{\text{train}}$ samples drawn at that length-scale and evaluate on test sets drawn at all $\ell_{\text{test}}$. Each $(\ell_{\text{train}},\ell_{\text{test}})$ pair thus defines a cell in a $10\times 10$ error matrix, reported separately for Tasks A and B. Unless otherwise stated, we use $N_{\text{virt}}=500$ virtual inputs for PILNO.
    
\begin{figure}[H]
        \centering
        \begin{subfigure}[t]{0.49\linewidth}
            \centering \includegraphics[width=\linewidth]{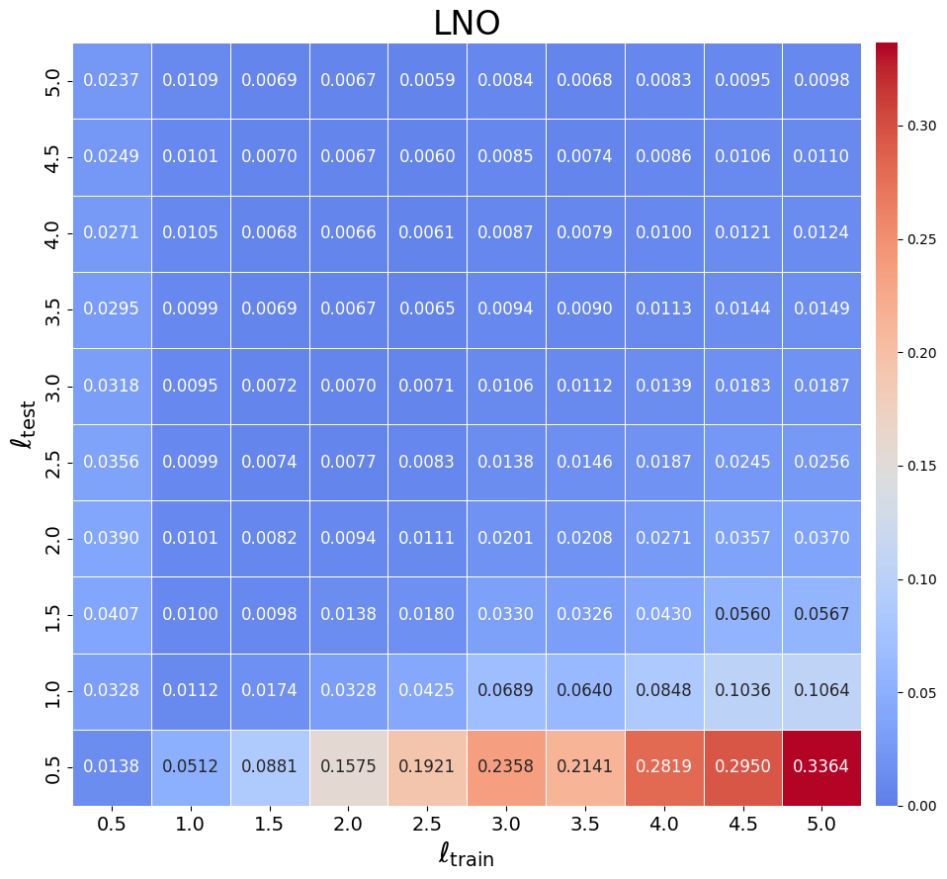}
            \caption{}
        \end{subfigure}
        \begin{subfigure}[t]{0.49\linewidth}
            \centering \includegraphics[width=\linewidth]{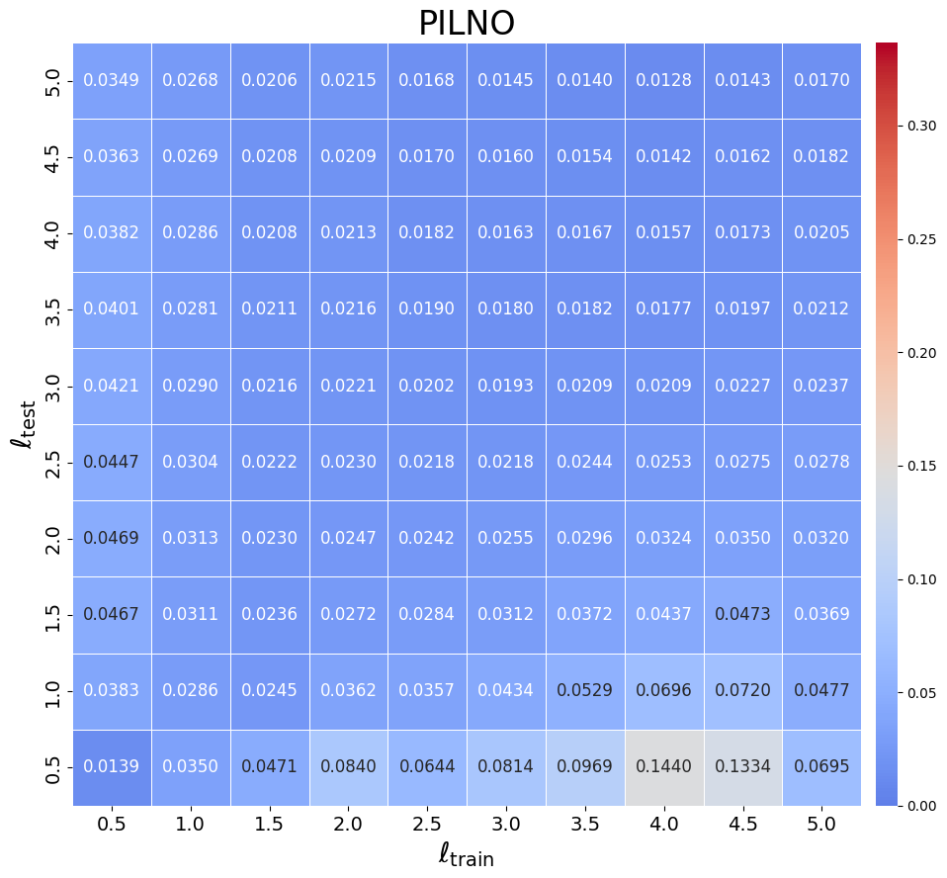}
            \caption{}
        \end{subfigure}
        \caption{Generalization across initial–condition length–scales for reaction-diffusion equation. Heatmaps show relative $L^2$ error when training at length–scale $\ell_{\text{train}}$ (horizontal axis) and testing at $\ell_{\text{test}}$ (vertical axis), for (a) LNO and (b) PILNO. PILNO sustains lower errors off the diagonal (mismatched smooth/oscillatory regimes), demonstrating stronger out–of–distribution generalization.}
        \label{fig:reaction_diffusion_generalization_IC}
    \end{figure}

\paragraph{Task A (IC $\rightarrow$ Solution; zero forcing)}
    For Task A we set $f\equiv 0$, so that~\eqref{eqn:reaction_diffusion} reduces to
    \[
        D\frac{\partial^2 u}{\partial x^2} + k\,u^2 - \frac{\partial u}{\partial t} = 0,
        \qquad (x,t)\in[0,1]\times[0,1],
    \]
    with periodic boundary conditions $u(0,t)=u(1,t)$. The operator to be learned is
    \[
        \mathcal{G}^A_\Theta : u_0(x) \mapsto u(x,t).
    \]
    To assess OOD performance under limited paired data, we generate datasets at ten correlation length-scales $\ell\in\{0.5,1.0,\dots,5.0\}$. For each $\ell_{\text{train}}$, we train a model using $N_{\text{train}}=25$ paired samples and evaluate on $N_{\text{test}}=100$ test instances at all $\ell_{\text{test}}$. Each $(\ell_{\text{train}},\ell_{\text{test}})$ pair defines one cell in the $10\times 10$ heatmaps in Fig.~\ref{fig:reaction_diffusion_generalization_IC}.

    Figure~\ref{fig:reaction_diffusion_generalization_IC} compares LNO and PILNO on this length-scale grid. For example, when trained on smooth initial conditions ($\ell_{\text{train}}=5.0$) and tested on highly oscillatory ones ($\ell_{\text{test}}=0.5$), LNO attains a relative $L_2$ error of about $33.6\%$, whereas PILNO reduces the error to approximately $6.9\%$. This demonstrates that PILNO provides substantially more robust OOD generalization than LNO under the same limited amount of paired data.
    
\paragraph{Task B (Forcing $\rightarrow$ Solution; zero IC/BC)}
    For Task B, we impose zero initial and boundary conditions and learn the mapping from a spatial forcing profile to the full spatio-temporal solution, as in PI-DeepONet~\cite{wang2021learning}:
    \[
        \mathcal{G}^B_\Theta : f(x) \mapsto u(x,t).
    \]
    We use the same $51\times 51$ space--time grid as in Task A. The inputs are forcings $f(x)$ sampled from the GP family above, and the output is the corresponding solution $u(x,t)$. Zero initial and boundary conditions are enforced via the mollifier
    \[
        m(x,t) = \sin(\pi x)\sin(0.5\pi t),
    \]
    analogously to the Darcy-flow experiment.
    
    \begin{figure}[H]
        \centering
        \includegraphics[height=8cm]{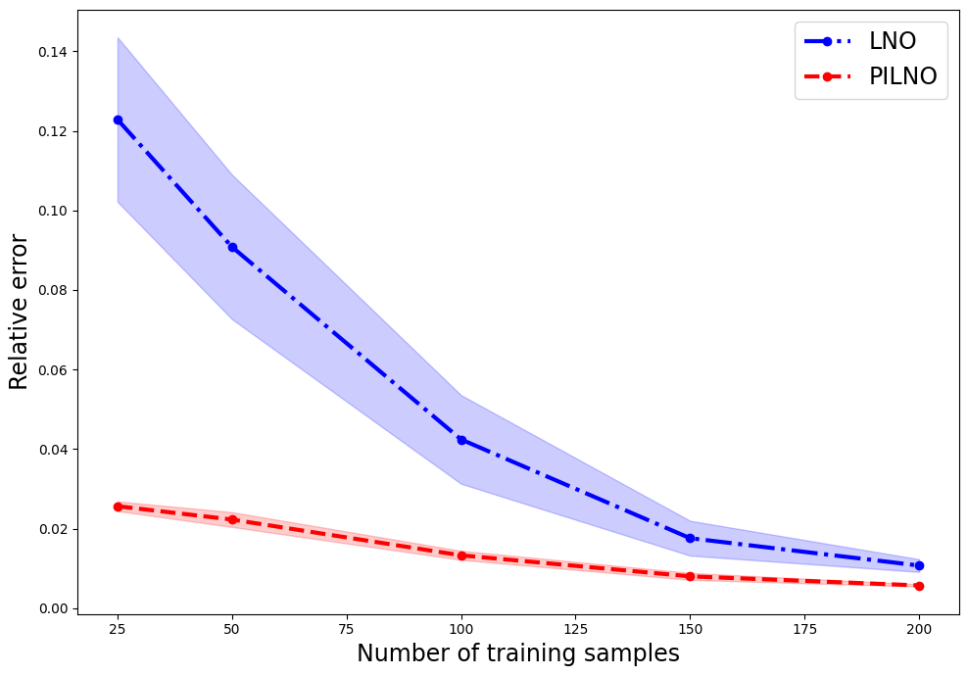}
        \caption{Relative $L_2$ error on the test set versus the number of training samples (mean $\pm$ std over 5 random seeds). As $N_{\mathrm{train}}$ decreases, LNO’s error rises sharply, whereas PILNO degrades more gracefully (e.g., $\approx\!2.5\%$ at $N_{\mathrm{train}}=25$).}
        \label{fig:reaction_diffusion_pilno_ntrain}
    \end{figure}
    
    \begin{figure}[H]
        \centering
        \includegraphics[height=9cm]{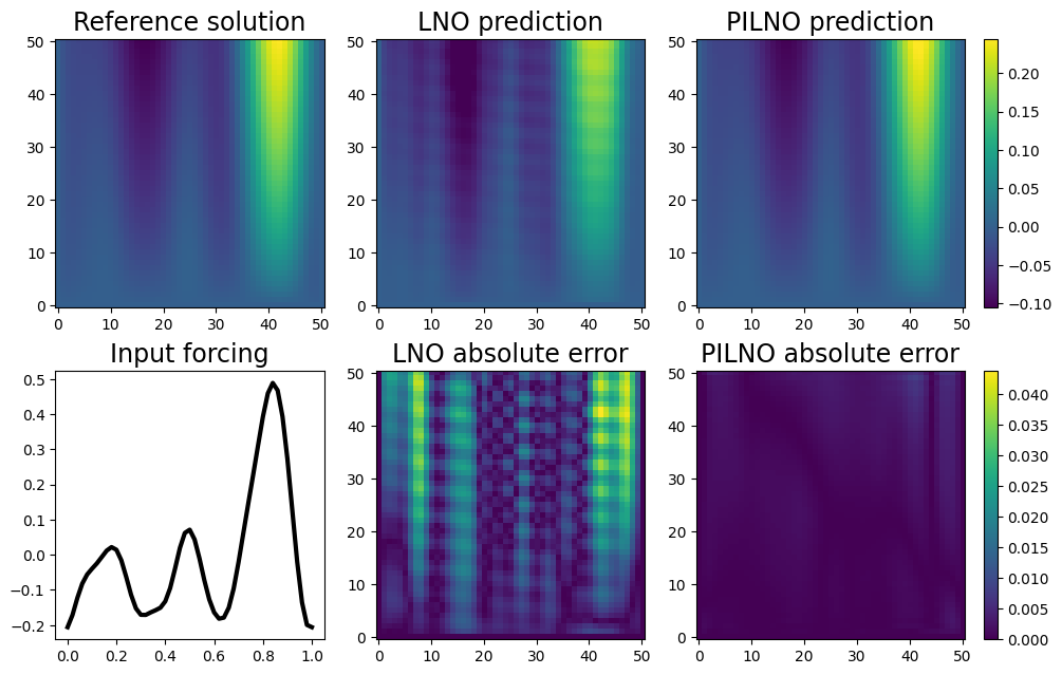}
        \caption{Comparison at $N_\text{train}=25$ under $\ell_\text{train}=\ell_\text{test}=0.5$ (reaction-diffusion). LNO attains a relative $L_2$ error of 12.3\%, whereas PILNO achieves 2.5\%, indicating substantially better performance by PILNO under extremely limited training data.}
        \label{fig:rd1d_plot}
    \end{figure}
    To study data efficiency in an in-distribution setting, we fix $\ell_{\text{train}}=\ell_{\text{test}}=0.5$ and vary $N_{\text{train}}\in\{25,50,100,150,200\}$, reporting mean $\pm$ standard deviation over five random seeds. Figure~\ref{fig:reaction_diffusion_pilno_ntrain} shows the resulting test errors: LNO's error increases sharply as $N_{\text{train}}$ decreases (reaching roughly $12.3\%$ at $N_{\text{train}}=25$), whereas PILNO degrades much more gracefully and remains near $2.5\%$ even at $N_{\text{train}}=25$. Figure~\ref{fig:rd1d_plot} illustrates the extreme small-data case $N_{\text{train}}=25$ at $\ell=0.5$: although both methods satisfy the enforced zero IC/BC by the mollifier, LNO exhibits substantial interior discrepancies (relative $L_2\approx 12.3\%$), whereas PILNO, aided by physics residuals, virtual inputs, and temporal-causality weighting, closely matches the reference solution (relative $L_2\approx 2.5\%$).
    
    \begin{figure}[H]
        \centering
        \begin{subfigure}[t]{0.49\linewidth}
            \centering \includegraphics[width=\linewidth]{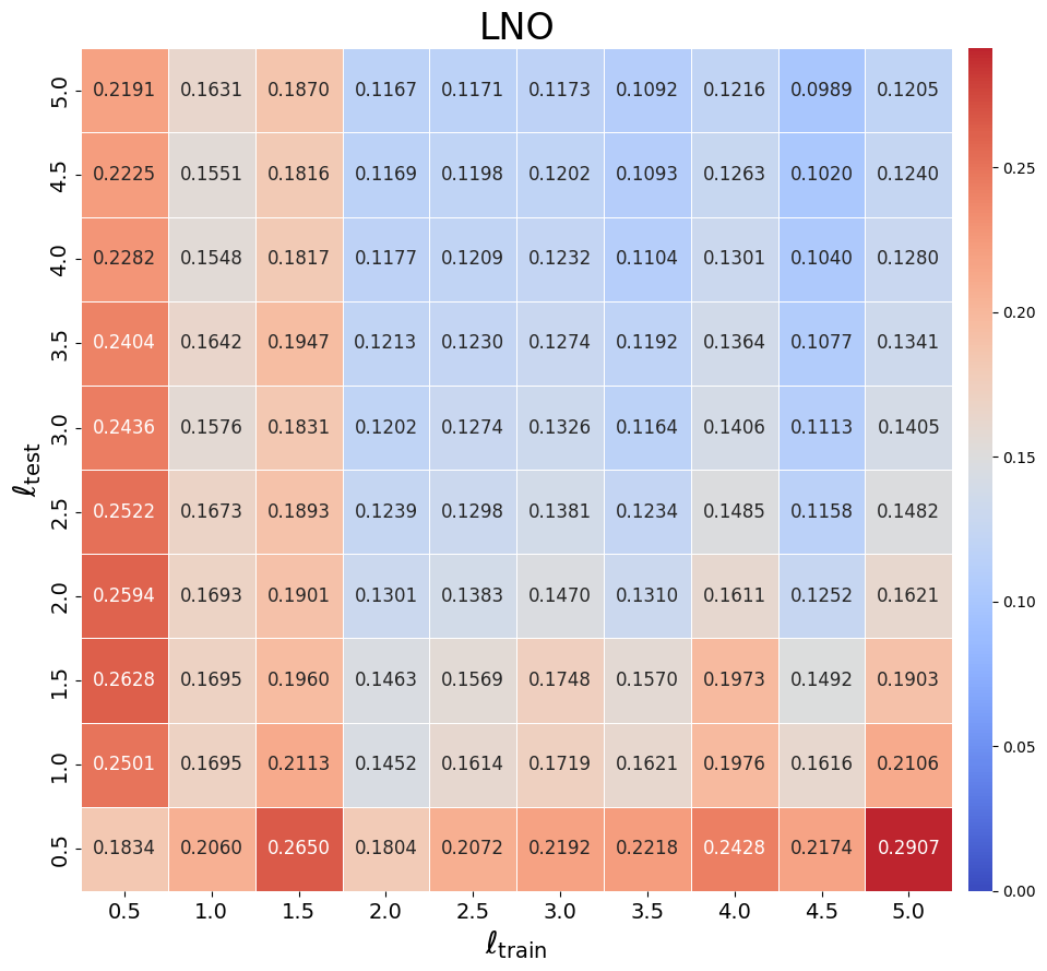}
            \caption{}
        \end{subfigure}
        \begin{subfigure}[t]{0.49\linewidth}
            \centering \includegraphics[width=\linewidth]{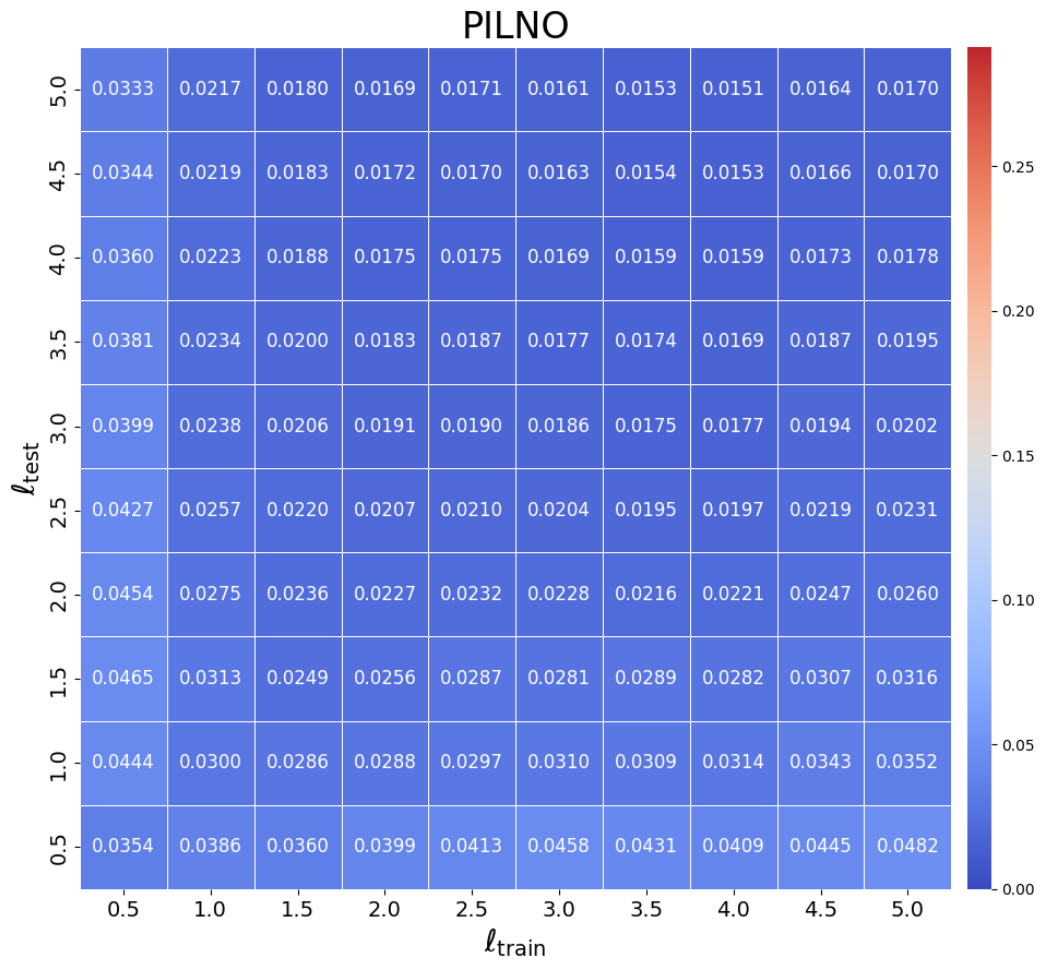}
            \caption{}
        \end{subfigure}
        \caption{Generalization across forcing term length–scales for reaction-diffusion equation. Heatmaps show relative $L^2$ error when training at length–scale $\ell_{\text{train}}$ (horizontal axis) and testing at $\ell_{\text{test}}$ (vertical axis), for (a) LNO and (b) PILNO. PILNO sustains lower errors off the diagonal (mismatched smooth/oscillatory regimes), demonstrating stronger out–of–distribution generalization.}
        \label{fig:reaction_diffusion_generalization_forcing}
    \end{figure}
    
    \begin{figure}[ht]
        \centering
        \includegraphics[height=9cm]{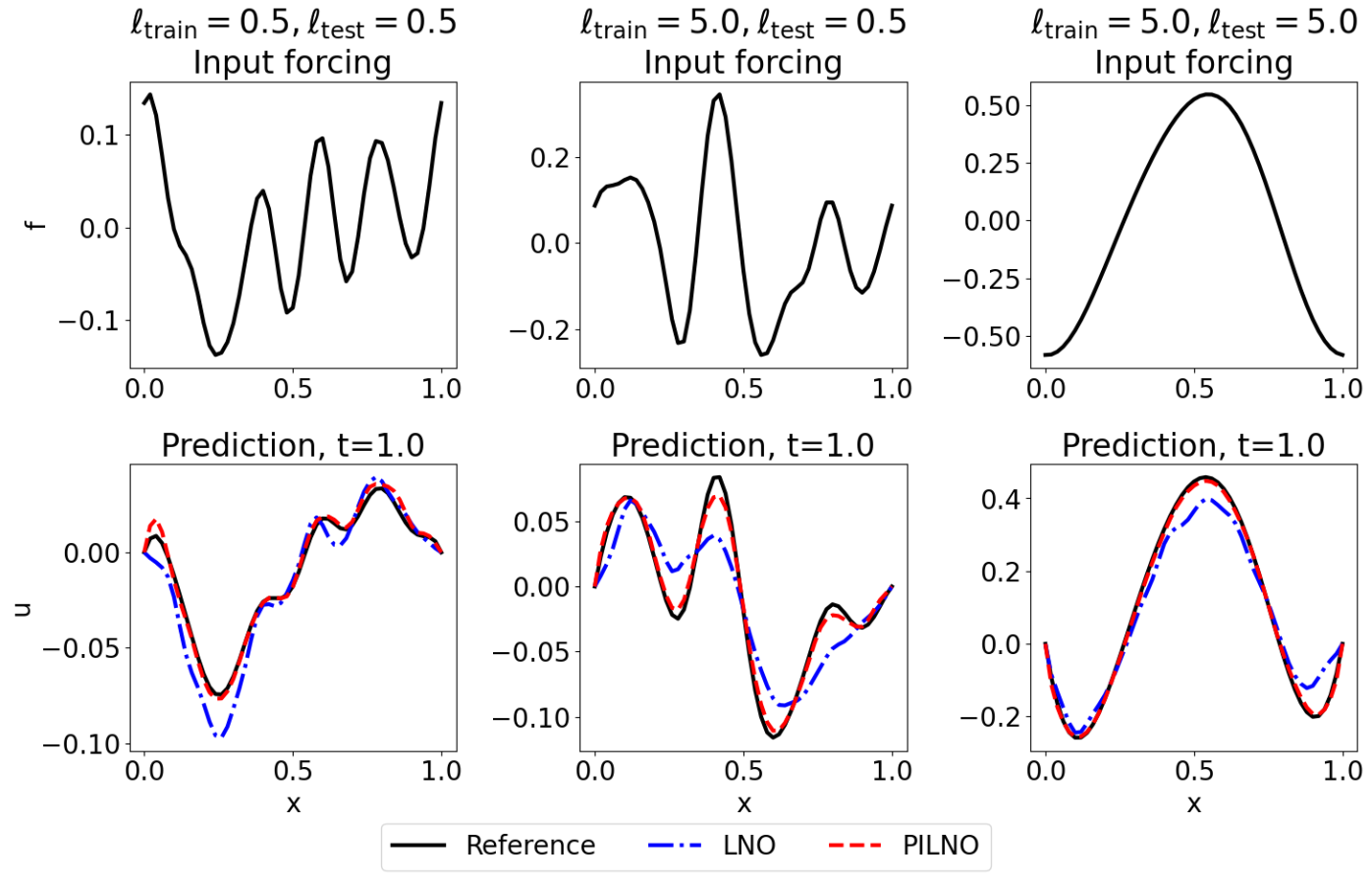}
        \caption{Comparison of PILNO and LNO on the reaction-diffusion equation under three train-test length scale configurations. Left: $\ell_\text{train}=0.5, \ell_\text{test}=0.5$. Middle: $\ell_\text{train}=5.0, \ell_\text{test}=0.5$. Right: $\ell_\text{train}=5.0, \ell_\text{test}=5.0$. Across all cases, PILNO produces predictions that more closely match the reference solution than LNO, as evidenced by the visual agreement of the predicted profiles with the reference.}
        \label{fig:rd1d_heatmap_plot}
    \end{figure}

    For OOD generalization, we generate datasets at the ten correlation length-scales $\ell\in\{0.5,1.0,\dots,5.0\}$, train a separate model with $N_{\text{train}}=25$ at each $\ell_{\text{train}}$, and evaluate on $N_{\text{test}}=100$ test instances at all $\ell_{\text{test}}$. The resulting $10\times 10$ error matrices are shown in Fig.~\ref{fig:reaction_diffusion_generalization_forcing}. In this small-data regime, the forcing-to-solution mapping is substantially more challenging than the IC-to-solution case: for Task A, diagonal entries ($\ell_{\text{train}}=\ell_{\text{test}}$) yield errors on the order of $1\%$, whereas for Task B the corresponding LNO errors exceed $10\%$. Nevertheless, PILNO attains uniformly lower errors across the grid, maintaining small diagonal errors and remaining robust when extrapolating from smooth forcings (large $\ell_\text{train}$) to highly oscillatory forcings (small $\ell_\text{test}$). Representative line plots for selected train-test pairs are given in Fig.~\ref{fig:rd1d_heatmap_plot}, where PILNO consistently tracks the reference solution more closely than LNO.

\paragraph{Importance of temporal-causality weighting}
    \begin{figure}[h]
        \centering
        \includegraphics[height=9cm]{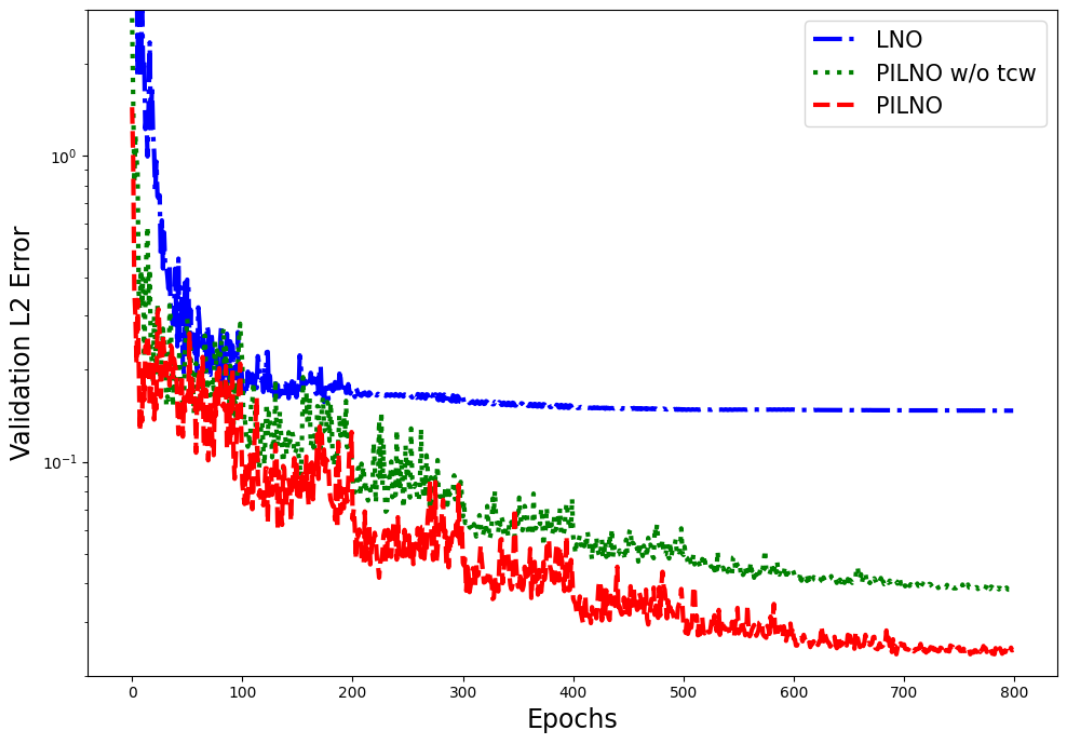}
        \caption{Validation relative $L_2$ error versus training epochs for the reaction-diffusion equation with $N_\text{train}=25$. Although all three models reduce the validation error as training progresses, the purely data-driven LNO saturates at approximately $18\%$ under this limited data regime. In contrast, PILNO achieves substantially lower error, and the inclusion of temporal causality weighting further reduces the validation error to about $2.5\%$, compared with roughly $4\%$ for PILNO without temporal causality weighting.}
        \label{fig:rd1d_tcw_validation_err}
    \end{figure}
    \begin{figure}[h]
        \centering
        \includegraphics[height=9cm]{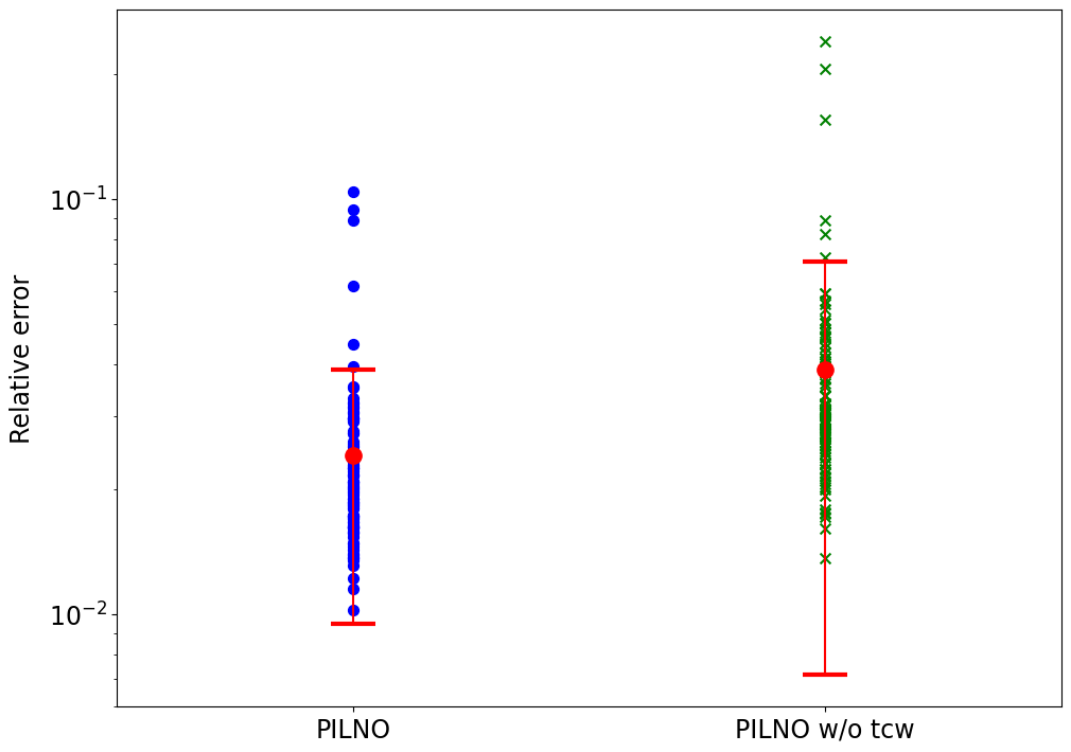}
        \caption{Distribution of relative $L_2$ errors over 100 test cases for PILNO and PILNO without TCW. Blue dots indicate the relative $L_2$ errors of PILNO, green crosses indicate those of PILNO without TCW, and red markers denote the mean error of each model with caps representing one standard deviation. PILNO and PILNO without TCW achieve mean relative $L_2$ errors of 2.42\% and 3.90\%, with standard deviations of $1.47\times10^{-2}$, $3.18\times 10^{-2}$, respectively, indicating that PILNO attains both a lower average error and reduced variability.}
        \label{fig:rd1d_tcw_error_plot}
    \end{figure}
    
    Consistent with Section~\ref{subsec:time_causal}, incorporating temporal causality is crucial for learning time-dependent PDEs. Figure~\ref{fig:rd1d_tcw_validation_err} compares LNO, PILNO (with TCW), and PILNO without TCW at $N_\text{train}=25$ and $\ell_\text{train}=0.5$, reporting validation relative $L_2$ error versus training epochs. In this small-data setting, LNO plateaus near $18\%$, which is reasonable because of its purely data-driven structure. Adding PDE constraints (PILNO) lowers the error, and TCW further improves both final accuracy and convergence: PILNO+TCW reaches $\approx 2.3\%$ of validation error, whereas PILNO without TCW settles around $3.5\%$ to $4.0\%$. The steeper decay of the PILNO with TCW model indicates faster optimization. 

    Figure~\ref{fig:rd1d_tcw_error_plot} compares the distribution of relative $L_2$ errors over 100 test cases for PILNO with and without TCW. Each blue dot corresponds to a single test error from PILNO with TCW, while green crosses denote errors from PILNO without TCW; red markers summarize the mean and one standard deviation for each model. PILNO with TCW achieves a lower mean error (2.42\%) than its counterpart without TCW (3.90\%), and its error distribution is more tightly concentrated, with smaller standard deviation ($1.47\times10^{-2}$ versus $3.18\times10^{-2}$). This indicates that temporal‑causality weighting not only improves average prediction accuracy but also reduces variability across test instances, leading to more reliable performance. Taken together, these results show that enforcing temporal causality yields more accurate and stable training for time-dependent PDEs.

\subsection{Forced KdV equation}\label{subsec:fkdv}
    We finally consider the one-dimensional forced Korteweg-de Vries (KdV) equation, a nonlinear dispersive PDE with external forcing, to compare PILNO against the DeepOMamba~\cite{hu2025deepomamba}, the strong spatio-temporal operator learning model. This example evaluates an in-distribution data efficiency study. We adopt the same benchmark setting and data-generation procedure used in DeepOMamba~\cite{hu2025deepomamba}, and evaluate data efficiency by varying the number of paired training samples.

\paragraph{PDE formulation and operator-learning task}
    Following the same formulation adopted in DeepOMamba~\cite{hu2025deepomamba}, the forced KdV equation is
    \begin{equation}\label{eqn:fkdv}
        \frac{\partial u}{\partial t} + u\frac{\partial u}{\partial x} + \beta \frac{\partial^3u}{\partial x^3} = \alpha \frac{\partial f}{\partial x}, \qquad (x,t)\in(-L,L)\times(0,T),
    \end{equation}
    with $L=5$ and $T=5$. For well-posedness, the system is associated with three boundary conditions $u(L,t)$, $u(-L,t)$, and $u_x(L,t)$ for $t\in(0,T)$, together with an initial condition $u(x,0)$ for $x\in(-L,L)$. The operator learning objective is to map the initial-boundary conditions, the external forcing $f_x(x,t)$, and PDE coefficients $(\alpha,\beta)$ to the full spatio-temporal solution $u(x,t)$ over the entire domain.
    \begin{equation}\label{eqn:fkdv_operator}
    \mathcal{G}_{\Theta}:\ [u(L,t),u(-L,t),u_x (L,t), u(x,0), f_x(x,t), \alpha, \beta] \ \longmapsto\ u(x,t),
    \end{equation}

\paragraph{Data generation and discretization}
    Following the data generation procedure of DeepOMamba~\cite{hu2025deepomamba}, we generate datasets from three analytic one-soliton solution families (Types A-C) with equal proportions. All inputs and outputs are discretized on a uniform $100\times 100$ space-time grid over $(x,t)\in(-5,5)\times (0,5)$. We first generate a full training pool of 27K instances (one third per type), and fix the validation and test sets to $N_\text{val}=3000$ and $N_\text{test}=3000$ instances, respectively. To evaluate data efficiency in the small-data regime, we construct training data by subsampling from the 27K training pool with $N_\text{train}\in\{2700,\ 270,\ 135,\ 54,\ 27\}$, always drawing equal numbers from Types A-C (e.g., $900/900/900$ samples from Type A, B, and C at $N_\text{train}=2700$).

\paragraph{Models and training protocol}
    \textbf{DeepOMamba.}
    We use the DeepOMamba architecture and forced-KdV hyperparameter choices reported in DeepOMamba~\cite{hu2025deepomamba}. Details of data generation procedure and model structure are provided in Appendix~D.
    \textbf{PILNO.}
    We instantiate PILNO with the same LNO/ALNO backbone used throughout this paper but with model depth of 2, and train it with data loss and physics-informed residual losses. For forced KdV, the physics term enforces Eq.~\eqref{eqn:fkdv}, together with residual penalties for the required IC/BC constraints. In addition, PILNO uses $N_\text{virt}=1800$ virtual inputs drawn from the same procedure stated above and uses temporal-causality weighting to stabilize time-dependent training.

    \begin{figure}[ht]
        \centering
        \includegraphics[height=9cm]{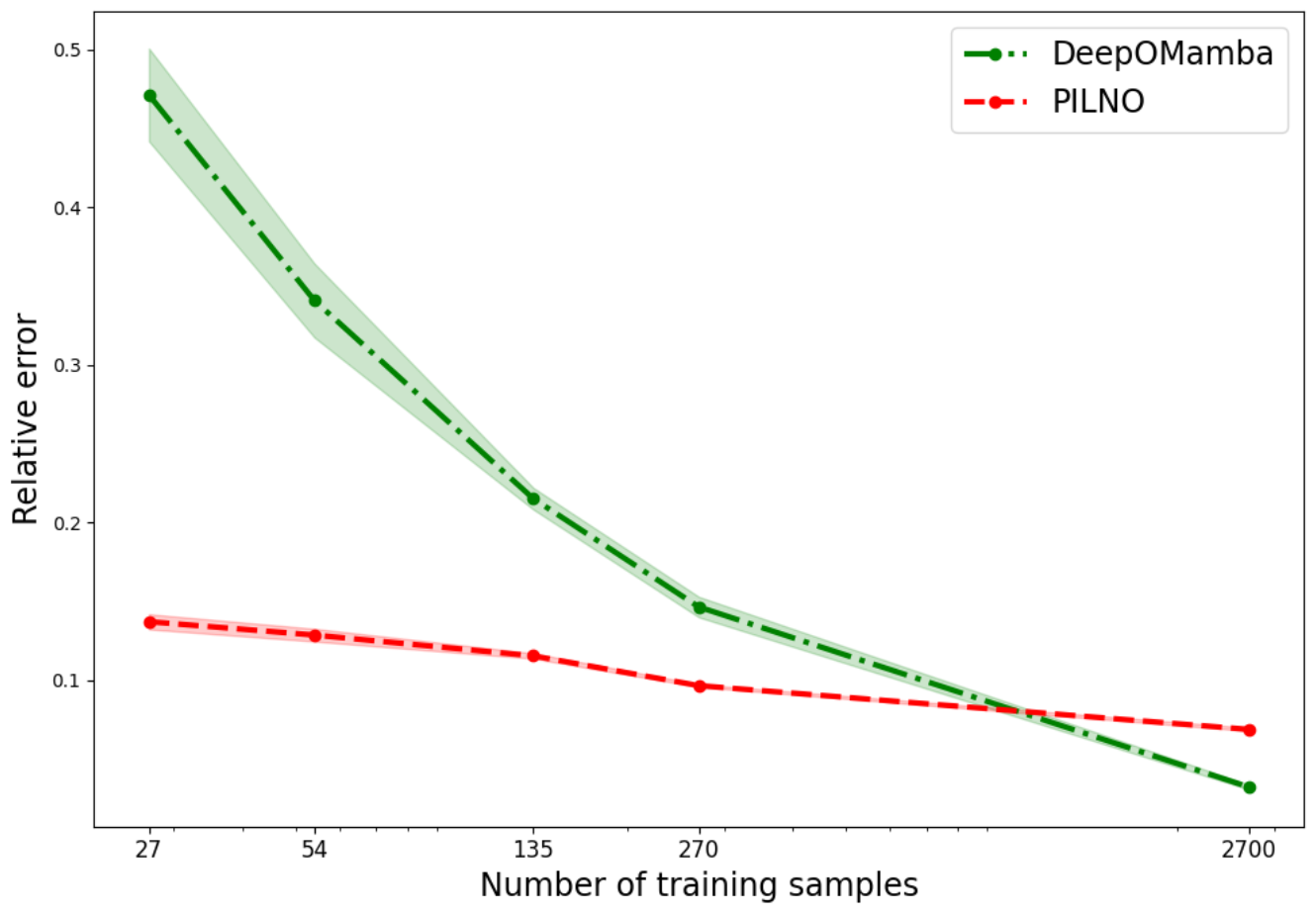}
        \caption{Relative $L_2$ error on the test set versus the number of training samples (mean $\pm$ std over 5 random seeds). As the number of training samples decreases from $N_\text{train}=2700$ to $N_\text{train}=27$, the error of DeepOMamba increases sharply from $3.24\%$ to $47.13\%$, whereas PILNO's error rises only moderately from $6.89\%$ to $13.72\%$.}
        \label{fig:fkdv_ntrain}
    \end{figure}
    
    For a fair comparison under the DeepOMamba benchmark, both models are trained using the same train/validation splits. We report the relative $L_2$ error on the test set, averaged over five random seeds (mean $\pm$ std). Figure~\ref{fig:fkdv_ntrain} reports the test error as a function of the number of training samples, $N_\text{train}$, providing a systematic assessment of performance across large sample and small sample regimes. At $N_{\text{train}}=2700$, DeepOMamba achieves a test error of 3.24\% compared with 6.89\% for PILNO, showing the advantage of fully supervised spatio-temporal operator learning in the data-abundant regime for time-dependent PDEs. As $N_{\text{train}}$ is reduced, however, PILNO exhibits a markedly more graceful degradation, which is consistent with the regularizing effect of physics loss and virtual inputs in the small-data regime. This divergence is most evident at $N_{\text{train}}=27$, where the paired training set is extremely limited: PILNO achieves a test error of 13.72\%, whereas DeepOMamba's error increases sharply to 47.13\%. Moreover, PILNO exhibits more stable performance in this extreme low-data setting, with a substantially smaller standard deviation at $N_\text{train}=27$ (0.48\%) than DeepOMamba (2.94\%).
   
    \begin{figure}[ht]
        \centering
        \includegraphics[width=16cm]{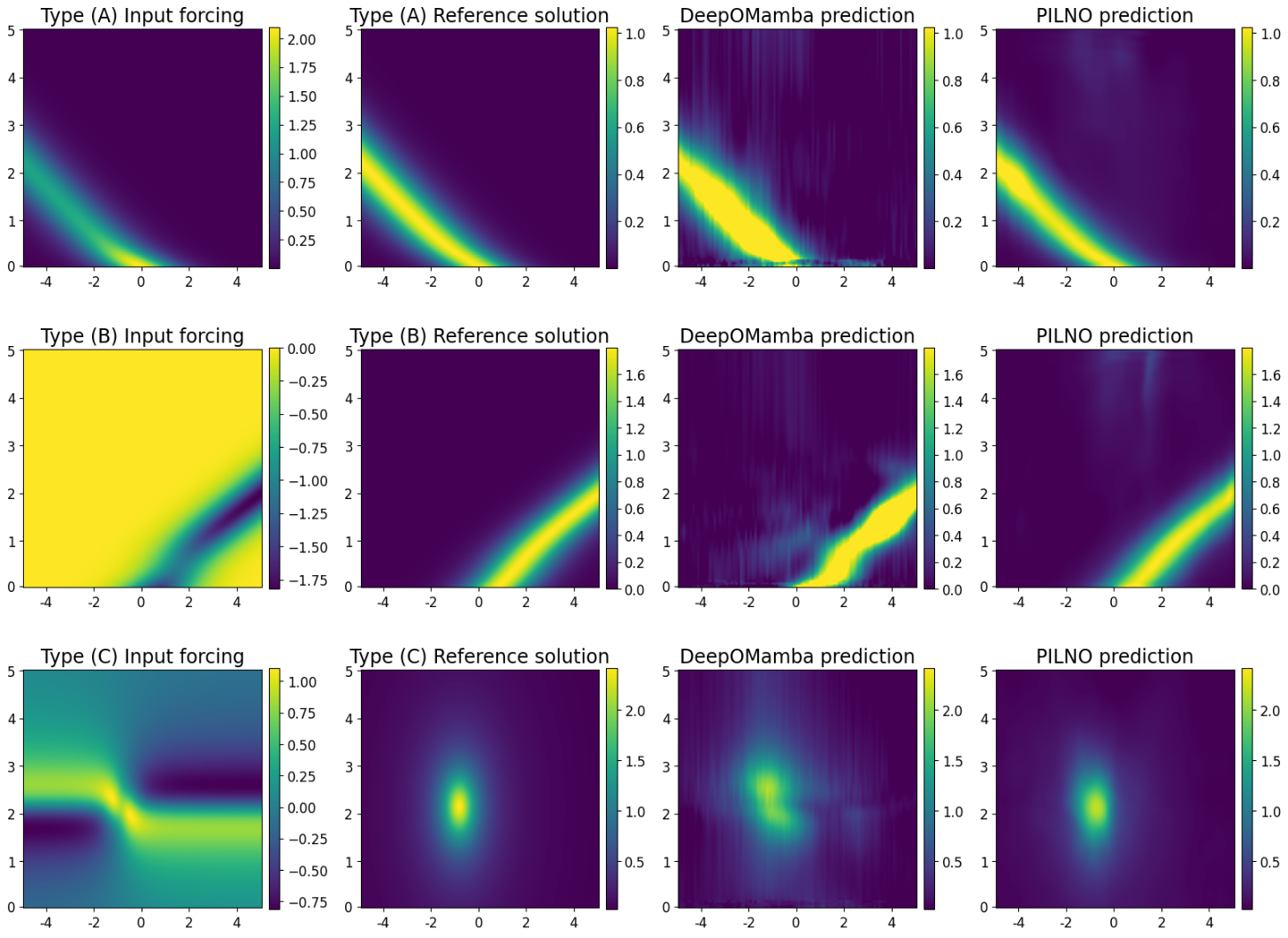}
        \caption{Qualitative comparison of DeepOMamba and PILNO at $N_\text{train}=27$ on representative test instances. The instances are randomly sampled from those whose errors fall within a $\pm 2\%$ band around the mean error of DeepOMamba and, simultaneously, within a $\pm 2\%$ band around the mean error of PILNO. Rows correspond to Types A, B, and C (top to bottom). Columns show the input forcing $f(x,t)$, the corresponding reference solution, the DeepOMamba prediction, and the PILNO prediction (left to right). The per-instance relative $L_2$ errors for DeepOMamba are 45.36\%, 48.50\%, and 45.71\% for Type A, B, and C, respectively, while those for PILNO are 10.90\%, 12.30\%, 14.14\% on the same instances.}
        \label{fig:fkdv_plot_prediction}
    \end{figure}

    \begin{figure}[ht]
        \centering
        \includegraphics[width=16cm]{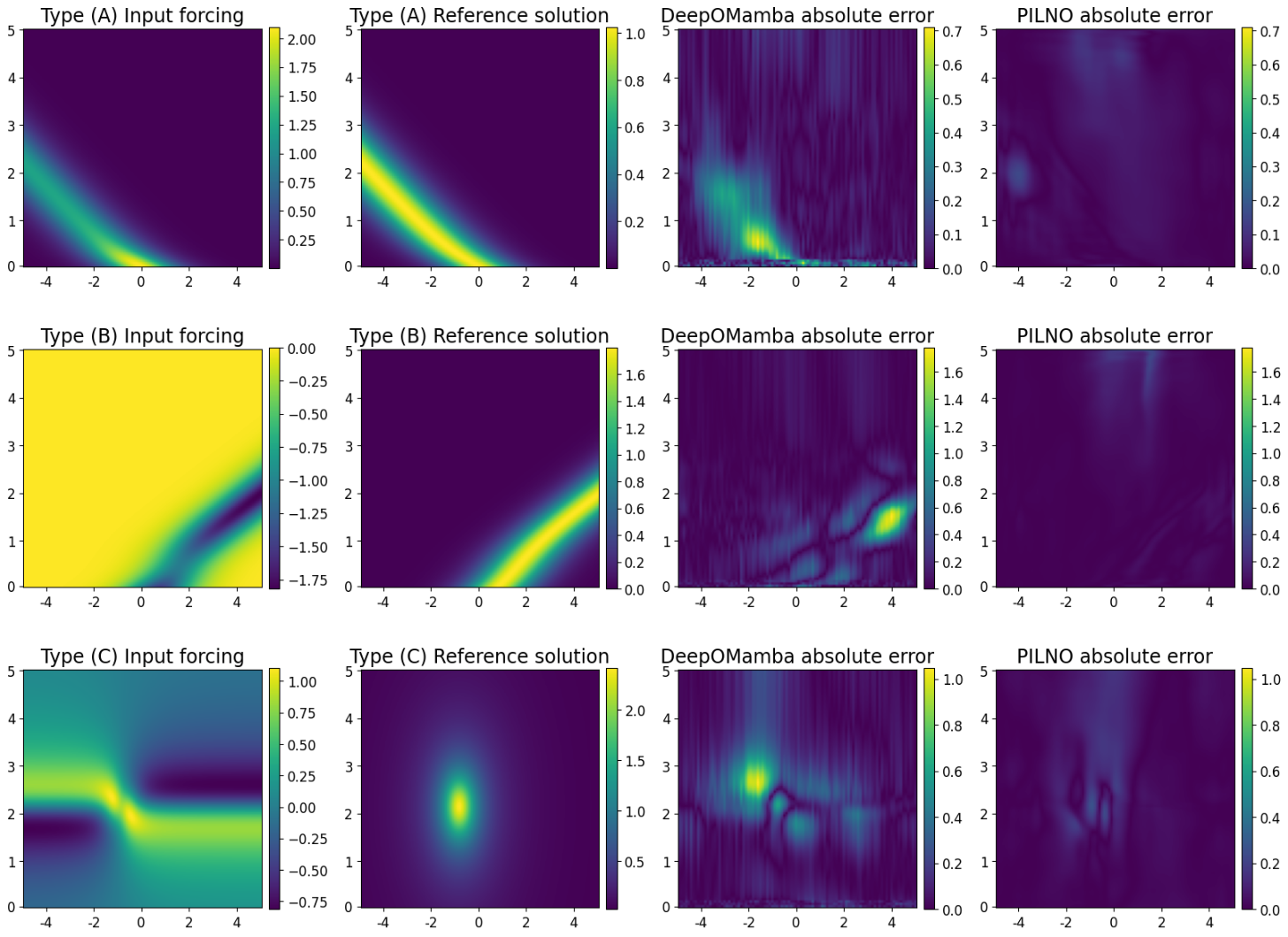}
        \caption{Absolute error comparison of DeepOMamba and PILNO at $N_\text{train}=27$ on the same test instances as in Fig.~\ref{fig:fkdv_plot_prediction}. Rows correspond to Types A, B, and C (top to bottom). Columns show the input forcing $f(x,t)$, the corresponding reference solution, the absolute error of DeepOMamba prediction, and the absolute error of PILNO prediction (left to right).}
        \label{fig:fkdv_plot_error}
    \end{figure}
    
    For qualitative visualization, we randomly sampled one representative test instance per solution type (A/B/C) from the intersection of test cases whose per-instance relative $L_2$ errors for DeepOMamba and for PILNO each fall within $\pm 2$ percentage points of the corresponding model's mean test error. We then visualize the forcing, the reference solution, and the predictions (and absolute errors) of both DeepOMamba and PILNO for the same instances. Figure~\ref{fig:fkdv_plot_prediction} provides qualitative comparisons in the small-data regime ($N_{\text{train}}=27$) across three representative forcing categories (Types A–C). For each case, we report the input forcing, the corresponding reference solution, and the predictions produced by DeepOMamba and PILNO. Across all types, PILNO exhibits closer agreement with the reference solution than DeepOMamba, more faithfully capturing the evolution of the solution and its salient spatiotemporal structures. Figure~\ref{fig:fkdv_plot_error} presents, for the same cases as in Figure~\ref{fig:fkdv_plot_prediction}, the input forcing, the reference solution, and the absolute error fields for DeepOMamba and PILNO. The error visualizations further indicate that PILNO achieves higher fidelity reconstructions, with markedly smaller absolute errors over the domain relative to DeepOMamba.

    Overall, this benchmark highlights a clear trade-off between fully supervised spatio-temporal operator learning and physics-informed regularization. DeepOMamba attains the best accuracy when paired supervision is abundant, whereas PILNO delivers substantially improved data efficiency and stability as paired data become scarce, consistent with the combined effect of residual constraints, virtual inputs, and temporal-causality weighting. Since both training and testing are drawn from the same analytic families (Types A--C), these results should be interpreted as an in-distribution data-efficiency and robustness study rather than an evaluation of out-of-distribution generalization. Note that, under the same labeled (paired) data budget, PILNO further exploits the known governing equation through PDE/IC/BC residual losses (and, when enabled, additional virtual inputs and temporal-causality weighting); accordingly, the comparison is intended to quantify the practical benefit of physics-informed operator learning when labeled supervision is limited.

\subsection{\tcd{Computational cost and accuracy-cost trade-off}}\label{subsec:computational_cost}
\tcd{We next examine the computational cost of PILNO and its accuracy-cost trade-off. The purpose of this comparison is not to suggest that PILNO is uniformly preferable to the data-driven baselines across all criteria. Rather, PILNO trades additional offline training cost for improved data efficiency, lower prediction error, and more stable training in small-data regimes. Since the physics residuals, virtual inputs, and temporal-causality weighting are used only during training, they do not increase the inference-time cost of the underlying neural-operator backbone. Detailed practical guidance on hyperparameter selection and scalability is provided in Appendix~E.}

\tcd{Table~\ref{tab:computational_table} reports the wall-clock training time and relative \(L_2\) test error of PILNO and the corresponding data-driven baselines. For each benchmark, the compared models were trained and profiled on the same GPU and under the same training configuration described in the preceding subsections. For the Darcy-flow problem, the profiling comparison was conducted on the same \(61\times 61\) grid for both LNO and PILNO, thereby enabling a direct comparison under identical grid-level settings. Because different benchmark problems were run on different GPUs and grid resolutions, the absolute wall-clock times should be interpreted within each benchmark rather than across different PDEs.}

\tcd{Across all benchmark problems, PILNO requires a larger offline training time than the corresponding data-driven baseline because it additionally evaluates physics residuals and, when enabled, residuals on virtual inputs. However, this additional cost is accompanied by a substantial reduction in test error. For example, in the reaction--diffusion Task B problem, PILNO reduces the relative \(L_2\) test error from \(1.229\times 10^{-1}\) to \(2.563\times 10^{-2}\), while increasing the training time from \(459~\mathrm{s}\) to \(2521~\mathrm{s}\). Similar accuracy gains are observed for Burgers' equation, Darcy flow, and the forced KdV benchmark. These results indicate that PILNO should be interpreted as a method oriented toward accuracy and data efficiency rather than as a method designed to minimize offline training time.}

\begin{table}[!htbp]
    \centering
    \small
    \setlength{\tabcolsep}{10pt}
    \renewcommand{\arraystretch}{1.4}
    \newcommand{\headerpad}{\rule[-4pt]{0pt}{24pt}}
    \begin{tabular}{c|c|c|c|c|c}
    \hline
    \headerpad
    Problem & GPU & Grid & Model & \shortstack{Wall-clock\\training time} & \shortstack{Relative $L_2$\\test error} \\
    \hline
    \multirow{2}{*}{Burgers} & \multirow{2}{*}{\shortstack{RTX 4090\\(24 GB)}} & \multirow{2}{*}{32 $\times$ 26}
    & LNO & 459 s & 9.386e-2  \\
    \cline{4-6}
    & & & PILNO & 6039 s & 1.420e-2  \\
    \hline

    \multirow{2}{*}{Darcy} & \multirow{2}{*}{\shortstack{A100\\(40 GB)}} & \multirow{2}{*}{61 $\times$ 61}
    & LNO & 612 s & 1.499e-1  \\
    \cline{4-6}
    & & & PILNO & 3299 s & 1.235e-2  \\
    \hline

    \multirow{2}{*}{\shortstack{Reaction-diffusion \\ Task B}} & \multirow{2}{*}{\shortstack{A100\\(40 GB)}}  & \multirow{2}{*}{51 $\times$ 51}
    & LNO & 459 s & 1.229e-1  \\
    \cline{4-6}
    & & & PILNO & 2521 s & 2.563e-2  \\
    \hline

    \multirow{2}{*}{Forced KdV} & \multirow{2}{*}{\shortstack{RTX 4090\\(24 GB)}} & \multirow{2}{*}{100 $\times$ 100}
    & DeepOMamba & 206 s & 4.713e-1  \\
    \cline{4-6}
    & & & PILNO & 3002 s & 1.372e-1  \\
    \hline

    \end{tabular}
    \caption{Wall-clock training time and relative $L_2$ test error of PILNO and the corresponding data-driven baselines in the small-data regime. For each benchmark, the compared models were trained and profiled on the same GPU; therefore, wall-clock times should be interpreted within each benchmark rather than across different PDEs.}
    \label{tab:computational_table}
\end{table}

\begin{figure}[!htbp]
        \centering
        \begin{subfigure}{0.31\linewidth}
            \centering \includegraphics[height=4.5cm]{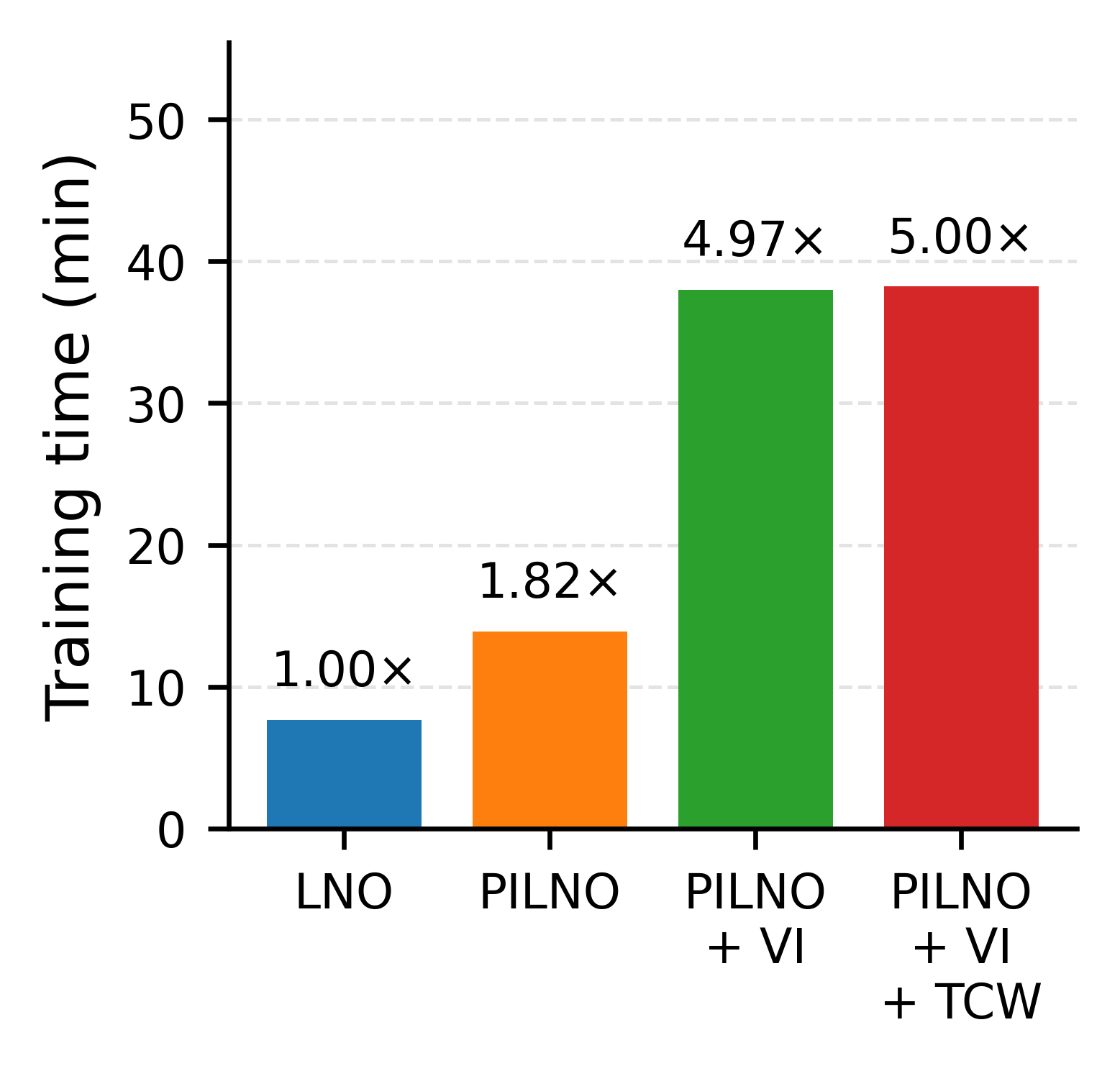}
            \caption{Wall-clock training time}
        \end{subfigure}
        \centering
        \begin{subfigure}{0.31\linewidth}
            \centering \includegraphics[height=4.5cm]{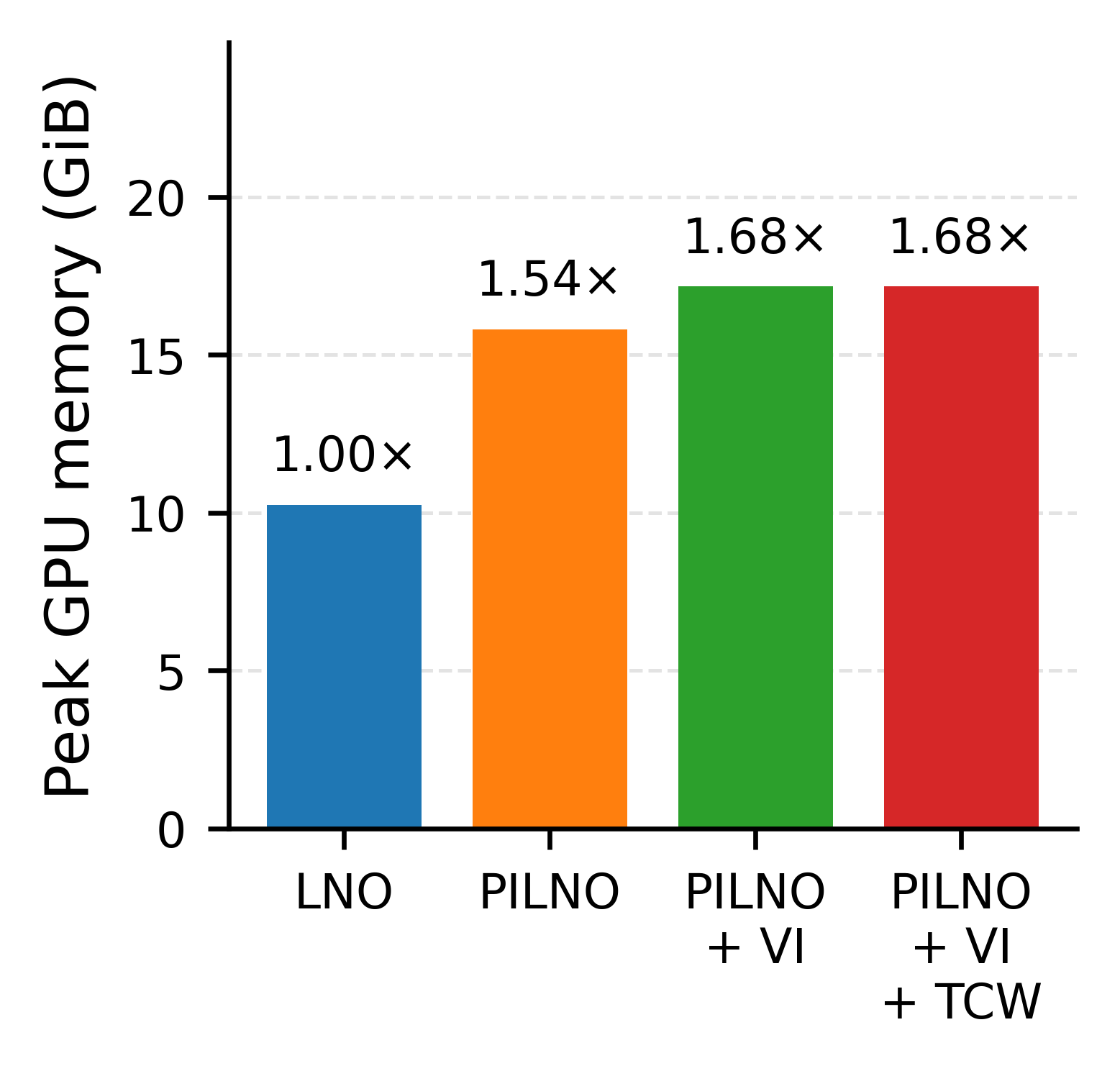}
            \caption{Peak GPU memory}
        \end{subfigure}
        \centering
        \begin{subfigure}{0.31\linewidth}
            \centering \includegraphics[height=4.5cm]{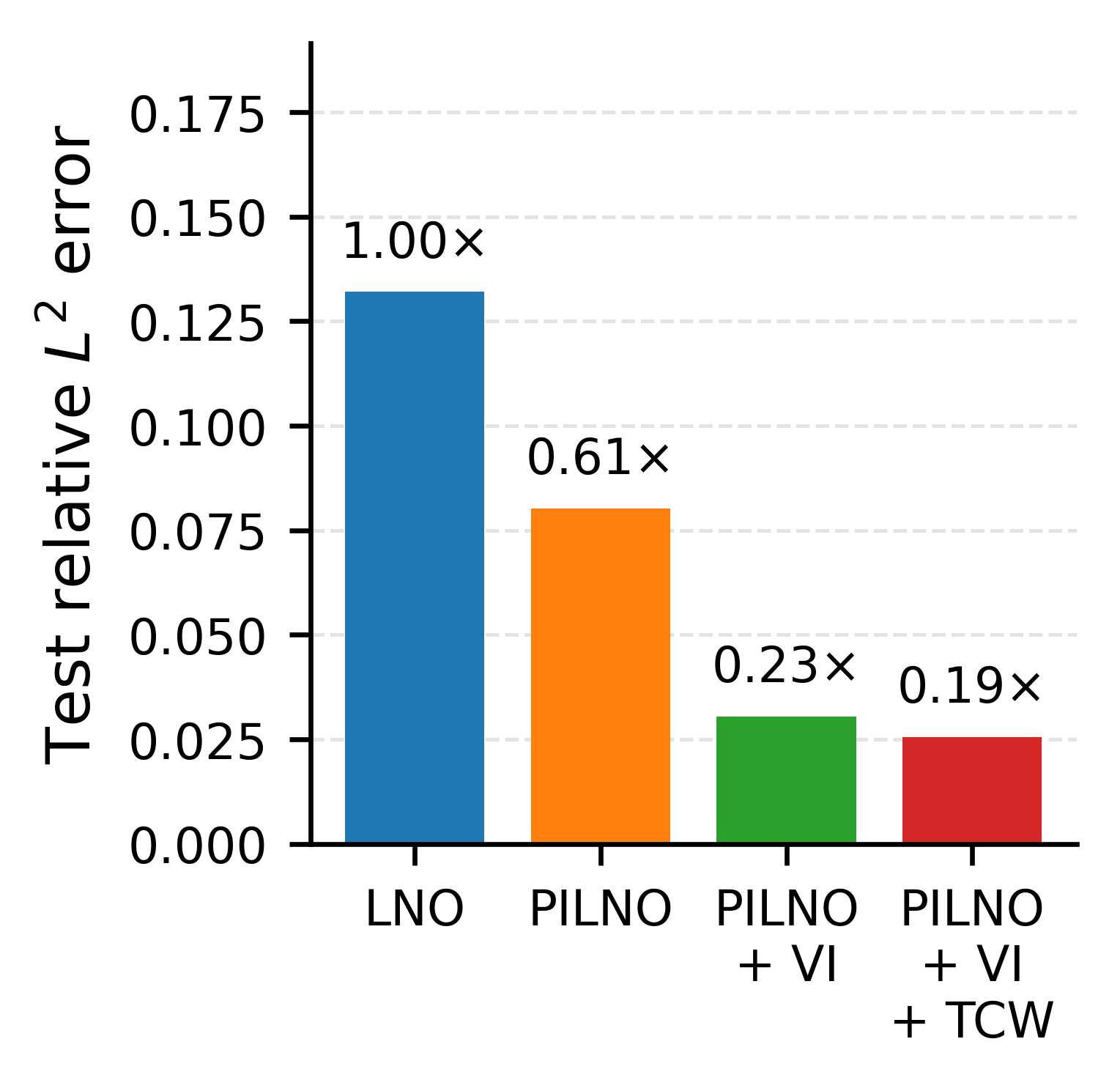}
            \caption{Test relative $L_2$ error}
        \end{subfigure}
        \caption{Cost versus performance ablation study for LNO and PILNO variants. We report wall-clock training time, peak GPU memory usage, and test relative $L_2$ error on the same benchmark and GPU. Here, PILNO denotes the physics-informed model without virtual inputs and without TCW. VI denotes virtual inputs, and TCW denotes temporal-causality weighting. The values above the bars are normalized by the LNO baseline.}
        \label{fig:comp_cost_ablation}
    \end{figure}

\tcd{To further identify the sources of the additional computational cost, we perform a component-wise ablation study on the reaction--diffusion Task B problem. In this ablation, "PILNO" denotes the physics-informed model without virtual inputs and without temporal-causality weighting. We then progressively add virtual inputs and temporal-causality weighting. Figure~\ref{fig:comp_cost_ablation} summarizes the resulting wall-clock training time, test error, and peak GPU memory usage.}

\tcd{The accuracy trend shows that each component contributes to the final performance. Adding the physics residual alone already improves the prediction accuracy, reducing the relative \(L_2\) test error by approximately 39\% compared with the data-driven LNO. This behavior differs from the Darcy-flow small-data case, where virtual inputs are essential for maintaining reliable physics-informed learning when very few paired samples are available. Adding virtual inputs further reduces the error, yielding an approximately 77\% reduction relative to LNO and demonstrating that label-free physics supervision substantially improves robustness in the small-data regime. Finally, TCW provides an additional improvement over the virtual-input model, indicating that temporal reweighting is beneficial for learning time-dependent dynamics.}

\tcd{The computational overhead follows a different trend. The physics residual increases both memory usage and training time because derivatives and residual terms must be evaluated on the predicted solution field. Virtual inputs introduce the largest additional training-time cost, since they increase the number of physics-residual evaluations without adding paired solution labels. In contrast, TCW adds almost no additional memory overhead and only a marginal increase in wall-clock cost once the time-slice residuals have already been computed. In the reaction--diffusion ablation, adding TCW leaves the peak GPU memory essentially unchanged and increases the wall-clock training time by less than 1\% relative to the PILNO+VI variant. This confirms that TCW acts primarily as a low-cost reweighting of the already-computed temporal residuals.}

\begin{figure}[!htbp]
        \centering
        \includegraphics[height=8cm]{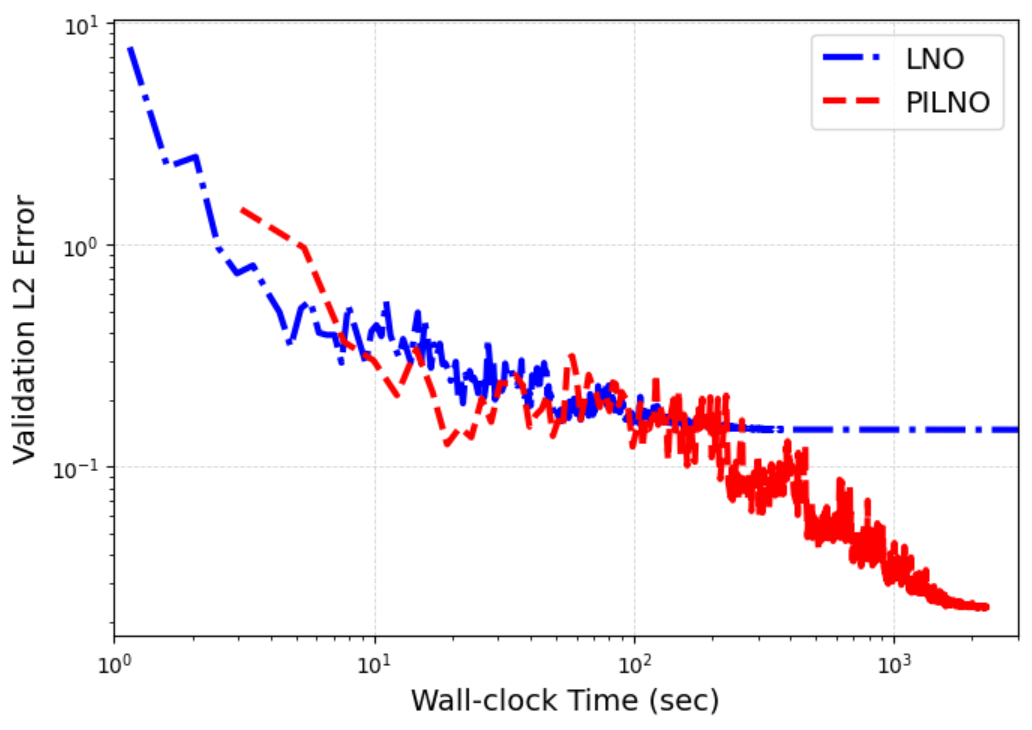}
        \caption{Validation relative $L_2$ error versus wall-clock training time for LNO and PILNO on the reaction-diffusion Task B problem. LNO's validation error decreases rapidly during the early stage of training but soon saturates at approximately 14\%, showing no further improvement even with extended training. In contrast, PILNO continues to decrease and eventually reaches a substantially lower validation error, highlighting the accuracy-cost trade-off of PILNO.}
        \label{fig:rd_wallclock}
    \end{figure}

\tcd{Figure~\ref{fig:rd_wallclock} shows the validation relative \(L_2\) error as a function of wall-clock time for the reaction-diffusion Task B problem. Because the data-driven LNO has a substantially shorter wall-clock training time than PILNO, we additionally trained LNO for up to \(5{,}000\) epochs to assess whether longer optimization alone could close the performance gap. Although the validation error of LNO decreases rapidly during the early stage of training, it soon saturates at approximately \(14\%\) and shows no further improvement even with extended training. This result indicates that the relatively high error of the data-driven LNO cannot be explained solely by insufficient training time. In contrast, PILNO requires more wall-clock time because of the physics-residual evaluations, but it continues to decrease and eventually reaches a substantially lower validation error. This behavior highlights the main accuracy--cost trade-off of PILNO: it is not designed to minimize the cost of each training step, but to obtain a more accurate and stable operator when labeled data are scarce.}

\section{Conclusion}\label{sec:conclusion}
    \tcb{This study enhances data-driven operator learning through PILNO, which incorporates governing-equation constraints into the ALNO backbone.} PILNO augments the operator-learning objective with PDE, boundary, and initial-condition residuals, and addresses two persistent challenges—scarce labeled data and sensitivity to \tcb{OOD} inputs—by complementing standard physics losses with label-free virtual inputs. For time-dependent PDEs, we further employ temporal-causality weighting to bias the residual toward early times, promoting stable optimization and causally coherent predictions. Together, these components yield a principled framework for physically constrained and sample-efficient operator learning.

    Empirically, across Burgers’ equation, Darcy flow, reaction--diffusion systems, and forced KdV equation, PILNO consistently achieves lower errors than LNO, the data-driven baseline; on the forced KdV benchmark, we additionally compare against DeepOMamba and observe substantially improved data efficiency for PILNO as paired data become extremely scarce. In particular, test errors with respect to the number of training samples demonstrate improved data efficiency, while error heatmaps over input length-scales demonstrate stronger OOD generalization, with PILNO sustaining substantially lower errors on off-diagonal train--test combinations than purely data-driven operators. Ablations further highlight that virtual inputs are critical for maintaining accuracy in the extreme small-data regime (e.g., Darcy flow), and that TCW further improves both training stability and predictive accuracy for time-dependent problems (e.g., reaction--diffusion).

    \tcb{Nevertheless, there remain important avenues for future work. The present OOD analysis is mainly restricted to spectral/statistical shifts in input functions, represented by changes in correlation length-scales. Extending this analysis to broader OOD settings, such as unseen PDE parameters and different boundary conditions, and improving robustness under such shifts are important directions for future work.}

    Moreover, because the use of physics residuals is rooted in the core premise of physics-informed neural networks (PINNs)~\cite{raissi2019physics}, and because both virtual inputs and temporal-causality weighting adapt techniques proposed to mitigate known PINN pathologies~\cite{wang2022respecting,jung2024ceens}, systematically transferring, refining, and assessing recent PINN advances within neural-operator architectures is a promising direction for future work. Examples include adaptive residual weighting, curriculum strategies for collocation sampling, and multi-physics constraints tailored to operator learning. We expect that such cross-fertilization between PINNs and operator learning further improves robustness, data efficiency, and long-horizon stability in scientific machine learning.

\section*{Acknowledgements}
This work was supported by the National Research Foundation of Korea [NRF-2021R1C1C1007875] and by the National Research Council of Science \&
Technology (NST) grant by the Korea government (MSIT) (CRC20014-000).

\section*{Appendix A. Additional Comparison of ALNO, LNO, and FNO on Burgers’ Equation}\label{sec:appendix_a}
    In Section~\ref{subsec:alno}, we introduced ALNO, which retains the interpretable pole-residue transient structure of LNO while increasing expressivity by replacing the steady-state branch with an FNO-style spectral module. Here we provide a focused comparison of ALNO, vanilla LNO, and FNO to assess whether this hybrid design yields tangible gains in approximation accuracy.

    All results in this appendix use the same Burgers' equation setup as in Section~\ref{subsec:burgers}: periodic domain $x\in[0,1]$, viscosity $\nu=0.01$, time horizon $t\in[0,1]$ discretized with $h=1/25$ (26 snapshots), and $N_x=32$ spatial nodes. Initial conditions are drawn from the Gaussian process with exponentiated squared-sine kernel at length-scale $\ell_{\text{train}}=\ell_{\text{test}}=0.5$. Each model is trained with $N_{\text{train}}=200$ paired samples under the same optimizer, learning-rate schedule, and a comparable number of parameters.
    
    \begin{table}[ht]
        \centering
        \renewcommand{\arraystretch}{1.4}
        \begin{tabular}{l c c c}
            \hline
            & \textbf{LNO} & \textbf{ALNO} & \textbf{FNO} \\
            \hline
            \textbf{Relative \(L_2\) Error} & \(3.444 \times 10^{-2}\) & \(3.312 \times 10^{-3}\) & \(8.089 \times 10^{-3}\) \\
            \hline
        \end{tabular}
        \caption{Relative $L^2$ errors on Burgers’ equation (periodic, $\nu=0.01$, $h=1/25$, $N_x=32$). Lower is better.}
        \label{tab:burgers_error}
    \end{table}
    
    \begin{figure}[ht]
        \centering
        \includegraphics[width=0.8\linewidth]{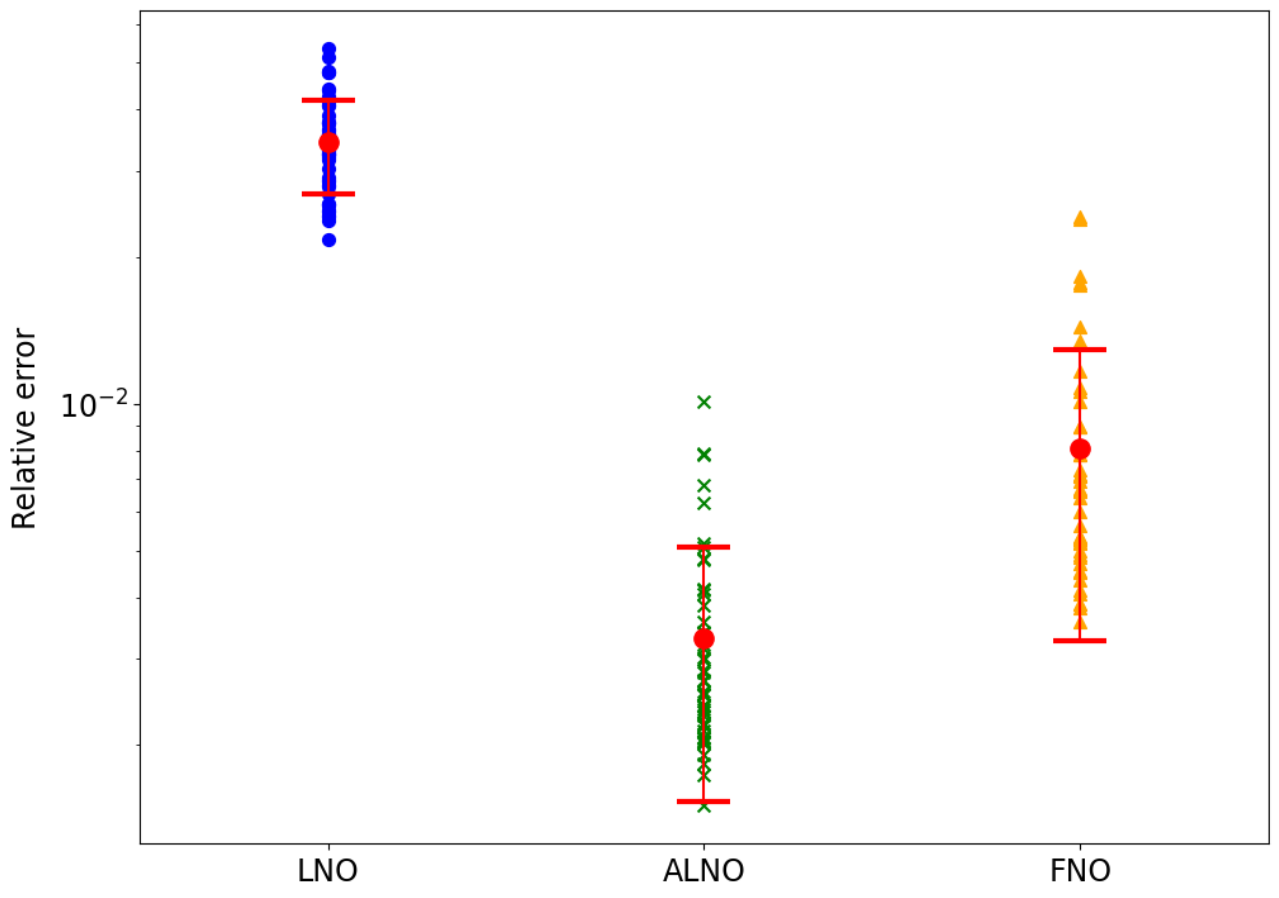}
        \caption{Distribution of relative $L_2$ errors over 50 test cases for LNO, ALNO, and FNO. Blue dots indicate the relative $L_2$ errors of LNO, green crosses indicate those of ALNO, and yellow triangles indicate those of FNO. Red markers denote the mean error of each model with caps representing one standard deviation. LNO achieves mean relative $L_2$ error of $3.444\times10^{-2}$ and standard deviation of $7.504\times10^{-3}$. ALNO achieves mean relative $L_2$ error of $3.312\times10^{-3}$ and standard deviation of $1.778\times10^{-3}$. FNO achieves mean relative $L_2$ error of $8.089\times10^{-3}$ and standard deviation of $4.816\times10^{-3}$}
        \label{fig:burgers_model_err}
    \end{figure}
    Table~\ref{tab:burgers_error} shows that vanilla LNO incurs the largest test error ($3.444\times 10^{-2}$), FNO improves substantially ($8.089\times 10^{-3}$), and ALNO further reduces the error to $3.312\times 10^{-3}$, an order of magnitude better than LNO and roughly a factor of $2.4$ better than FNO. Figure~\ref{fig:burgers_model_err} provides a finer view of the error distribution across the 50 test instances. ALNO not only achieves the lowest mean error but also exhibits the smallest spread, with noticeably fewer large-error outliers than either LNO or FNO. These results support the claim that decoupling the transient and steady-state branches---while keeping an interpretable pole-residue transient component and using a flexible Fourier multiplier for the steady response---yields a empirically more expressive and robust backbone for operator learning.

\section*{Appendix B. Diffusion equation}\label{appendix:diffusion}
    Here we consider the diffusion equation, a canonical model for macroscopic transport emerging from microscopic Brownian motion:
    \begin{equation}
        \begin{aligned}
            &D\dfrac{\partial^2u}{\partial x^2}- \dfrac{\partial u}{\partial t} = f(x,t), \quad x\in[0,1], \quad t\in[0,1],\\
            &u(t,0) = u(t,1)
        \end{aligned}
        \label{eqn:diffusion}
    \end{equation}
    where $u(t,x)$ denotes the scalar field and $D>0$ is the diffusivity. To train and evaluate the models, we discretize time step size $h=1/25$, yielding 26 snapshots $\{ t_k \}_{k=0}^{25}$, and select $N_x=32$ equidistant spatial nodes on $[0,1]$, identical to the Burgers' setup. The input to LNO and PILNO is the initial condition, and the target is the corresponding spatio-temporal solution.

    As our goal here is to assess generalization, we sample initial $u_0(x)$ from a Gaussian random field with the exponentiated squared sine kernel, using length-scales $\ell \in \{ 0.5,1.0,1.5,\dots,5.0 \}$. For each $\ell_{train}$, a model is trained and then evaluated on all $\ell_{test}$, producing a $10\times 10$ heat map of test errors (relative $L_2$), analogous to the Burgers' experiment.

\begin{figure}[H]
        \centering
        \begin{subfigure}[t]{0.49\linewidth}
            \centering \includegraphics[width=\linewidth]{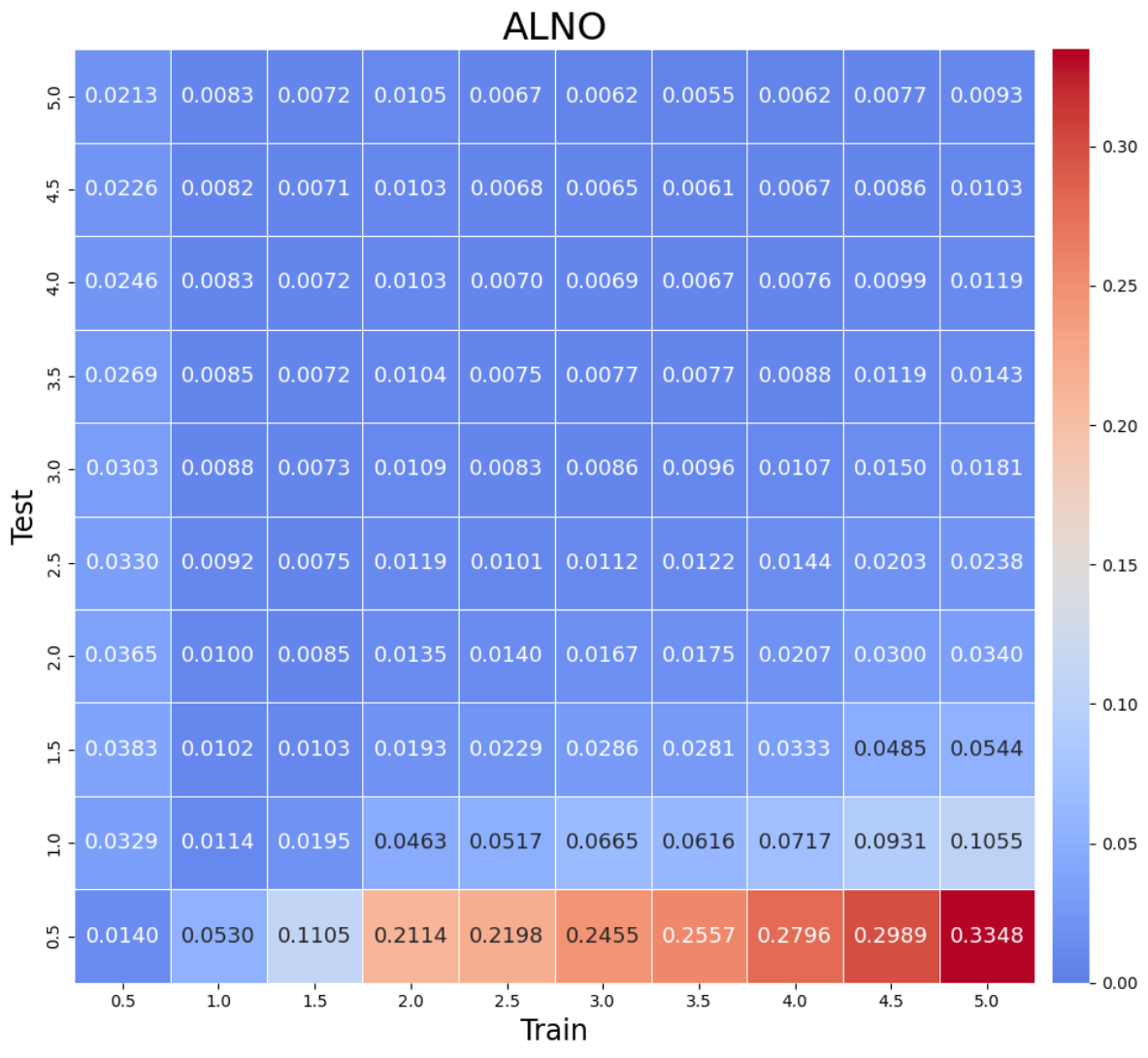}
            \caption{}
        \end{subfigure}
        \begin{subfigure}[t]{0.49\linewidth}
            \centering \includegraphics[width=\linewidth]{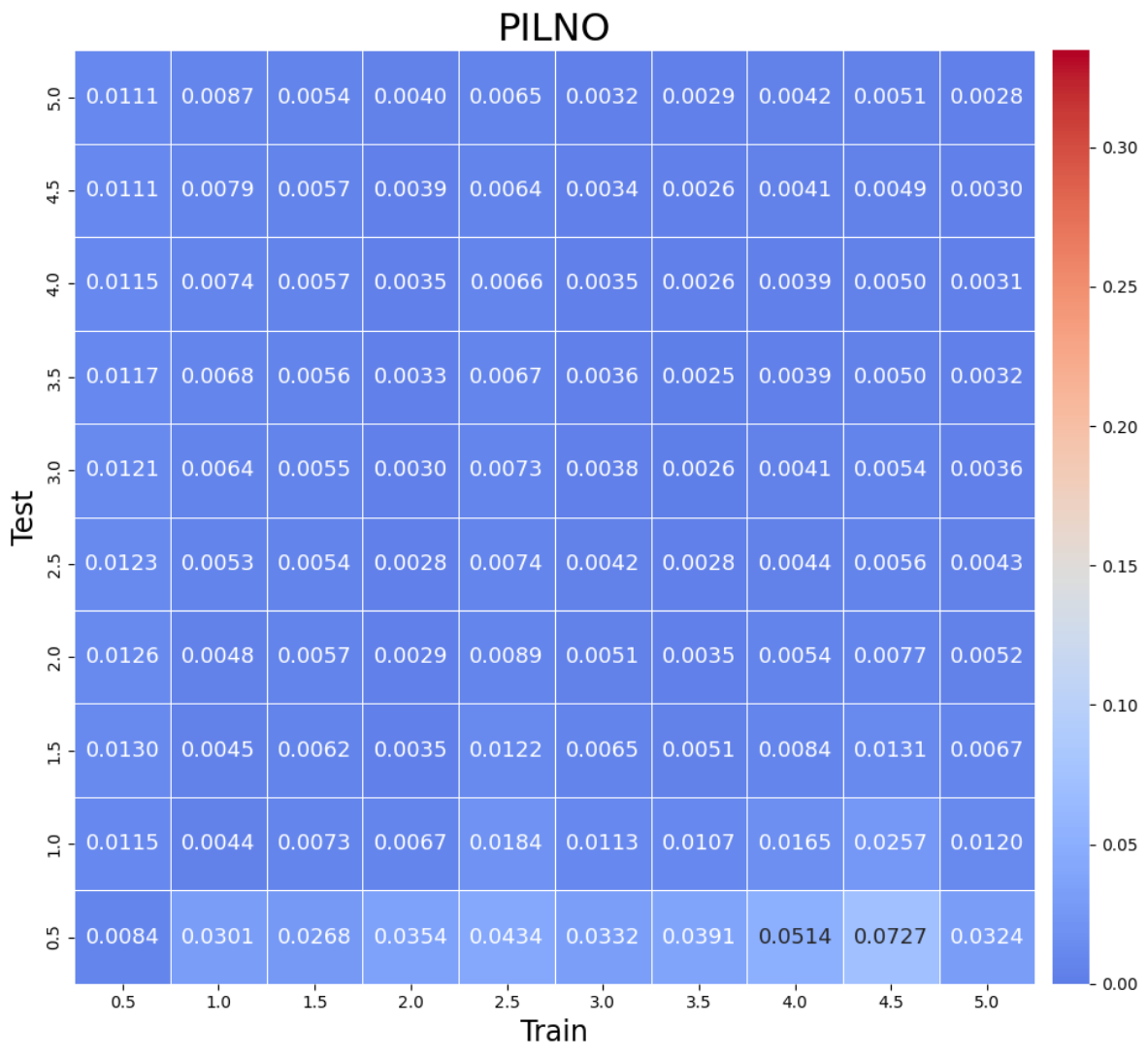}
            \caption{}
        \end{subfigure}
        \caption{Generalization across initial–condition length–scales for diffusion equation. Heatmaps show relative $L^2$ error when training at length–scale $\ell_{\text{train}}$ (horizontal axis) and testing at $\ell_{\text{test}}$ (vertical axis), for (a) ALNO and (b) PILNO. PILNO sustains lower errors off the diagonal (mismatched smooth/oscillatory regimes), demonstrating stronger out–of–distribution generalization.}
        \label{fig:diffusion_generalization}
    \end{figure}

    Figure~\ref{fig:diffusion_generalization} reports generalization heatmaps for ALNO and PILNO. As expected, errors are lowest near the diagonal $\ell_{train} \approx \ell_{test}$. Off-diagonal evaluations (mismatch between smooth and oscillatory initial conditions for training and test datasets) are more challenging: ALNO exhibits a pronounced degradation when extrapolating across widely separated length-scales (e.g., from large $\ell$ to small $\ell$). In contrast, PILNO maintains substantially lower errors across these off-diagonal regimes, indicating that the physics-informed residual effectively constrains the learned dynamics beyond the training distribution.

    On diffusion, where the underlying operator favors smoothing dynamics, PILNO consistently improves out–of–distribution performance across initial–condition length–scales, mirroring the robustness observed in Burgers’ equation while isolating the benefit of the physics–informed structure under a linear parabolic PDE.

\section*{Appendix C. Temporal-Causality Weighting (TCW) Ablation Study Without Virtual Inputs}\label{appendix:tcw_wo_virt}
    In Section~\ref{subsec:darcy}, we demonstrated the effect of virtual inputs, and in Section~\ref{subsec:reaction_diffusion} we evaluated the effect of temporal-causality weighting (TCW). Since the experiments in Section~\ref{subsec:reaction_diffusion} were conducted with virtual inputs, it is unclear whether TCW provides a standalone benefit when virtual inputs are removed. To isolate this effect, we repeat the training protocol of Section~\ref{subsec:reaction_diffusion} with $N_{\mathrm{virt}}=0$; all other models and hyperparameters are kept identical.

    TCW is designed to address a temporal optimization pathology in physics-informed training: during long-horizon prediction, early-time errors can propagate forward, and the physics loss may be reduced by fitting late-time residuals while leaving substantial early-time inconsistencies (late-time masking). TCW mitigates this instability by emphasizing early-time residuals. Its practical impact is expected to be largest when the physics loss dominates optimization. Virtual inputs create such a regime by removing supervised anchors on additional inputs, whereas in the fully labeled setting ($N_{\mathrm{virt}}=0$) the data loss typically anchors training and the physics loss acts more as a regularizer, making TCW’s standalone gains harder to observe. In other words, the standalone effect of TCW becomes harder to observe and may yield only marginal improvements.

\begin{figure}[ht]
        \centering
        \includegraphics[width=0.8\linewidth]{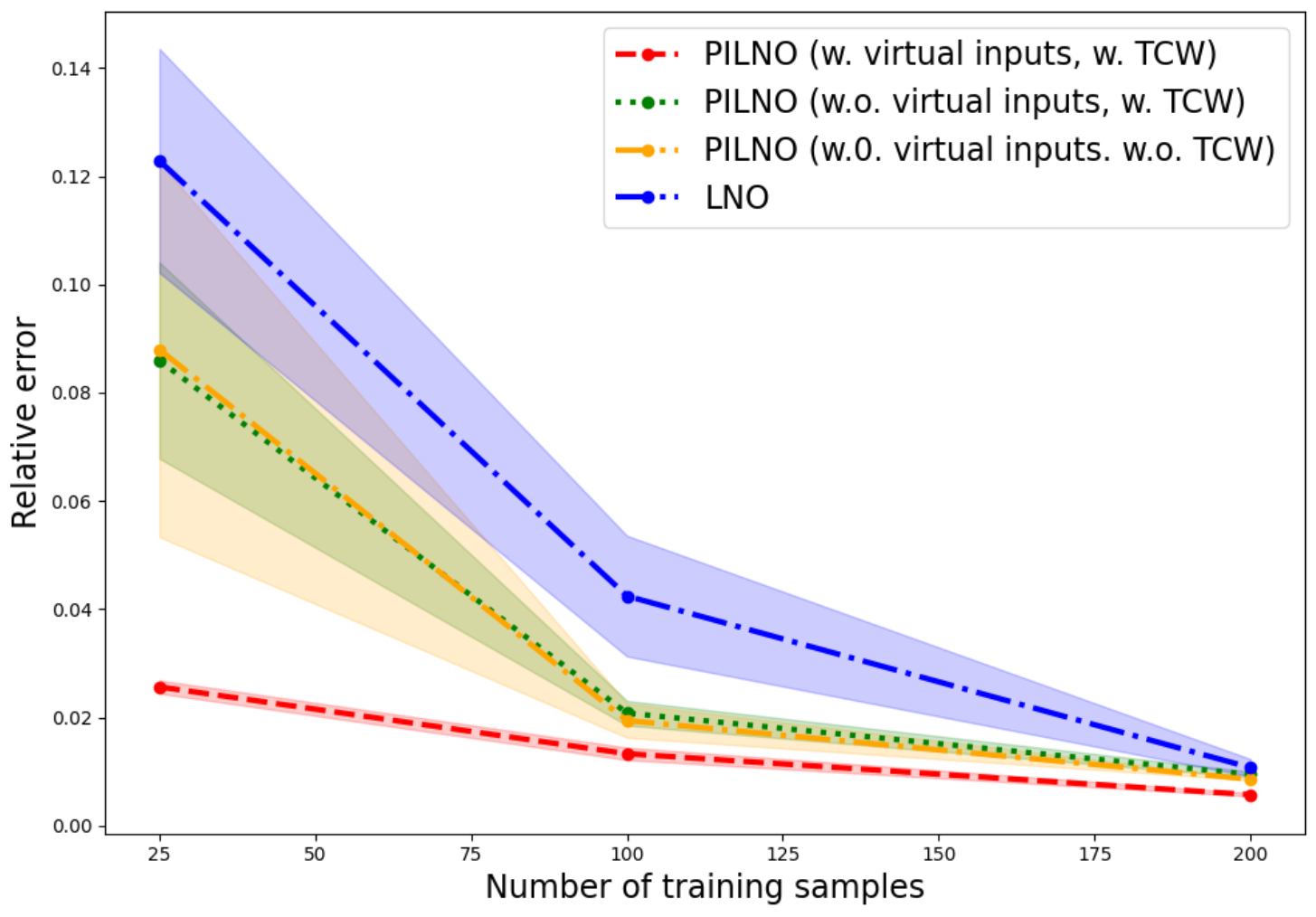}
        \caption{Relative $L_2$ error on the test set versus the number of training samples (mean $\pm$ std over five random seeds). In the absence of virtual inputs, TCW yields error curves comparable to those obtained without TCW across all training set sizes.}
        \label{fig:RD1D_tcw_wo_vi}
    \end{figure}

    Figure~\ref{fig:RD1D_tcw_wo_vi} supports this interpretation. Without virtual inputs, TCW and non-TCW models achieve comparable mean test errors across training-set sizes, while TCW yields a modest reduction in the standard-deviation band, indicating improved run-to-run stability. Overall, TCW remains mildly beneficial in the labeled-only regime, but its effect is substantially weaker than when combined with virtual inputs, where physics-driven optimization is more prominent.
    
\section*{Appendix D. Detailed setting for forced KdV equation}\label{appendix:fkdv}
    In this section, we detail the data generation procedure used for the forced KdV example and provide a description of the DeepOMamba model architecture employed in our experiments.
\paragraph{Data generation}
    Following the DeepOMamba~\cite{hu2025deepomamba} benchmark in forced KdV equation, we generate data from three analytic solution families, denoted as Types A, B, and C. We have three forcing types, formulated as
    \begin{equation}
        \begin{aligned}
            f_A(x,t) &= \frac{12k\beta/\alpha \left[ k^3(4\beta-\delta) - bA/(1+A^2t^2)-b_1 \right]}{\cosh^2[k(x-\delta k^2t)-b\arctan (At) - b_1t - b_0]} \\
            f_B(x,t) &= \frac{12k\beta/\alpha [k^3(4\beta - \delta) - (2b_2t+b_1)\exp(b_2t^2 + b_1t + b_0)]}{\cosh^2[k(x-\delta k^2t) - \exp(b_2t^2 + b_1t + b_0)]} \\
            f_C(x,t) &= \frac{12\beta \gamma}{\alpha} \left[ \frac{6\beta(\gamma+1)}{F(x,t)^2} - \frac{8\beta(x+a)^2}{H(x,t)^2F(x,t)^3} - \frac{(t+b)H(x,t)^2}{(x+a)F(x,t)} - \frac{(t+b)H(x,t)^3}{(x+a)^3}\arctan(H(x,t))  \right]
        \end{aligned}
    \end{equation}
    and corresponding solutions
    \begin{equation}
        \begin{aligned}
            u_A(x,t) &= \frac{12\beta k^2}{\cosh^2[k(x-\delta k^2t)-b\arctan (At) - b_1t - b_0]} \\
            u_B(x,t) &= \frac{12\beta k^2}{\cosh^2[k(x-\delta k^2t) - \exp(b_2t^2 + b_1t + b_0)]} \\
            u_C(x,t) &= \frac{12\beta \gamma}{F(x,t)}
        \end{aligned}
    \end{equation}
    where $F(x,t) = (x+a)^2 + (t+b)^2 + d$ and $H(x,t) = \frac{x+a}{\sqrt{d+(b+t)^2}}$. Here, $k,a,A,\alpha,\beta,\delta,\gamma,b_0,b_1,b,b_2,d$ are the coefficients of the forced KdV equation.
    
\paragraph{DeepOMamba baseline}

We use a $N_x\times N_t=100\times100$ space-time grid with $C_{\mathrm{in}}=7$ input channels. The Mamba model is configured with hidden dimension $d_{\mathrm{model}}=256$, $n_{\mathrm{layer}}=1$. The network uses LayerNorm with $\epsilon=10^{-5}$ and linear projections map $700\!\to\!256$ at the input and $256\!\to\!100$ at the output. Moreover, in training procedure, we use Adam with learning rate 0.001, batch size 16, and $E=100$ epochs. We apply a linear learning-rate decay via \texttt{LambdaLR} with $\eta_e=\eta_0(1-e/E)$, and optimize the relative $L_2$ loss $\|\hat{u}-u\|_2/\|u\|_2$.

\section*{Appendix E. Practical hyperparameter guidance and scalability}\label{appendix:hyperparameter_and_scalability}
    \paragraph{Implementation details and practical hyperparameter guidance}
    \tcb{Tables~\ref{tab:hyperparameter_model}--\ref{tab:hyperparameter_optimization} summarize the benchmark-specific architecture and physics-training configurations together with the common optimization settings used in our experiments. These configurations should be interpreted as the settings used to obtain the reported results rather than as universally optimal choices. In practice, the model depth, channel width, and numbers of retained Fourier and pole-residue modes should be selected based on validation performance and the spectral complexity of the target solution. Increasing these quantities enhances model capacity but also increases computational and memory costs.}

    \tcb{The virtual-input configuration involves two related choices: the sampling distribution \(\mathbb{P}_{\mathrm{virt}}\) and the number of virtual inputs \(N_{\mathrm{virt}}\). First, following the principles described in Section~\ref{subsec:virtual_dataset}, \(\mathbb{P}_{\mathrm{virt}}\) should be defined over admissible problem inputs for which the PDE and IC/BC residuals can be evaluated without reference solutions. When prior knowledge about the intended deployment regime is available, it may also be incorporated into \(\mathbb{P}_{\mathrm{virt}}\) to broaden physics supervision beyond the paired-data distribution. In the Gaussian-random-field benchmarks considered in this study, we implemented this principle by varying the correlation length scale, thereby exposing the model to input functions with a broader range of spectral characteristics.}

    \tcb{Second, \(N_{\mathrm{virt}}\) should be treated as an accuracy-cost hyperparameter. Although virtual inputs can be generated without solving for the corresponding reference solution fields, they are not computationally free because each input requires additional physics-residual evaluations during training. Increasing \(N_{\mathrm{virt}}\) provides denser coverage of \(\mathbb{P}_{\mathrm{virt}}\) and can improve generalization beyond the paired-data distribution, but it also increases the total wall-clock training cost. We therefore recommend increasing \(N_{\mathrm{virt}}\) until the validation error or validation residual shows
no meaningful further improvement.}

    \tcb{For the time-dependent benchmarks, we used fixed epoch-independent exponential TCW profiles. A larger decay rate $\gamma_{\mathrm{TCW}}$ places stronger emphasis on early-time residuals, whereas a smaller value approaches uniform temporal weighting. However, an excessively large value may suppress late-time residuals too strongly and delay the learning of later dynamics. Adaptive TCW schedules constitute a possible extension, but they were not evaluated in the present study since fixed monotone weighting schedules were sufficient to improve accuracy.}

\begin{table}[!htbp]
    \centering
    \small
    \setlength{\tabcolsep}{10pt}
    \renewcommand{\arraystretch}{1.7}
    \newcommand{\headerpad}{\rule[-4pt]{0pt}{24pt}}
    \begin{tabular}{c|c| c| c| c| c}
    \hline
    \headerpad
    Problem & \makecell{Number of \\ ALNO layers} & Width & Fourier modes & \makecell{Pole-residue \\ modes} & Activation \\
    \hline
    Burgers & 4 & 32 & 14 & 14 & tanh  \\
    Darcy flow & 4 & 64 & 20 & 20 & gelu  \\
    \makecell{Reaction-diffusion\\Task A} & 1 & 32 & 14 & 14 & sin  \\
    \makecell{Reaction-diffusion\\Task B} & 4 & 64 & 20 & 20 & sin  \\
    Forced KdV & 2 & 24 & 20 & 20 & gelu  \\
    \hline
    \end{tabular}
    \caption{Architecture configurations of the ALNO backbone used for each benchmark. Within each benchmark, PILNO and ALNO use the same backbone architecture and differ only in the definition of the training objective. The Fourier and pole-residue mode counts are reported per dimension.}
    \label{tab:hyperparameter_model}
\end{table}

\begin{table}[!htbp]
    \centering
    \small
    \setlength{\tabcolsep}{10pt}
    \renewcommand{\arraystretch}{1.7}
    \newcommand{\headerpad}{\rule[-4pt]{0pt}{24pt}}
    \begin{tabular}{c|c| c| c| c| c}
    \hline
    \headerpad
    Problem & \makecell{$N_{\mathrm{virt}}$} & \makecell{$B_{\mathrm{data}}$} & \makecell{$B_{\mathrm{phys}}$} & \makecell{TCW profile,\\$w(t)$} & TCW update scheme \\
    \hline
    Burgers & 1000 & 3 & 50 & $10e^{-5t}$ & Fixed  \\
    Darcy flow& 900 & 1 & 50 & \makecell{N/A\\(steady problem)} & N/A  \\
    \makecell{Reaction-diffusion\\Task A} & 500 & 3 & 50 & $10e^{-5t}$ & Fixed  \\
    \makecell{Reaction-diffusion\\Task B} & 500 & 3 & 50 & $10e^{-0.5t}$ & Fixed  \\
    Forced KdV & 1800 & 3 & 50 & $10e^{-2t}$ & Fixed  \\
    \hline
    \end{tabular}
    \caption{Problem-specific physics-informed training configurations used for PILNO. Here, $N_{\mathrm{virt}}$ denotes the total number of virtual inputs, $B_{\mathrm{data}}$ is the labeled-data batch size, and $B_{\mathrm{phys}}$ is the batch size of virtual input used for physics residual evaluation. For time-dependent problems, the TCW profile is fixed throughout optimization and not applicable to the steady-state Darcy flow problem.}
    \label{tab:hyperparameter_training}
\end{table}

\begin{table}[!htbp]
    \centering
    \small
    \setlength{\tabcolsep}{10pt}
    \renewcommand{\arraystretch}{1.7}
    \newcommand{\headerpad}{\rule[-4pt]{0pt}{24pt}}
    \begin{tabular}{c|c}
    \hline
    \headerpad
    \makecell{Optimization\\hyperparameter} & Common setting \\
    \hline
    Optimizer & Adam \\
    Adam coefficients & \((\beta_1,\beta_2)=(0.9,0.999)\) \\
    Initial learning rate & 0.01 \\
    Weight-decay coefficient & 1e-4 \\
    Learning-rate schedule & Step decay (StepLR) \\
    Scheduler step size & 100 \\
    Learning-rate decay factor $\gamma$ & 0.5 \\
    Maximum training epochs & 1000 \\
    Early stopping patience & 500 \\
    \hline

    \end{tabular}
    \caption{Common optimization configuration used for all ALNO and PILNO experiments unless otherwise stated. The scheduler step size and early stopping patience are expressed in epochs. Benchmark-specific baselines, such as DeepOMamba, follow their own optimization configurations described separately in Appendix~D.}
    \label{tab:hyperparameter_optimization}
\end{table}

    \paragraph{Scalability considerations}

    \tcb{PILNO inherits the resolution-dependent computational cost of its ALNO backbone and incurs additional training overhead from physics-residual evaluation. Finer residual grids and higher-dimensional tensor-product domains increase forward/backward operations and derivative evaluations, often requiring residual subsampling, smaller physics batches, mixed precision, or domain decomposition to control memory usage and training time. Because physics residuals, virtual inputs, and TCW are used only during training, PILNO and ALNO have the same inference cost at a fixed resolution. However, this cost still grows with the evaluation-grid resolution.}

    \tcb{To empirically examine this resolution dependence, we varied the residual-grid resolution in the Darcy-flow problem from \(31\times31\) to \(61\times61\) and \(121\times121\). This experiment is intended as a resolution-sensitivity study rather than a comprehensive asymptotic scalability benchmark. We used a simplified PILNO configuration with \(N_{\rm train}=200\), \(N_{\rm virt}=400\), and 10 Fourier and pole-residue modes per dimension. All runs were conducted on the same NVIDIA A100 GPU using the same model configuration and training protocol, except for the grid resolution. The results are summarized in Table~\ref{tab:darcy_scale}.}

    \tcb{Both wall-clock training time and peak GPU memory increased substantially with the grid resolution. The training time increased from \(778\) s on the \(31\times31\) grid to \(2139\) s on the \(61\times61\) grid and \(7114\) s on the \(121\times121\) grid. Peak GPU memory exhibited a similar increase, reaching approximately \(32.6\) GiB at \(121\times121\). In contrast, the relative \(L_2\) test error did not decrease monotonically with grid refinement. The error decreased from \(2.753\times10^{-2}\) at \(31\times31\) to \(2.183\times10^{-2}\) at \(61\times61\), but remained at a comparable level, \(2.290\times10^{-2}\), when the resolution was further increased to \(121\times121\). This observation suggests that, once the residual grid is sufficiently fine to resolve the dominant spatial structures relevant to the learned operator, further grid refinement may produce diminishing accuracy gains while continuing to increase computational cost. For this Darcy-flow setting, the \(61\times61\) grid therefore provided a favorable accuracy-cost trade-off among the tested resolutions. This conclusion is specific to the present benchmark and model configuration and should not be interpreted as a universally optimal residual-grid resolution.}

    \begin{table}[!htbp]
        \centering
        \small
        \setlength{\tabcolsep}{10pt}
        \renewcommand{\arraystretch}{1.6}
        \newcommand{\headerpad}{\rule[-4pt]{0pt}{24pt}}
        \begin{tabular}{c|c|c|c|c|c}
        \hline
        \headerpad
        Problem & GPU & Grid & \shortstack{Wall-clock\\training time} & \shortstack{Peak\\GPU memory} & \shortstack{Relative $L_2$\\test error} \\
        \hline
        \multirow{3}{*}{Darcy Flow} & \multirow{3}{*}{\shortstack{A100\\(40 GB)}} 
        & 31 $\times$ 31 & 778 s & 3660 MiB & 2.753e-2  \\
        \cline{3-6}
        & & 61 $\times$ 61 & 2139 s & 10024 MiB & 2.183e-2  \\
        \cline{3-6}
        & & 121 $\times$ 121 & 7114 s & 33378 MiB & 2.290e-2  \\
        \hline

        \end{tabular}
        \caption{Empirical resolution-sensitivity study of PILNO on the Darcy-flow problem. The experiments use a simplified configuration with \(N_{\rm train}=200\), \(N_{\rm virt}=400\), and 10 Fourier and pole-residue modes per dimension. All experiments were conducted on the same NVIDIA A100 GPU using the same training protocol except for the residual-grid resolution.  Increasing the grid resolution substantially increases wall-clock training time and peak GPU memory, whereas the relative \(L_2\) test error remains at a comparable level between the \(61\times61\) and \(121\times121\) settings.}
        \label{tab:darcy_scale}
    \end{table}

\bibliographystyle{elsarticle-num}
\bibliography{refer}

\end{document}